\documentclass[10pt,twocolumn,letterpaper]{article}

\usepackage[pagenumbers]{cvpr}

\usepackage[accsupp]{axessibility} % Improves PDF readability for those with visual impairments.

\usepackage{graphicx}
\usepackage{blindtext}
\usepackage{natbib}
\usepackage{adjustbox}
\usepackage{subcaption}
\usepackage{caption}
\usepackage[font=small,labelfont=bf]{caption}
\usepackage{capt-of}
\usepackage{hhline}

\usepackage{multirow}
\usepackage{soul}
\usepackage[normalem]{ulem}

\usepackage[dvipsnames]{xcolor}

\definecolor{cvprblue}{rgb}{0.21,0.49,0.74}
\usepackage[pagebackref,breaklinks,colorlinks,citecolor=cvprblue]{hyperref}

\def\paperID{4186} % *** Enter the Paper ID here
\def\confName{CVPR}
\def\confYear{2024}

\title{Action-slot: Visual Action-centric Representations for Multi-label Atomic Activity Recognition in Traffic Scenes}

\author{Chi-Hsi Kung\textsuperscript{1}
\and
Shu-Wei Lu\textsuperscript{1}
\and
Yi-Hsuan Tsai\textsuperscript{2}
\and
Yi-Ting Chen\textsuperscript{1}
\and
National Yang Ming Chiao Tung University\textsuperscript{1}\\
{\tt\small \{chkung,tomy45651.sc06,ychen\}@nycu.edu.tw}
\and
Google\textsuperscript{2}\\
{\tt\small yhtsai@google.com}
}

\begin{document}
% \maketitle
\twocolumn[{%
    \renewcommand\twocolumn[1][]{#1}%
    \maketitle
    \begin{center}
        \centering
        \captionsetup{type=figure}
        \includegraphics[width=1.0\textwidth]{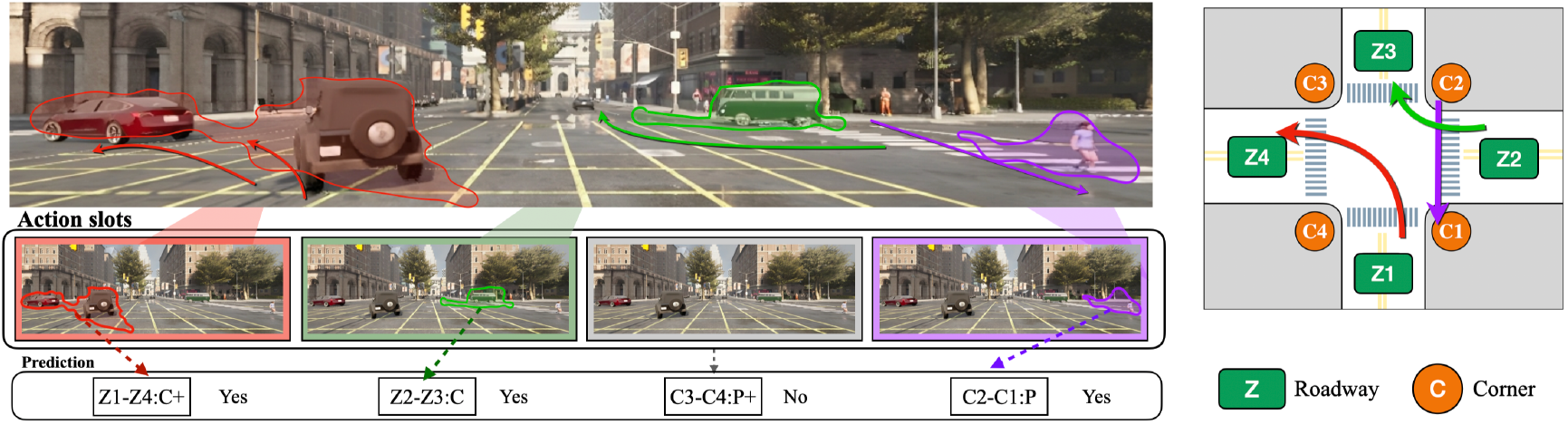} 
        \captionof{figure}{
        Illustration of the concept of multi-label atomic activity recognition and our proposed Action-slot. In the scene, three atomic activities are presented and depicted by colored arrows. For example, the red arrow represents the \textbf{Z1-Z4: C+} atomic activity, indicating a group of vehicles turning left. Atomic activities are defined based on road user's type and their motion patterns grounded in the underlying road structure. We introduce Action-slot to learn visual action-centric representations that enable decomposing multiple atomic activities in videos. We demonstrate that our framework can effectively recognize multiple atomic activities via learned representations.
%We 
%a slot attention-based framework that 
%with specific slots to focus regions (indicated by colored borders) corresponding to the respective atomic activity.  
%The framework achieves compelling performance on multi-label atomic
%activity recognition.
        %Illustration of the key idea of Action-slot.  Distinct categories of traffic atomic activities in the upper image, represented by colored arrows, are defined based on types of road users and their motion on the road topology. Our goal is to learn visual action-centric representations that decompose each individual activity from the scene by leveraging slots to pay attention (as shown in the colored borders) to each category of activity. 
        %\ychen{several concerns: 1. the slot concept is unclear. 2. the figure looks like a segmentation problem. 3. what are atomic activities? Need more effort.}
            % Our insight is to leverage slot attention to improve atomic activity recognition by learning compositional action-centric representations. To help the slots discover atomic activities, Action-slot excludes the background region from the \textit{competition space} of action slots.
            }
    \label{fig:teaser}
\end{center}%
}]

\begin{abstract}
In this paper, we study multi-label atomic activity recognition. Despite the notable progress in action recognition, it is still challenging to recognize atomic activities due to a deficiency in holistic understanding of both multiple road users’ motions and their contextual information. In this paper, we introduce Action-slot, a slot attention-based approach that learns visual action-centric representations, capturing both motion and contextual information. Our key idea is to design action slots that are capable of paying attention to regions where atomic activities occur, without the need for explicit perception guidance. To further enhance slot attention, we introduce a background slot that competes with action slots, aiding the training process in avoiding unnecessary focus on background regions devoid of activities. Yet, the imbalanced class distribution in the existing dataset hampers the assessment of rare activities. To address the limitation, we collect a synthetic dataset called TACO, which is four times larger than OATS and features a balanced distribution of atomic activities. To validate the effectiveness of our method, we conduct comprehensive experiments and ablation studies against various action recognition baselines. We also show that the performance of multi-label atomic activity recognition on real-world datasets can be improved by pretraining representations on TACO. Our source code, dataset, and visualization videos are available at \href{https://hcis-lab.github.io/Action-slot/}{https://hcis-lab.github.io/Action-slot/}
\end{abstract}    
\label{sec:intro}

\begin{figure*}[t!]
\centering
    \includegraphics[width=16cm]{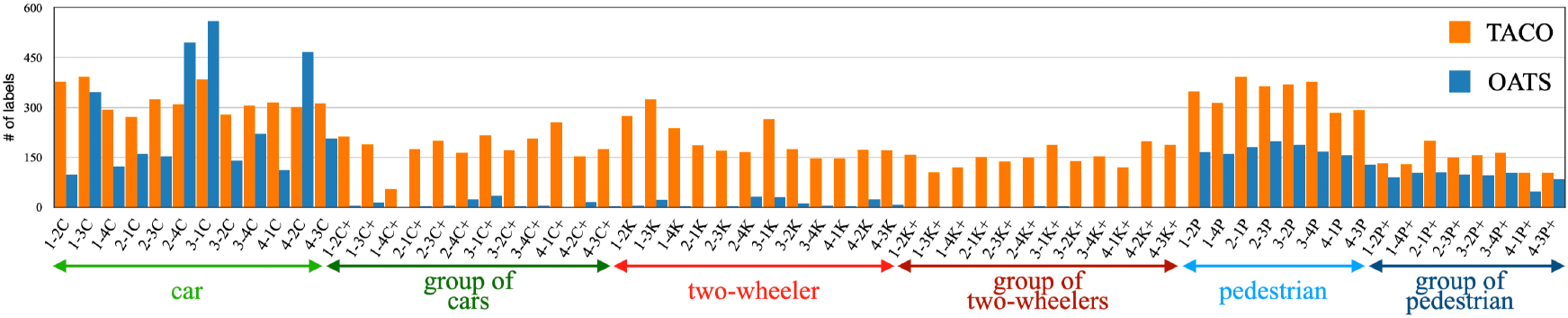}
        \caption{
        The distribution of atomic activity classes in our TACO dataset compared to the OATS dataset. Please note that, for space considerations, we omit the road topology notations of corner (C) and roadway (Z) on the x-axis of the figure.
        %Atomic activity class distribution of our TACO and the OATS dataset. Note that we neglect the road topology notation of C and Z in the x-axis of the figure for the space.
        }
\label{fig:distribution}
\vspace{-3mm}
\end{figure*}

\section{Introduction}
%Traffic activity recognition \ychen{i felt this term should be changed}\hank{done}, enabling understanding of individual events in the scenes shown in the upper illustration in Figure \ref{fig:teaser}, is a critical aspect of the development of intelligent driving systems.
There has been a noteworthy rise of community interest in activity recognition within traffic scenes~\cite{chen2016atomic,ramanishka2018toward,pieiccv2019,malla2020titan,stipicra2020,singh2022road,Agarwal_2023_ICCV}. 
Recognizing activities of road users is essential for advancing the development of intelligent driving systems, as it enables various applications, such as intent prediction~\cite{brain4car2015,stipicra2020,semanticregioncorl2022}, scenario retrieval~\cite{lin2014visual,driverwacv2020,Segal_universal_corl2020,Naphade23AIC23}, scenario-based assessment~\cite{carlachallenge2022,xu2022safebench}, and safety-critical scenario generation~\cite{xu2022safebench,rempe2022strive,cao2022advdo,rempeluo2023tracepace}.  

In this paper, we focus on multi-label atomic activity recognition, a new 
% traffic activity recognition 
task proposed by the OATS dataset for interactive traffic scenario understanding~\cite{Agarwal_2023_ICCV}. 
%
%An atomic activity is defined with a road topology-aware description that formulates the movement of one or grouped road users on road topology, as shown in Figure~\ref{fig:teaser}.
An atomic activity is a higher-level semantic motion pattern rooted in the underlying road topology.
For instance, a group of vehicles turning left is described as \textbf{Z1-Z4: C+}, as shown in Figure~\ref{fig:teaser}.
The notions \textit{Z1} and \textit{Z4} are roadways within an intersection.
The left-turn motion pattern is represented as \textbf{Z1-Z4}.
%as road topology.  \textit{C1} to \textit{C4} \textit{Z1} to \textit{Z4}.
%
We use \textit{C+} to denote a group of vehicles.
The combination of such motion patterns and road user types can compose 64 classes of atomic activities.
This task presents distinctive challenges as it demands a comprehensive understanding of contextual information 
% related to road topology 
and motions of road users.
Moreover, it requires the ability to break down activities within crowded and intricate traffic scenes from videos.
%The task poses unique challenges of requiring a holistic understanding of context information for road topology and road users to recognize the actions and the ability to decompose activities from crowded and complex traffic scenes. 

We identify the limitations of current action recognition models, showing less desirable performance in the OATS benchmark~\cite{Agarwal_2023_ICCV}. 
We argue that video-level representations, 
%which model the entire video with unified feature maps using 3D ConvNet~\cite{carreira2017quo,tran2019video,feichtenhofer2019slowfast,feichtenhofer2020x3d} or transformer~\cite{fan2021multiscale,tong2022videomae}, 
simply employing 3D ConvNet~\cite{carreira2017quo,tran2019video,feichtenhofer2019slowfast,feichtenhofer2020x3d} or transformers~\cite{fan2021multiscale,tong2022videomae} models,
% that capture the entire video with unified feature maps, 
pose a challenge for classifiers in distinguishing individual atomic activities within traffic scenes due to the complicated structure of traffic scenes.
% present a challenge for classifiers to distinguish individual atomic activities within traffic scenes because of complex compositionality of traffic scenes.
%video-level representations modeling the whole video into unified feature maps with 3D ConvNet~\cite{carreira2017quo,tran2019video,feichtenhofer2019slowfast,feichtenhofer2020x3d} or transformer~\cite{fan2021multiscale,tong2022videomae} make it difficult for the classifier to distinguish each atomic activity from the scenes. 
%
On the other hand, object-aware representations~\cite{Baradel_2018_ECCV,CVPR2019_ARG,Agarwal_2023_ICCV} using object features~\cite{he2017mask} struggle to associate the relationship between the road structures and motions of objects. 
%
%To explore the alternatives that can address the limitation of existing action recognition methods for traffic atomic activity recognition, 
%
Then the question we like to answer in this paper: \textit{Can we learn visual representations that decompose multiple atomic activities from videos, without using object proposals?} 

% \ychen{i felt "compositional" is not a good term because of the following reasons.}
% \textcolor{red}{
% \begin{enumerate}
%     \item the term "compositional" implies it composes of other sub-concepts, e.g., activity composes "object" and "action." As you said in the text, what you do is "decompose" the scene. I felt "compositional scene representations" might be a better term.
%     \item however, i do not encourage to put "compositional scene representations" because others would ask you to prove if you learn "compositional scene representations," which is not the focus of your work.
%     \item another note: if you really want to push for this term, your text (you only use this term twice in your paper) should show your intention, e.g., related work should discuss them. experiments should prove it. if not, i felt you just want to use a "fancy" term.  
%     \item to sum up, i felt the term "visual action-centric representation" suffices, and that is the term you used in the title.
% \end{enumerate}
% }

To this end, we propose Action-slot, a slot attention-based framework, inspired by the recent success of slot attention for unsupervised object discovery~\cite{locatello2020object}.
% To this end, we propose Action-slot to learn visual action-centric representations from videos, inspired by the recent success of slot attention that decomposes objects from images~\cite{locatello2020object}.
%,  for atomic activity recognition. learns object-centric representations 
%
%we leverage a slot to represent a category of atomic activity, which decomposes the scene into activities, as shown in Figure \ref{fig:teaser}.
% Our insight is that slot attention uses a slot to represent an object has demonstrated its decomposition ability on object-aware tasks, such as object discovery~\cite{locatello2020object,kipf2022conditional,elsayed2022savi++,bao2022discovering,biza2023invariant}, panoptic segmentation~\cite{zhou2022slot}, and, novel view synthesis~\cite{NEURIPS2022_3dc83fcf}. 
Action-slot incorporates four crucial design choices that allow the model to decompose multiple atomic activities from videos and represent them without relying on object proposals.
First, we allocate a fixed number of action slots, assigning each action slot to focus on regions where a specific atomic activity occurs, such as \textbf{C2}-\textbf{C1: P}. 
Second, we introduce an additional background slot and attention guidance to enforce the slot paying attention to regions without active atomic activities. 
Third, we discourage action slots allocated with negative classes (i.e., those
activities not present in a video), from attending to any regions.
%
% we encourage the remaining slots to concentrate on regions that differ from the attention of the background slot by designing a novel loss function.
%
Fourth, we modify the slot updating strategy used in~\cite{kipf2022conditional,elsayed2022savi++,bao2022discovering} to a parallel updating approach that integrates spatial-temporal information at the video level into a slot.
In our experiments, we demonstrate the effectiveness of \textit{Action-slot} in accurately identifying a variety of atomic activities annotated in the OATS dataset~\cite{Agarwal_2023_ICCV}.
Furthermore, we conduct thorough experiments, comparing various action recognition baselines and presenting ablation studies that validate the design of Action-slot.
Moreover, we provide qualitative evidence demonstrating that the learned action-centric representations can reliably identify distinct atomic activities in the visual domain, even without the need for supervision signals such as object locations~\cite{elsayed2022savi++}.
However, due to imbalanced class distribution within the dataset, OATS only uses 35 out of 64 atomic activity classes for training and evaluation.
%
%because the activities are rare in the real world,
%
%Particularly, no activities involved with group two-wheelers (\textit{K+}). 
Specifically, there are no activities associated with group two-wheelers (\textit{K+}).
%
% This remains the analysis for nearly half of the classes of activities unknown and hinders the evaluation for rare classes.
This observation extends to nearly half of the activity classes, impeding the evaluation of rare classes.

To address this limitation, we introduce the Traffic Activity Recognition  (TACO) dataset, an extensive dataset for atomic activity recognition. 
We utilize the CARLA simulator~\cite{Dosovitskiy17} to gather instances of all conceivable activity classes, ensuring a well-balanced distribution, as illustrated in Figure~\ref{fig:distribution}.
% To address the limitation, we propose Traffic activity reCOgnition dataset (TACO), the largest dataset for atomic activity recognition. 
% %
% We leverage the CARLA simulator~\cite{Dosovitskiy17} to collect all possible classes of activity and achieve a balanced distribution, as shown in Figure~\ref{fig:distribution} (b). 
%
We again benchmark all methods on TACO and show that Action-slot outperforms all baselines by a large margin. 
%
%More interesting, We find that Action-slot can localize atomic activities spatial-temporally by training with weak supervision (i.e., video-level activity labels) and without perception modules (e.g., object detectors). We conduct comprehensive ablation studies and qualitative results to demonstrate the effectiveness of the proposed design for atomic activity recognition. 

Beyond comprehensive performance analysis, we demonstrate that TACO can enhance the efficacy of atomic activity recognition on real-world datasets through pretraining feature representations on TACO.
%We demonstrate the real-world value of TACO by conducting transfer learning from sim-to-real and show the dataset can provide strong pertaining representations. 
%
%Specifically, we fine-tune the pretrained model on OATS and a new annotated nuScenes dataset~\cite{nuscenes}. 
%
Then we finetune the pretrained model on both the OATS dataset and a newly annotated 
% real-world dataset utilizing the 
nuScenes dataset~\cite{nuscenes}.
We present significant improvements in all experiments through TACO pre-training, underscoring the robust real-world transferability of TACO. Our main contributions are as follows:
%We find all models obtain significant improvement from TACO pre-training, which shows the strong real-world transferability of the TACO.
\begin{enumerate}
    \item We introduce \textit{Action-slot}: an action-centric slot attention-based framework that decomposes multiple atomic activities from videos.
    %We propose \textit{Action-Slot}: an action-centric slot attention-based framework with modified architectural designs enabling traffic atomic activity recognition.
    \item We present the TACO dataset, an extensive and balanced dataset for multi-label atomic activity recognition, facilitating comprehensive performance analysis. 
    %We introduce the TACO dataset, the largest, full-category, and balanced traffic atomic activity recognition dataset for comprehensive performance analysis.
    % \ychen{i am not convinced why the dataset can enable developing and evaluating models in a more systematic manner. we should focus on the benefit of a balanced dataset.}
    \item We conduct extensive evaluations and ablation studies across multiple datasets to justify the effectiveness and generalization of Action-slot.
    %Extensive experimental results show that Action-slot achieves state-of-the-art atomic activity recognition performance on multiple datasets.
    \item We provide qualitative evidence demonstrating that the learned action-centric representations can reliably identify distinct atomic activities in the visual domain without perception modules. 
    %We demonstrate that \textit{Action-Slot}'s action-centric representations can localize atomic activities via attention maps with weak supervision. 
\end{enumerate}

\label{sec:related_work}
\section{Related Work}
\paragraph{Traffic Scenario Understanding Datasets.}
The development of intelligent driving systems, such as scene analysis~\cite{ramanishka2018toward,stipicra2020,semanticregioncorl2022}, scenario retrieval~\cite{lin2014visual,driverwacv2020,Segal_universal_corl2020,Naphade23AIC23}, and safety-critical scenario generation~\cite{xu2022safebench,rempe2022strive,cao2022advdo,rempeluo2023tracepace}, heavily relies on traffic activity understanding.
While previous efforts have designed datasets~\cite{chen2016atomic,ramanishka2018toward,li2019dbus,malla2020titan} by labeling high-level actions (e.g., left turn) for traffic pattern recognition, they fall short in supporting fine-grained scenario analysis. 
%Most previous efforts have designed datasets~\cite{chen2016atomic,ramanishka2018toward,li2019dbus,malla2020titan} by labeling high-level actions (e.g., left turn) for traffic activity recognition, they fall short in supporting fine-grained scenario analysis. 
%
For instance, high-level actions cannot differentiate ``left turns'' that start from different locations, such as the ego lane versus the oncoming lane.
To address this limitation, both Inner-City~\cite{chen2016atomic} and ROAD~\cite{singh2022road} datasets augment high-level action labels (e.g., turning left) with location labels (e.g., on the oncoming lane or on our side of the road).
%While Inner-City~\cite{chen2016atomic} and ROAD~\cite{singh2022road} datasets augment high-level action labels (e.g., turning left) with location labels (e.g., on the oncoming lane or on our side of the road) for each frame as a detection problem. 
%However, the per-frame prediction requires temporal label aggregation when used for video analysis or scenario retrieval. 
%
Recently, OATS~\cite{Agarwal_2023_ICCV} propose a topology-aware traffic activity description language to unify the action and location labels via the topology-aware pattern description. 
Specifically, the language decomposes an interactive scenario into a set of atomic activities, that are defined based on the types of road users and the corresponding motion patterns grounded in road structures.
% They first define each roadway and corner in a 4-way intersection as the semantic regions \textit{R1}~\textit{R4} and \textit{C1}~\textit{C4}, as shown in the right of Figure~\ref{fig:teaser}. Then they associate objects with two semantic regions to describe a traffic pattern, as shown in the left of Figure~\ref{fig:teaser}. In this way, one can easily describe the traffic scenes with the temporal development of actions. 
% However, traffic atomic activities in the real world are rare and thus make it difficult to construct a well-balanced category distribution, which can make the analysis of rare classes of activities unclear. 
% %
% Therefore, we propose to collect traffic activities in the CARLA simulator~\cite{Dosovitskiy17} to construct a balanced distribution TACO dataset. 
Nevertheless, some real-world instances of traffic atomic activities are scarce, presenting a challenge in forming a well-balanced category distribution and impeding the evaluation of rare classes. 
In response, we construct the TACO dataset utilizing the CARLA simulator~\cite{Dosovitskiy17} to gather instances of all conceivable activity classes, ensuring a well-balanced distribution.

\vspace{-3mm}
\paragraph{Video Action Recognition.}
% As traffic atomic activity recognition can be framed as a multi-label action recognition, we revisit the existing literature.
Substantial improvements in spatial-temporal modeling via 3D ConvNet~\cite{carreira2017quo,feichtenhofer2020x3d,tran2019video,feichtenhofer2019slowfast} and transformer~\cite{bertasius2021space,arnab2021vivit,fan2021multiscale,tong2022videomae} have been observed because of the Kinetics human action dataset~\cite{kay2017kinetics} that facilitates model pre-training.
The pre-trained models can further be finetuned on other action recognition datasets such as the HMDB~\cite{kuehne2011hmdb}, THUMOS~\cite{THUMOS14}, and Charades~\cite{sigurdsson2016hollywood} datasets.
Meanwhile, the community also explores multi-label action recognition and establishes benchmarking datasets, such as
%On the other hand, there has been a line of research that focuses on multi-label action recognition. 
MultiTHUMOS~\cite{yeung2015every}, Charades~\cite{sigurdsson2016hollywood}, and AVA~\cite{gu2018ava}.
% datasets.
%, capturing one or few humans performing actions.
%
% , are the representative ones. 
%However, traffic activity recognition involves  is more challenging because it requires recognizing multiple activities present in the scenes.
%
The most relevant task to multi-label atomic activity recognition is group activity recognition~\cite{choi2009they,volleyball}, where multiple players engage in distinct actions. 
%
%However, most action labels in group activity recognition can be easily classified with the objects' appearance information. 
%
%For example, a ball-lifting player in the Volleyball~\cite{volleyball}. 
%
In contrast, atomic activity recognition in traffic scenes demands a comprehensive understanding of objects and their motion patterns within the underlying topology. 
In addition, a noteworthy distinction is that the majority of objects in traffic scenes do not exhibit any activity, constituting a negative class, a concept absent in group activity recognition datasets. 
% \dennis{One critical question, how do we reason that our Action-Slot cannot be applied to the existing multi-label action recognition datasets? Maybe we want to mention more differences from the existing datasets.}
\vspace{-3mm}
\paragraph{Slot Attention for Representation Learning.}
Locatello et al.,~\cite{locatello2020object} propose slot attention as a new perspective for learning object-centric representations in an unsupervised manner. 
They have successfully stimulated the community to explore strategies to bootstrap slot attention from synthetic datasets~\cite{johnson2017clevr} to real-world dynamic visual scenes with moving cameras such as the Waymo~\cite{sun2020scalability} and KITTI~\cite{Geiger2012CVPR} datasets.
Particularly, they utilize auxiliary information such as optical flow~\cite{kipf2022conditional}, motion segmentation~\cite{bao2022discovering}, object initial locations~\cite{elsayed2022savi++}, and depth~\cite{elsayed2022savi++} to facilitate object-centric representation learning in challenging real-world applications.

%In this work, we aim to learn action-centric representations for 
Multi-label atomic activity recognition poses three challenges for existing slot attention methods.
First, the slots are permutation invariant. This property is designed to discover arbitrary objects~\cite{locatello2020object,kipf2022conditional,elsayed2022savi++,bao2022discovering,zhou2022slot,NEURIPS2022_3dc83fcf,biza2023invariant} but is not suitable for a classification task with fixed-length classes. 
Second, current slot attention architectures used in videos, such as SAVi~\cite{kipf2022conditional} and Slot-VPS~\cite{zhou2022slot}, are designed for object-centric tasks, such as tracking. 
Specifically, they update a slot for each frame in a recurrent manner, which is not suitable for action recognition that requires a holistic temporal understanding. 
Third, to enable slot attention in complex scenes, they leverage additional object signals to guide the attention during training, such as flow~\cite{kipf2022conditional}, depth~\cite{elsayed2022savi++}, or motion segmentation~\cite{bao2022discovering}. 
Object signals may lead to misinterpretations in atomic activity recognition since not all objects participate in activities (e.g., pedestrians strolling on the sidewalk and vehicles waiting at traffic lights).
In this work, we propose a novel action-centric slot attention-based framework and demonstrate the effectiveness of the framework on multiple datasets.
% Object signals can be misleading in traffic atomic activity recognition because not all objects are involved in an activity (e.g., pedestrians wandering on sidewalk and vehicles waiting for traffic lights.
%
% in Figure~\ref{fig:teaser}) and multiple objects can contribute to a single activity (e.g., C3-C4:P+ in Figure~\ref{fig:teaser}).

% \begin{figure}
% \centering
% \begin{subfigure}{.3\textwidth}
%   \centering
%   \includegraphics[width=.5\linewidth]{Figure/slot_comparison_a.pdf}
%   \caption{Prior work: Recurrent}
%   \label{fig:recurrent}
% \end{subfigure}%
% \begin{subfigure}{.3\textwidth}
%   \centering
%   \includegraphics[width=.2\linewidth]{Figure/slot_comparison_b.pdf}
%   \caption{Action-Slot (Ours): Parallel}
%   \label{fig:parallel}
% \end{subfigure}
% \caption{(a) The slots are updated recurrently along the temporal dimension, where each time the slot only considers a frame-wise feature~\cite{locatello2020object,kipf2022conditional,elsayed2022savi++,bao2022discovering}.
% %
% (b) Our parallel scheme considers updating slots based on the spatial-temporal features for all the frames, which can be more suitable for the 
% traffic pattern recognition task.
% }
% \label{fig:slot-updating}
% \end{figure}

\begin{figure*}[t!]
\centering
    \includegraphics[width=17cm]{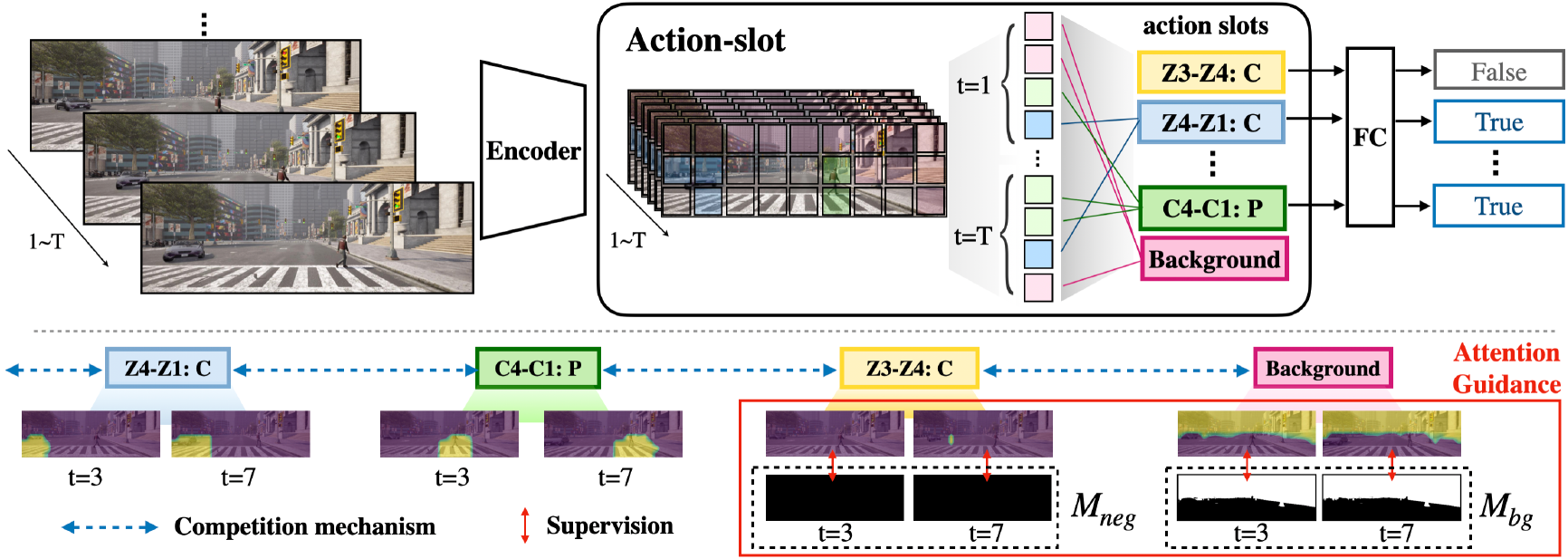}
        \caption{The top of the figure illustrates the proposed framework. Action-slot takes video as input and uses a CNN encoder to extract feature patches. All patches are then processed with individual slots simultaneously to find the most relevant spatial-temporal patches corresponding to each action slot. The updated action slots are fed into a fully connected layer to predict the probability of the corresponding action class, excluding the background slot. The bottom of the figure depicts the attention maps of action and background slots. We propose to incorporate a background mask $M_{\mathtt{bg}}$ to supervise the background slot. The design facilitates other action slots to capture action signals. Furthermore, we design a regularization for slots allocated to negative classes (e.g., Z3-Z4: C) using an all-zero mask $M_{\mathtt{neg}}$. 
        %Moreover, we regularize slots allocated to negative classes (e.g., Z3-Z4: C) with an all-zero mask $M_{\mathtt{neg}}$.
        }
        \label{fig:main_arch}
        \vspace{-4mm}
\end{figure*}

\label{sec:method}

\section{Method}
In this section, we first introduce the background of slot attention.
Then, we provide an overview of the problem definition in multi-label atomic activity recognition and the proposed method.
Finally, we explain the specific modifications we make to the slot attention for our task.

\subsection{Preliminary: Slot Attention}
\label{sec:slot_attention}
The Slot Attention module~\cite{locatello2020object} can be viewed as a clustering algorithm that maps image patches input to a set of $K$ output slots $S \in \mathbb{R}^{K \times D_{\mathtt{slot}}}$, where $D_{\mathtt{slot}}$ is the dimension of each slot.
%
% Slot attention can be viewed as a type of clustering method, where a fixed set of $K$ vectors $S \in \mathbb{R}^{K \times D_{slot}}$, where $D_{slot}$ denotes the dimension of each slot.
% \dennis{What is $D_{slot}$? Dimension of each slot?}
%
%Previous works exploit slot attention to segment objects in an image or video by matching each object to a slot \cite{locatello2020object,kipf2022conditional,elsayed2022savi++,bao2022discovering,zhou2022slot}.
%
% Slot attention has been successfully deployed for object segmentation and tracking~\cite{locatello2020object,kipf2022conditional,elsayed2022savi++,bao2022discovering,zhou2022slot}.
%
Specifically, the slots $S$ are first initialized by randomly sampling $K$ vectors from a Gaussian distribution, where the parameter $K$ is usually defined as the maximum number of objects in an image.
% or video.
An input frame, i.e., image features $F \in$ $\mathbb{R}^{H \times W \times D_{\mathtt{in}}}$ with size $H \times W$ and dimension $D_{\mathtt{in}}$, is first flattened to patch tokens $F^{'} \in$ $\mathbb{R}^{N \times D_{\mathtt{in}}}$, where $N = H \times W$. Then the tokens are mapped to the slots using the dot product attention module.
That is, the attention weight can be calculated as $A = \frac{1}{\sqrt{D}} k(F^{'}) \cdot q(S) \in \mathbb{R}^{N \times K}$, and $q$ and $k$ are linear transformations that map the input $F$ and slots $S$ to a common dimension $D$.

We obtain the updated values for slots through $U = \bar{A}^{T} \cdot v(F^{'}) \in \mathbb{R}^{K \times D}$, where $\bar{A}$ is the normalized attention weight calculated via the softmax operation and $v$ is a linear transformation. 
%
%With the normalized attention weight $\bar{A}$ via the softmax operation, we obtain the updated values for slots as $U = \bar{A}^{T} \cdot v(F) \in \mathbb{R}^{K \times D}$, where $v$ is a linear transformation.
%
% The slot attention then takes an image's features $I \in$ $\mathbb{R}^{H \times W \times D_{in}}$, where $H$, $W$, and $D_{in}$ denote features' height, width, and dimension, as input and computes the dot product with slots to obtain the attention weights $W = \frac{1}{\sqrt{D}} k(I) \cdot q(S) \in \mathbb{R}^{N \times K}$, where $N = H \times W$, and q and k are linear transformation.
% The weights are then used to compute the update values $U = W^{T}v(I) \in \mathbb{R}^{K \times D}$, where $W$ are the normalized
% attention weights and $v$ is another linear transformation.
%
% \dennis{Please do not just copy from the other paper, even though this can be common sentences...}
%
Finally, the slots are updated via a GRU: $S^{'} = \mathtt{GRU}(S, U)$~\cite{cho-etal-2014-learning}. To refine the slots, the updating process repeats $M$ iterations~\cite{locatello2020object,kipf2022conditional,elsayed2022savi++}.
Note that, the key difference between classical attention~\cite{vaswani2017attention} and slot attention is that the attention weights $A$ are normalized with the softmax operation slot-wise instead of token-wise. This difference enables slots to compete with each other so that each slot attends to different relevant regions of input. 
%
%occupies relevant regions on feature maps.
%
To extend slot attention to video tasks, the previous works~\cite{kipf2022conditional,elsayed2022savi++,bao2022discovering} propagate slots recurrently, i.e., $S^{t} = \mathtt{GRU}(S^{t-1}, U)$. 
% The recurrent updating mechanism is illustrated in Figure~\ref{fig:slot-updating} (a).
% ------------------------------------------------------------

\subsection{Overview of Proposed Method}
\label{sec:overview}
% We introduce the overall idea of our Action-slot framework for multi-label atomic activity recognition.

% \vspace{-2mm}
\paragraph{Problem Formulation.}
Given a video clip $V_i$ with $T$ image frames $\{I_t^i\}_{t=1}^T$, our goal is to recognize whether there are any atomic activities $Y$ presenting in the video. The ground truth of $Y^i$ for video $V_i$ is a binary multi-label vector, i.e., $Y^i= \{y_c\}_{c=1}^{N_{\mathtt{cl}}}$, where $y_c$ is 1 if the corresponding activity appears and vice versa, and $N_{\mathtt{cl}}$ denotes the total number of possible activities.
% (i.e., $N_{\mathtt{cl}} = 64$ for non-ego vehicle patterns.
Note that, we do not constrain when a certain activity appears, e.g., $y_c$ can happen in any image frames of $\{I_t^i\}_{t=1}^T$, or may appear multiple times in one video. 

\vspace{-5mm}
\paragraph{Overview of Action-slot.}
We propose Action-slot, an action-centric slot-attention-based framework to decompose multiple atomic activities from videos.
% for traffic atomic activity recognition.
Our idea is to assign each slot to learn specific action-centric representations for the corresponding atomic activity.
% Given video frames, we use a convolution-based encoder~\cite{he2016deep,carreira2017quo,feichtenhofer2019slowfast,feichtenhofer2020x3d} to extract features $F$
% 3D convolution-based encoder~\cite{feichtenhofer2020x3d} to extract spatial-temporal features $F$ 
% that are the input to our slot attention module.
Figure~\ref{fig:main_arch} presents the overview of Action-slot.
Different from existing slot attention algorithms designed for unsupervised object discovery and tracking \cite{kipf2022conditional,elsayed2022savi++,bao2022discovering}, we make the following three modifications.
First, we allocate a fixed number of slots $K$, which is equal to the number of atomic activities $N_{\mathtt{cl}}$. 
% In this work, we refer to our slots as action slots.
We supervise the prediction of each action slot with the corresponding ground truth label $y_c$, i.e., whether the atomic activity $c$ appears.
% or not.
% This is attributed to the fact that the atomic activities are temporally evolving,
% is not a recurrent problem but more like a video activity discovery task
% so it is more suitable to consider all the frames at the same time.
% (see Figure~\ref{fig:slot-updating} for a comparison with the prior works~\cite{kipf2022conditional,elsayed2022savi++,bao2022discovering})
Second, we introduce a background slot to focus on the regions that are not possible to present an activity, so that the other action slots can compete with this background slot.
In other words, the design forces other action slots to pay attention to the important regions.
Third, we discourage action slots associated with negative classes (i.e., those
activities not present in a video), from attending to any regions.
Lastly, we update all action slots in a parallel fashion by considering all the image frames together, instead of recurrently updating slots along the temporal domain utilized in~\cite{kipf2022conditional,elsayed2022savi++,bao2022discovering}.

\subsection{Action-slot}
\label{sec:action_slot}
% To alleviate the computation cost of the slot update process for videos, previous work~\cite{bao2022discovering} discard the iterative refinement and use learnable parameters for slots initialization, which is similar to learnable query set for object anchors proposed in DETR~\cite{carion2020end}.
In this section, we describe the details of our design for learning action-centric representations.
We follow \cite{bao2022discovering} to use learnable parameters for slot initializations.
We extract image features
% spatial-temporal features 
$F \in \mathbb{R}^{T \times H \times W \times D_{in}}$ from the
% 3D convolution-based encoder~\cite{feichtenhofer2020x3d}. 
convolution-based encoder~\cite{he2016deep,carreira2017quo,feichtenhofer2019slowfast,feichtenhofer2020x3d} for each frame in a video clip and then flatten them to $F^{'} \in \mathbb{R}^{N \times D_{\mathtt{in}}}$, where $N = T \times H \times W$.
We add learnable 3D spatial-temporal positional embeddings $E \in \mathbb{R}^{T \times H \times W} $ to the tokens $F^{'}$~\cite{bertasius2021space,arnab2021vivit,fan2021multiscale,bao2022discovering,tong2022videomae}.
% As shown in Figure~\ref{fig:slot-updating}(b),

\vspace{-5mm}
\paragraph{Allocated slot.}
% We then define a set of slots $\{S_k\}_{k=1}^{K}$, where $K = N_{\mathtt{cl}} + 1$, representing the number of action slots (i.e., number of activity classes) plus a background slot. 
We then define a set of slots $\{S_k\}_{k=1}^{K}$, where $K = N_{\mathtt{cl}}$, representing the number of action slots (i.e., number of activity classes).
In contrast to the previous slot attention works~\cite{locatello2020object,kipf2022conditional,elsayed2022savi++,zhou2022slot} where they do not specify a task to each slot and use a Hungarian matcher for bipartite matching with objects, we instead allocate each slot a class of activity, namely, action slots. Each action slot is processed with an independent binary classifier to output a prediction $\hat{y}_c$ for each class $c$, where a binary cross entropy (BCE) function is used as the objective for each data sample: $L_{\mathtt{act}} = \sum_{c=1}^{N_{\mathtt{cl}}} \textnormal{BCE}(\hat{y}_c, y_c)$.

\begin{figure}[t!]
\centering
    \includegraphics[width=8cm]{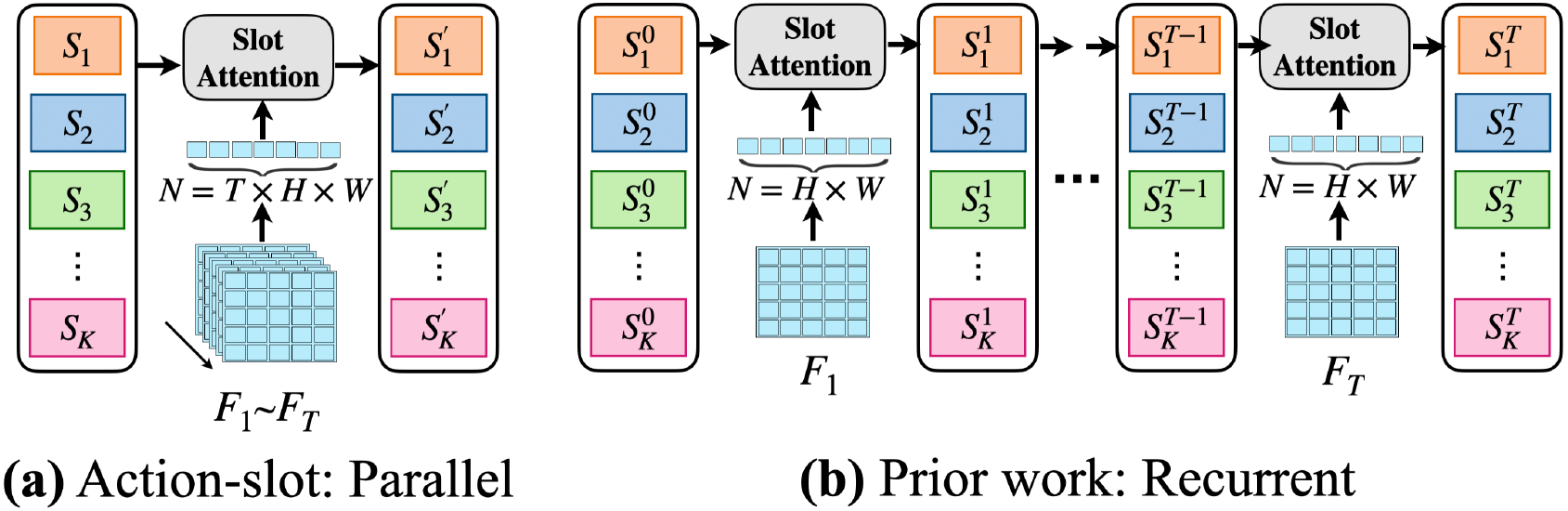}
        \vspace{-1mm}
        \caption{
       (a) Our parallel scheme considers updating slots based on the spatial-temporal features for all the frames.
       (b) The slots are updated recurrently along the temporal dimension, where each time the slot only considers a frame-wise feature.
        }
        \label{fig:parallel}
        \vspace{-4mm}
\end{figure}

\vspace{-4mm}
\paragraph{Parallel updating.}
We update slots across image frames in a parallel manner, as shown in Figure~\ref{fig:parallel} (a). Specifically, we calculate normalized attention weight $\bar{A} \in \mathbb{R}^{N \times K}$ while considering temporal dimension at the same time, i.e.,  $N = T \times H \times W$. We then update the $K$ slots with attention weights $\bar{A}$ in a single pass. This is different from the previous works~\cite{kipf2022conditional,elsayed2022savi++,bao2022discovering} where they update slots across frames recurrently, namely, using the previous state of slots and current feature maps to update the current state of slots for $T$ iterations, as shown in Figure~\ref{fig:parallel} (b).

% Note that, previous works \cite{kipf2022conditional} treat slots as equivariant permutation vectors and require a Hungarian matcher for bipartite matching.
% In contrast, we do not require the operation to output predictions because we allocate an action slot to a specific atomic activity.

\vspace{-5mm}
\paragraph{Background slot and attention guidance.}
% Although allocating the action slots to their corresponding activity classes helps the learning process, it remains a challenging task without additional perception guidance on where the action slots should pay attention.
% Moreover, 
% We discuss how to provide guidance to help actions find relevant regions.
It is non-trivial to provide 
% such
guidance for positive classes as the objects are not necessarily involved in activities all the time, and not all objects are involved in activities. 
Therefore, we propose to use a background slot that is not supervised by any activity classes. By the softmax operation in slot attention, this background slot would attend regions that are not relevant to any activities and force other action slots to focus on other regions, as shown in the bottom of Figure~\ref{fig:main_arch}.
% be encouraged to attend regions that the action slots should not focus on.
% \paragraph{Background attention guidance.}
To further enhance the effectiveness of the background slot, we provide an attention mask $M_{\mathtt{bg}}$ as the supervision, directly guiding the background slot where to pay attention.
The mask is generated by excluding pixels with the classes of vehicles, pedestrians, drivable areas, crosswalks, and sidewalks.
% pixel-wise instance segmentation collected in the TACO dataset. For the masks in OATS and nuScenes experiments, we use an off-the-shelf DeepLabV3+~\cite{chen2018encoder}.
% With the mask $M_{\mathtt{bg}}$, 
We guide the background attention via the loss function:
$L_{\mathtt{bg}} = \textnormal{BCE}(\bar{A}_{\mathtt{bg}}, M_{\mathtt{bg}})$, where $\bar{A}_{\mathtt{bg}}$ is the normalized attention from the background slot. %\textcolor{red}{YC: we need to highlight how to generate background labels via instance segmentation camera}
% With explicit background attention guidance, the competition mechanism of slot attention would further force the action slots to ignore background regions and thus help them easier to find the regions of interest.  

\vspace{-5mm}
\paragraph{Regularization for action slots.}
\label{subsec:reg}
Instead of providing attention guidance for positive classes, which is non-trivial as explained previously, we introduce a regularization term to discourage action slots associated with negative classes (i.e., those activities not present in a video), from attending to any regions.
% in spatial-temporal features. 
We construct an attention mask $M_{\mathtt{neg}}$ wherein all element values are set to 0.
We design a loss function: $L_{\mathtt{neg}} = \sum_{\{c | y_{c}=0\}} \textnormal{BCE}(\bar{A}_{c}, M_{\mathtt{neg}})$,
where $\bar{A}_{c}$ is the normalized attention output of slot $S_c$ for negative class $c$.
By regularizing the action slots with negative classes, the competition mechanism makes slots with positive classes more competitive in the important regions.
% identify non-relevant regions for negative classes 
Note that the regularization does not require additional annotations.
Finally, the full objective function of Action-slot is: 
\[L_{\mathtt{all}} = L_{\mathtt{act}}+ w_{\mathtt{bg}}L_{\mathtt{bg}} + w_{\mathtt{neg}}L_{\mathtt{neg}}\] 
% \begin{equation}
%    L_{\mathtt{all}} = L_{\mathtt{act}}+ w_{\mathtt{bg}}L_{\mathtt{bg}} + w_{\mathtt{neg}}L_{\mathtt{neg}},
%    \label{loss_all}
% \end{equation}
%
, where $w_{\mathtt{bg}}$ and $w_{\mathtt{neg}}$ are the weights to balance two terms (set as 0.5 and 1, respectively in this paper).

\label{sec:experiments}

\section{Experiments}
\subsection{Datasets}
% We evaluate baselines and Action-Slot on the three datasets:
\noindent \textbf{OATS}~\cite{Agarwal_2023_ICCV}. The dataset was collected in San Francisco with an instrumented vehicle and comprises 1026 labeled clips and is divided into 3 splits for cross-validation. OATS labels 59 traffic atomic activity classes. We follow the same experimental protocol to train and evaluate the 35 classes.
%used in OATS's experiment. 
%in which have enough samples. 
%
The dimensions of an image are 1200 $\times$ 1920 pixels. 
We downsample the image size to 224 $\times$ 224, as in~\cite{Agarwal_2023_ICCV}. 

\noindent \textbf{TACO}. We construct the TACO dataset in the CARLA simulator~\cite{Dosovitskiy17} to attain a balanced distribution encompassing 64 classes of traffic atomic activities. 
We leverage three approaches to collect data: auto-pilot, scenario-runner~\cite{scenario_runner}, and automatic scenario generation~\cite{kung2023riskbench}.
The TACO dataset comprises 5178 clips. Among them, 1148 are used for testing.
% Besides the 4-way intersection, we extend the topology-aware activity description for T-intersections and collect them for diverse road topologies.
We collect image and instance segmentation data with a size of 512 $\times$ 1536 pixels, and subsequently downsample each frame to 256 $\times$ 768 as input to the models.
%which have a size of 512 $\times$ 1536 pixels and we downsample a frame to 256 $\times$ 768 for the models' input. 

\noindent \textbf{nuScenes}~\cite{nuscenes}. We annotate the $train\_val$ set of nuScenes dataset with atomic activities. 
The $train\_val$ set comprises 850 videos.
Our annotation protocol yields a total of 426 short clips, each comprising 16 frames.
%remains 426 short clips with 16 frames after we annotate it. 
%
The annotated nuScenes encompasses 42 classes of atomic activity.
%processes 42 classes of atomic activity.
%Please refer to the supplementary for more details on the TACO dataset construction and the annotation process for nuScenes. 
% \noindent
Please refer to the supplementary material for additional details on the dataset construction and annotation.
 % of the TACO dataset and the annotation process for nuScenes.

\subsection{Baselines}

\noindent \textbf{Video-level models.}
We implement our baselines using the PyTorchVideo~\cite{fan2021pytorchvideo} library to construct different video-level architectures, including I3D~\cite{carreira2017quo}, X3D~\cite{feichtenhofer2020x3d}, CSN~\cite{tran2019video}, SlowFast~\cite{feichtenhofer2019slowfast}, MViT~\cite{fan2021multiscale}, and VideoMAE~\cite{tong2022videomae}.
%
%, pre-trained on Kinetics-400~\cite{kay2017kinetics}. 
%
Note that we modify the models to perform multi-label prediction with one linear layer that outputs $N_{\mathtt{cl}}$ class channels.

\noindent \textbf{Object-aware models.}
% For object-aware approaches~\cite{Baradel_2018_ECCV-,CVPR2019_ARG,Agarwal_2023_ICCV}, we perform multi-class prediction for each object proposal.
%
The models~\cite{Baradel_2018_ECCV,CVPR2019_ARG,Agarwal_2023_ICCV} first extract object features via the RoI-Align~\cite{he2017mask} and learn the relations between objects. ORN~\cite{Baradel_2018_ECCV}, ARG~\cite{CVPR2019_ARG}, and OATS~\cite{Agarwal_2023_ICCV} uses MLP,  Graph Convolutional Network (GCN)~\cite{kipf2016semi}, and spatical-temporal GCN~\cite{mohamed2020social} to enhance object features. OATS also uses an additional graph branch to encode low-level object trajectories.
% To extract trajectories of objects, we use Deep OC-SORT~\cite{maggiolino2023deep}, the state-of-the-art model on the MOT20 benchmark~\cite{dendorfer2020mot20} in terms of HOTA~\cite{luiten2021hota}.
We use Deep OC-SORT~\cite{maggiolino2023deep} to extract trajectories of objects.
We follow OATS by setting the number of tracklets as 20 for the input of the models.
To encourage the object-aware models~\cite{Baradel_2018_ECCV,CVPR2019_ARG} to learn the context information, we concatenate object features with the global features, which are obtained by convolving X3D~\cite{feichtenhofer2020x3d} features with a 3D kernel size of 1.
These models then output multi-class predictions for each proposal by a linear layer with a channel number of $N_{cl}$ plus one negative class.
% The object-aware models perform multi-class predictions for each object proposal with a channel number of $N_{cl}$ plus one negative class. 
%
Note that we do not implement the model proposed in OATS~\cite{Agarwal_2023_ICCV} due to the absence of source code.
% For more details on object-aware methods, please refer to the supplementary material.

%========================================
% \\Supplementary\\
% During training, we pad the ground truth label with the negative class to match the length of tracklet proposals and use a Hungarian matcher to associate them.
% %
% It is worth noting that because object-aware baselines output multi-instance results, i.e., each proposal can be any action class, we rearrange the outputs to a set and calculate the metrics. 
%========================================

\noindent \textbf{Slot-based models.}
% For slot-based models, we use the $5^{\mathtt{th}}$ ResBlock of X3D~\cite{feichtenhofer2020x3d} as the backbone features.
The models take features from the last ConvBlock of the encoders~\cite{he2016deep,feichtenhofer2020x3d} as input.
%the models' input.
Note that we remove the last projection layer of X3D, which makes the size of some slot-based models slightly smaller than X3D.
We re-implement existing slot-attention-based methods for videos~\cite{kipf2022conditional,elsayed2022savi++,bao2022discovering,zhou2022slot} to use the proposed ``allocated slots" for multi-label atomic activity recognition. 
% by modifying their slot designs to the ``allocated ones" for better performance. 
%Since existing slot-attention-based methods for videos~\cite{kipf2022conditional,elsayed2022savi++,bao2022discovering,zhou2022slot} are designed for object tracking and video panoptic segmentation, we re-implement their architectures and modify their slot as our allocated ones for better performance.
%
Specifically, SAVi~\cite{kipf2022conditional} randomly initialize slots for each forward pass from a learned distribution, while MO~\cite{bao2022discovering} and Slot-VPS~\cite{zhou2022slot} treat slots as learnable queries~\cite{carion2020end}. 
For the slot update mechanism, SAVi and MO recurrently update slots in time, 
% (see Figure~\ref{fig:slot-updating}(a))
while Slot-VPS uses an extra self-attention~\cite{vaswani2017attention} to update slots across frames. 
We then use their slots in the last frame as the input to the classifier.

% We also need to specify the slot updating mechanism.
% Specifically, slot-based approaches use $N_{\mathtt{cl}}$ independent linear layers $\in \mathbb{R}^{\mathtt{D_{\mathtt{slots}}} \times 1}$ for $N_{\mathtt{cl}}$ binary classifications, where $D_{\mathtt{slots}}$ is the dimension of a slot.

% Since the slot-based baselines follow the object-centric fashion and they update slot features frame by frame, we aggregate the slot features at different times with sum as input to the classifier.
% For all the models, we also predict the ego vehicle's pattern based on backbone features extracted from the corresponding CNN architectures.
%

\begin{table}[h!]
\centering
\scriptsize
\caption{Quantitative results on the OATS dataset. ``Seq'' denotes the input sequence length. The symbol $\ddag$ denotes the re-implementation of slot-based methods, where each slot is allocated to a specific atomic activity class.
%to achieve reasonable performance 
\textit{S1}, \textit{S2}, and \textit{S3} represents the three defined splits in the OATS dataset.
}

        \resizebox{0.98\linewidth}!{
        \begin{tabular}
            {@{}l@{\;}c @{\;} c @{\;} | @{\;} c @{\;} c @{\;} c @{\;} c @{\;} | c }
            \toprule
            \multicolumn{1}{l}{Method}& 
            \multicolumn{1}{c}{Backbone}  & 
            \multicolumn{1}{l}{Seq}& 
            \multicolumn{1}{c}{Pretrain}  & 
            \multicolumn{1}{c}{S1}  & 
            \multicolumn{1}{c}{S2}  & 
            \multicolumn{1}{c}{S3}  & 
            \multicolumn{1}{c}{mAP}   
            \\ 
             \midrule
            CSN~\cite{tran2019video,Agarwal_2023_ICCV} & ResNet152 &32 & IG65M & 12.1 & 12.6 & 12.9&12.5
            \\
            TPN~\cite{yang2020temporal,Agarwal_2023_ICCV} & ResNet50 &32 & ImageNet & 11.6& 13.3& 12.9 & 12.6
            \\
            SlowOnly~\cite{feichtenhofer2019slowfast,Agarwal_2023_ICCV} & ResNet50 &32 & ImageNet & 11.2 & 14.7& 12.9 & 13.0
            \\
            SlowFast~\cite{feichtenhofer2019slowfast,Agarwal_2023_ICCV} & ResNet50 &32 & None & 10.8 &15.1 & 14.5 & 13.5
            \\
            I3D (NL)~\cite{wang2018non,Agarwal_2023_ICCV}& ResNet50&32 & ImageNet & 11.9 &15.5 & 14.0 & 13.8
            \\
            I3D~\cite{carreira2017quo,Agarwal_2023_ICCV} &  ResNet50 &32& ImageNet & 11.8 &14.3 &16.8 & 14.3
            \\
            \midrule
             ORN~\cite{Baradel_2018_ECCV,Agarwal_2023_ICCV} & ResNet50 &32& ImageNet & 16.8& 13.4& 18.1 & 16.1
             \\
             ARG~\cite{CVPR2019_ARG,Agarwal_2023_ICCV} & Inceptionv3 &32& ImageNet & 20.2 &21.3 &19.3 & 20.3
             \\
            OATS~\cite{Agarwal_2023_ICCV} & Inceptionv3 &32& ImageNet & \uline{24.3}& \uline{28.6} & \uline{27.2} & \uline{26.7}
            
            \\
            \midrule
             SAVi$\ddag$~\cite{kipf2022conditional,elsayed2022savi++}& ResNet50 &32& ImageNet & 21.1 & 22.4 & 22.5 &22.0
              \\
             MO$\ddag$~\cite{bao2022discovering} &ResNet50 &32& ImageNet &14.3 & 15.6 & 18.2 & 16.0
             \\ 
             Slot-VPS$\ddag$~\cite{zhou2022slot} &ResNet50 &32& ImageNet & 15.7 & 17.8 &17.2 &16.9
              \\
             Action-slot (Ours) & ResNet50 &32& ImageNet & \textbf{26.6} & \textbf{28.6} & \textbf{30.8} & \textbf{28.6}
              \\
            \hhline{========}
             % \bottomrule
             I3D~\cite{carreira2017quo} & ResNet50 &8& Kinetics-400 & 21.7 & 24.6 &24.4 & 23.6
            \\
             X3D~\cite{feichtenhofer2020x3d} & N/A &16& Kinetics-400 & 30.4	& 33.2 &30.6 & 31.4
              \\
             CSN~\cite{tran2019video} & ResNet101 &32& Kinetics-400 & \uline{43.1} & \uline{47.1} & \uline{44.3} & \uline{44.8}
            \\
             SlowFast~\cite{feichtenhofer2019slowfast} & ResNet50 &16& Kinetics-400 & 36.1	& 36.6 & 34.2 & 35.6
             % \midrule
             \\
             \midrule
             ORN~\cite{Baradel_2018_ECCV} & X3D &16& Kinetics-400 & 19.3 & 22.5 & 23.6 & 21.8
             \\
             ARG~\cite{CVPR2019_ARG} & X3D &16& Kinetics-400 & 24.8 & 25.9 & 29.3 & 26.7
             \\
            % OATS~\cite{Agarwal_2023_ICCV} & X3D & Kinetics-400 & 
            % \\
             % \midrule
             \midrule
             SAVi$\ddag$~\cite{kipf2022conditional,elsayed2022savi++}& X3D &16& Kinetics-400 & 19.0&22.1& 21.6 &20.9
              \\
             MO$\ddag$~\cite{bao2022discovering} &X3D &16& Kinetics-400 & 25.3 & 25.0 & 24.2& 24.8
             \\
             Slot-VPS$\ddag$~\cite{zhou2022slot} &X3D &16& Kinetics-400 & 24.7	& 24.6 & 25.5 & 24.9
              \\
              Action-slot (Ours) & X3D &16& Kinetics-400 & \textbf{48.1}& \textbf{47.7}&\textbf{48.8}& \textbf{48.2}
              \\
             % \midrule
            \bottomrule
        \end{tabular}
}
%\vspace{8pt}
\label{table:oats}
\vspace{-2mm}
\end{table}

\begin{table}[h!]
\centering
\small
\caption{Quantitative results on the TACO dataset. ``Seq'' denotes the input sequence length. \textit{C}, \textit{K}, \textit{P}, \textit{C+}, \textit{K+}, and \textit{P+} denote activities involved with different types of road users.
The symbol $\ddag$ denotes the re-implementation of slot-based methods, where each slot is allocated to a specific atomic activity class. %the slots are allocated to specific action classes to achieve reasonable performance.
}
\resizebox{1\linewidth}{!}{
        \begin{tabular}
            {@{}l@{\;}c  c  | c @{\;} c@{\;} c @{\;} c @{\;}c @{\;}c @{\;}|c }
            \toprule
            \multicolumn{1}{l}{Method}& 
            \multicolumn{1}{c}{Para. (M)}  & 
            \multicolumn{1}{c}{Seq}  & 
            \multicolumn{1}{c}{C}  & 
            \multicolumn{1}{c}{K}  & 
            \multicolumn{1}{c}{P} &
            \multicolumn{1}{c}{C+} &
            \multicolumn{1}{c}{K+} &
            \multicolumn{1}{c}{P+} &
            \multicolumn{1}{c}{mAP} 
             \\
             \midrule

             I3D~\cite{carreira2017quo} & 27.3 & 8 & 
             27.3 & 19.4 & 30.5 &  34.6&33.6 & 34.8 & 29.7
            \\
             X3D~\cite{feichtenhofer2020x3d} & 3.0 & 16 &
             37.5 & 20.3 & 34.6& \uline{56.3} & 51.5 & 38.8 & 38.3
              \\
             CSN~\cite{tran2019video} &21.4& 32 & 
             \uline{43.5} & \uline{35.5} & \uline{43.0} & 52.5 & \uline{46.1} & \uline{43.4} &\uline{44.0}
            \\
             SlowFast~\cite{feichtenhofer2019slowfast} & 33.7 & 16 & 
             33.9 & 19.7 & 36.7 & 39.9& 42.6 & 41.0 &35.2
                \\
             MViT~\cite{fan2021multiscale} & 36.6 &16 & 
             21.4 & 13.8 & 26.3 & 43.7& 30.0 & 33.8 & 27.9
               \\
             VideoMAE~\cite{fan2021multiscale} &  57.9&16 & 
             30.6 & 18.6 & 27.1 &  51.6& 33.0 &37.4 & 33.1
            \\
             \midrule
             ORN~\cite{Baradel_2018_ECCV} & 4.8 & 16 & 25.5 & 15.8 & 24.6 & 31.6 & 19.8 & 13.9 & 22.2 
             \\
            ARG~\cite{CVPR2019_ARG} &12.2& 16 & 27.8 & 15.0 & 27.2 & 35.6 & 13.7 & 20.0 & 23.2
             % ARG*~\cite{CVPR2019_ARG} &12.2 & 16 & 
             % \\
             \\
             % ORN*~\cite{Baradel_2018_ECCV} & 4.8 & 16 &
             %   \\
            % OATS~\cite{Agarwal_2023_ICCV} & 4.8 & 16 &
             % OATS*~\cite{Agarwal_2023_ICCV} & 4.8 & 16 &
             %   \\
            % GCN~\cite{li2020learning} &&16&&&
            % \\
             \midrule
             SAVi$\ddag$~\cite{kipf2022conditional,elsayed2022savi++}& 2.3 & 16 & 
             19.2 & 13.6 & 27.5 &  23.3& 25.9 & 35.9& 23.3
              \\
             MO$\ddag$~\cite{bao2022discovering} &2.3&16& 
             34.2 & 24.9 & 39.2 & 38.3 & 39.4 & 37.7 & 35.3
             \\
             Slot-VPS$\ddag$~\cite{zhou2022slot} &3.5&16& 
             31.9 & 21.3 & 32.0 & 51.9 & 44.7 & 31.3 & 36.0
              \\
             % \midrule
             % Action-Slot (X3D) & \textbf{2.3} & 16 & \textbf{73.4} & \textbf{66.8} & \textbf{70.8} & \textbf{62.5} & \textbf{66.9} & \textbf{54.4} & \textbf{47.3} &
             %  53.2 & \textbf{56.4} & \textbf{57.9}
             %  \\
              Action-slot (Ours) & \textbf{2.3} & 16 & 
              \textbf{48.1} & \textbf{41.2} & \textbf{49.2} & \textbf{70.1} & \textbf{62.6} & \textbf{52.8} &\textbf{54.4}
              \\
             % \midrule
            \bottomrule
        \end{tabular}
}
%\vspace{8pt}
\label{table:taco}
\vspace{-2mm}
\end{table}

% \subsection{Pretrain and Backbone}

\noindent \textbf{Pretraining and backbone choice.}
% \noindent \textbf{ImageNet}.
We first follow the setting of OATS and implement Action-slot and slot-based baselines with ResNet50~\cite{he2016deep} pretrained on ImageNet~\cite{deng2009imagenet}.
% \noindent \textbf{Kinetics}. 
To further study the effect of pre-trained models, we train models with state-of-the-art 3D bakcbones~\cite{carreira2017quo,feichtenhofer2019slowfast,feichtenhofer2020x3d}, pretrained on Kinetics-400~\cite{kay2017kinetics}, for the experiments on both OATS and TACO datasets. 
% To utilize the advantage of advanced video representations, w
Specifically, we use X3D~\cite{feichtenhofer2020x3d} as the backbone by default because of the compact model size.
% For experiments on TACO, all the models including our Action-Slot are trained for 100 epochs using AdamW optimizer~\cite{loshchilov2018decoupled} with a batch size of 8 and learning rate 5e-4.
Please refer to the supplementary material for the additional implementation details.
% and training hyperparameters of Action-Slot.

\subsection{Implementation Details}
 % \noindent \textbf{Ego-vehicle action}.  While OATS only collects videos where ego-vehicle has a specific action (e.g., Z1-Z3), we collect scenarios where ego-vehicle stops for traffic lights or slowly moves away from \textit{Z1}, denoted as \textit{Z1-Z1: E}. \ychen{the statement is not convincing. why?}

% The length of video sequences can be various, so we randomly sample sub-sequences with equal intervals between frames for the training set and fix the sub-sequences for the testing set~\cite{Agarwal_2023_ICCV}.
\noindent \textbf{Background masks.}
We extract background masks from the instance segmentation collected in TACO. For OATS and nuScenes, we use an off-the-shelf DeepLabV3+~\cite{chen2018encoder}.

\noindent \textbf{Metrics.}
We follow the common practice in multi-label action recognition~\cite{yeung2015every,sigurdsson2016hollywood,gu2018ava,Agarwal_2023_ICCV} to report mean average precision (mAP). 
% We report mean average precision (mAP)
% as it is the common metric used in multi-label action recognition~\cite{yeung2015every,sigurdsson2016hollywood,gu2018ava,Agarwal_2023_ICCV}. 

\noindent \textbf{Visualization.}
We visualize attention maps of each atomic activity according to the allocated action slots to validate if Action-slot learns action-centric representations. 
%This enhances the explainability of the proposed framework. 
We keep pixels with an attention score greater than 0.5 on OATS and 0.2 on TACO.
% We choose the thresholds empirically.
%due to the massive number of classes.

%============================================
% \\Supplementary\\ 
% it is worth mentioning that we eliminate the GRU from the slot updating process due to its negligible impact on enhancing performance.
% %
% To have fair comparisons, we fine-tune the last two 3D Conv blocks of the backbone for all the models. 
% %
% In addition, the positive-negative data imbalance is a well-known issue in multi-label recognition. 
% %
% We set the weight of positive classes as 5 for slot-based and CNN-based methods in binary cross-entropy and 15 for object-aware methods in cross-entropy.
%============================================
%

% We conduct all experiments on an NVIDIA 4090 GPU with 24 GB memory. 

%except MViT
%MViT is trained on 2 V100 GPUs because the increased number of patches introduces significant computation.

% \textcolor{red}{It's worth mentioning that we eliminate the GRU from the updating process due to its negligible impact on enhancing performance. This result is contrary to the original slot attention~\cite{locatello2020object} findings. Instead, we rely on the updated values $U$ as the final state of slots $S^{'}$.}

\begin{table*}[t!]
\begin{minipage}{.4\linewidth}
% \centering
% \scriptsize
% \caption{Results of the different backbone of Action-slot on the TACO dataset.}
% \begin{tabular}
%             {@{}l@{\;} @{\;} c @{\;}@{\;}}
%             \toprule
%             \multirow{1}{*}{ \begin{tabular}{@{\;}c@{\;}} \end{tabular}} & 
%              \multicolumn{1}{c}{mAP}  

%              \\
%              \midrule
% \
%              I3D & 29.7
%              \\
%              Action-slot &37.6 (+7.9)
%              \\
%             \midrule
%              X3D~\cite{feichtenhofer2020x3d}&  37.8 \\
%              Action-slot & \textbf{54.4} (+16.6)
%               \\
%              \midrule
%              SlowFast~\cite{feichtenhofer2019slowfast}& 35.2
%              \\
%              Action-slot & 46.7 (+11.5)
%               \\
             
%             \bottomrule
%         \end{tabular}
%         \label{table:backbone}
\centering
\scriptsize
\caption{Ablation study of Action-slot. The reported results are the average of the 3 splits
% defined 
in the OATS dataset.}
\vspace{-2mm}
% \resizebox{0.65\columnwidth}{!}{
\begin{tabular}
            {@{\;} @{\;}c @{\;}|@{\;}c @{\;}|@{\;}c @{\;}|@{\;}c @{\;}|@{\;}c @{\;}|@{\;}c @{\;} | @{\;}c@{\;} }
            \toprule
            \multicolumn{1}{c}{ID}  &
            \multicolumn{1}{c}{Allocated}  &
            \multicolumn{1}{c}{Update}  &
            \multicolumn{1}{c}{BG Slot}  &
            \multicolumn{1}{c}{$L_{\mathtt{bg}}$}  &
            \multicolumn{1}{c}{$L_{\mathtt{neg}}$}  &
            \multicolumn{1}{c}{mAP}
            \\
             \midrule
              1&&parallel&&&&   10.8
              \\
               2&\checkmark&recurrent&&&&  24.8
               \\
              3&\checkmark&parallel&&&&  42.7
              \\
               4&\checkmark&parallel&\checkmark&&&  40.8
              \\
              5&\checkmark&parallel&\checkmark&\checkmark& &  43.6
              \\
              6&\checkmark&parallel&\checkmark& & \checkmark & 43.0
              \\
              7&\checkmark&parallel&\checkmark&\checkmark&\checkmark &  \textbf{48.2}
             \\
             % \midrule
             % \midrule
            \bottomrule
        \end{tabular}
\label{tab:ablation}
\end{minipage}%
\hspace{0.03\linewidth}
\begin{minipage}{.3\linewidth}
\centering
\scriptsize
\caption{Comparisons of Action-slot and object-level guidance on TACO across different numbers of road users (denoted as $N$) present in a video. BG and Neg denote background slot and regularization in our method.}
\vspace{-2mm}
\begin{tabular}
            {@{}l@{\;} @{\;} c @{\;}@{\;} @{\;}@{\;}c @{\;}@{\;} @{\;} c @{\;} @{\;} c @{\;}  }
            \toprule
            \multirow{1}{*}{ \begin{tabular}{@{\;}c@{\;}} \end{tabular}} & 
             \multirow{1}*{ \begin{tabular}{@{}c@{}} $N$ $\leq$ 5 \end{tabular}} & 
             \multirow{1}*{ \begin{tabular}{@{}c@{}}  5 $<$ $N$ $\leq$ 15 \end{tabular}} & 
             \multirow{1}*{ \begin{tabular}{@{}c@{}} $N$ $>$ 15 \end{tabular}} 
             &
             
             \\
             \midrule
             object & 49.4 & 47.7 & 43.5
             \\
             BG+Neg & 55.2 & 50.9 & 46.3
             \\
            \bottomrule
        \end{tabular}
        
        \label{table:object_guidance}
\end{minipage}
\hspace{0.03\linewidth}
\begin{minipage}{.2\linewidth}
% \centering
% \scriptsize
% \caption{Quantitative results of transfer learning to OATS and nuScenes. FS denotes training from scratch on the target.}
% \begin{tabular}
%             {@{}l@{\;}  c @{\;}c   c @{\;} c  }
%             \toprule
%             \multirow{2}{*}{ \begin{tabular}{@{\;}c@{\;}} \\\end{tabular}} & 

%             \multicolumn{2}{c}{OATS}  & 
%              \multicolumn{2}{c}{nuScenes}  

%              \\
%              \cmidrule(lr){2-3} \cmidrule(lr){4-5}
%             &
%              \begin{tabular}{c@{\;}}  FS  \end{tabular} & 
%              \begin{tabular}{c}  +TACO   \end{tabular} & 
%             \begin{tabular}{c@{\;}} FS   \end{tabular} & 
%              \begin{tabular}{c@{\;}} +TACO  \end{tabular} 
            
%               \\
%             \midrule
%             X3D~\cite{feichtenhofer2020x3d} & 31.4& 34.9 (+3.5) & 19.8 & 27.8 (+8.0)
%             \\
%             ARG & 26.7 & 31.1 (+4.4) & 12.2 & 17.0 (+4.8)
%             \\
%             % Slot-VPS & 25.5 &  & 20.7 &
%             %  \\
%              Action-slot & 48.2 & \textbf{59.5} (+11.3) & 23.6 & \textbf{32.3} (+8.7)
%              \\
%             \bottomrule
%         \end{tabular}
        
%         \label{table:transfer}

\centering
\scriptsize
\caption{Results of various backbones for Action-slot on TACO.}
\vspace{-2mm}
\begin{tabular}
            {@{}l@{\;} @{\;} c @{\;}@{\;}}
            \toprule
            \multirow{1}{*}{ \begin{tabular}{@{\;}c@{\;}} \end{tabular}} & 
             \multicolumn{1}{c}{mAP}

             \\
             \midrule
\
             I3D & 29.7
             \\
             Action-slot &37.6 (+7.9)
             \\
            \midrule
             X3D~\cite{feichtenhofer2020x3d}&  37.8 \\
             Action-slot & \textbf{54.4} (+16.6)
              \\
             \midrule
             SlowFast~\cite{feichtenhofer2019slowfast}& 35.2
             \\
             Action-slot & 46.7 (+11.5)
              \\
             
            \bottomrule
        \end{tabular}
        \label{table:backbone}
\end{minipage}
\vspace{-3mm}
\end{table*}

% \begin{table}[!t]
% \centering
% \scriptsize
% \caption{Ablation study of Action-slot. The reported results are the average of the 3 splits of the OATS dataset.}
% % \resizebox{0.65\columnwidth}{!}{
% \begin{tabular}
%             {c @{\;}|c @{\;}|c @{\;}|c @{\;} | c@{\;} c }
%             \toprule
%             \multicolumn{1}{c}{Allocated Slot}  &
%             \multicolumn{1}{c}{BG Slot}  &
%             \multicolumn{1}{c}{$L_{\mathtt{bg}}$}  &
%             \multicolumn{1}{c}{$L_{\mathtt{neg}}$}  &
%             \multicolumn{1}{c}{mAP}
%             \\
%              \midrule
%               &&&&   10.8
%              \\
%               \checkmark& &&&  42.7
%               \\
%                \checkmark&\checkmark&&&  40.8
%               \\
%               \checkmark&\checkmark&\checkmark& &  43.6
%               \\
%               \checkmark&\checkmark& & \checkmark & 43.0
%               \\
%               \checkmark&\checkmark&\checkmark&\checkmark &  \textbf{48.2}
%              \\
%              % \midrule
%              \midrule
%             \bottomrule
%         \end{tabular}
% \label{tab:slot}
% \end{table}

\subsection{Main Results}
\label{sec:main}
\noindent \textbf{OATS.}
In Table~\ref{table:oats}, we show experimental comparisons with video-level-based, object-aware, and slot-attention-based methods.
In the top group, we use the ImageNet pre-trained ResNet50, following the evaluation protocol designed in OATS~\cite{Agarwal_2023_ICCV}. 
%
% The corresponding results are reported .
%
% Action-slot achieves state-of-the-art performance on all three splits.
%
%We found the object-centric slot-based methods can slightly improve the performance over the video-level models (e.g., I3D and SlowFast). 
%
We further compare all methods with advanced visual backbones 
%by leveraging X3D~\cite{feichtenhofer2020x3d} 
pretrained from Kinetics-400~\cite{kay2017kinetics} and report the corresponding results in the bottom group. 
%
%All models gain significant improvement for a more fair comparison. 
%
In both ImageNet and Kinetics pretrained settings, Action-slot achieves state-of-the-art performance on the OATS dataset.
Specifically, Action-slot improves the performance of atomic activity recognition by identifying the region of interest via the proposed slot attention mechanism, without explicit perception guidance.
%
%which is more robust and can avoid compounding errors~\cite{carion2020end}.

\noindent \textbf{TACO.}
We establish a new benchmark on the proposed TACO dataset to validate the effectiveness of Action-slot for a wide range of atomic activities, which is challenging for the OATS dataset~\cite{Agarwal_2023_ICCV} due to its imbalanced distribution. 
%
%Additionally, the TACO dataset enables the evaluation of AP of atomic activities involved with different types of road users because of its large-scale and balanced category distribution. 
%
In the top group of Table~\ref{table:taco}, we evaluate two recent transformer-based models, i.e., MViT~\cite{fan2021multiscale} and VideoMAE~\cite{tong2022videomae}. 
The models show worse performance compared to other video-level methods, which could be attributed to their data-hungry nature~\cite{dosovitskiy2021an}.
%
% We hypothesize that this is because the transformers are more data-hungry as demonstrated in the previous literature ~\cite{dosovitskiy2021an}. 
%

Object-aware models~\cite{Baradel_2018_ECCV,CVPR2019_ARG,Agarwal_2023_ICCV} show suboptimal performance, particularly for atomic activities related to grouped road users, as reported in the middle group.
% The performance of object-aware models~\cite{Baradel_2018_ECCV,CVPR2019_ARG,Agarwal_2023_ICCV} is reported in the middle group. 
% %
% We note that the models exhibit unsatisfactory results, especially for the atomic activities involved with grouped road users. 
%
The findings indicate that the relation modeling of objects learned from the MLP and GCN~\cite{kipf2016semi} is not effective for the task because it requires a holistic understanding of contextual information and motions of objects.
%
%Moreover, the results highlight the need 
% The results suggest the relationship of objects learned from the GCN~\cite{kipf2016semi} is not effective in traffic atomic activity recognition. 

In the bottom group of Table~\ref{table:taco}, 
% we evaluate object-centric slot-based models and the proposed Action-slot. 
we re-implement the object-centric slot-based models with allocated slots (denoted by $\ddag$).
% Slot-VPS performs slightly better than the recurrent slot updating methods (i.e., SAVi and MO) because it leverages self-attention for temporal consistency. Finally, 
Action-slot outperforms all models by a large margin on activities with all types of road users because of our parallel architecture design and the proposed attention guidance. Particularly, Action-slot excels in activities involved with grouped road users.
% In comparison, the proposed Action-Slot achieves an 18.0\% mAP gain over Slot-VPS. 
It is worth noting that the Action-slot is powerful yet efficient in terms of model size. The results validate the effectiveness of action-centric representations over the existing action recognition models and object-centric representations.
%in traffic atomic activity recognition.

\subsection{Ablation Study}
\label{sec:ablation}
We conduct ablation studies in Table~\ref{tab:ablation} to understand the effect of our designed components.
% on the three splits in OATS.
% We conduct ablation study on the proposed parallel-updating architecture in Table \ref{sec:ablation}.

% \begin{table}[!t]
% \centering
% \scriptsize
% \caption{Ablation study of our slot attention designs. Results are evaluated on the TACO dataset. \ychen{can we show the value of $L_{bg}$ and $L_{neg}$ yet?}}
% % \resizebox{0.65\columnwidth}{!}{
% \begin{tabular}
%             {c @{\;}|c @{\;}|c @{\;}|c @{\;}|c@{\;}c}
%             \toprule
%             \multicolumn{1}{c}{Allocated Slot}  &
%             \multicolumn{1}{c}{BG Slot}  &
%             \multicolumn{1}{c}{$L_{\mathtt{bg}}$}  &
%             \multicolumn{1}{c}{$L_{\mathtt{neg}}$}  &
%             \multicolumn{1}{c}{mAP}
%             \\
%              \midrule

%               &&&& 38.4
%              \\
%               \checkmark& &&& 52.8
%               \\
%                \checkmark&\checkmark&&& 48.9
%               \\
%               \checkmark&\checkmark&\checkmark& & 53.3
%               \\
%               \checkmark&\checkmark& & \checkmark & 51.4
%               \\
%               \checkmark&\checkmark&\checkmark&\checkmark & \textbf{53.9}
%              \\
%              % \midrule
%              \midrule
%             \bottomrule

%         \end{tabular}
% % }
% \label{tab:slot}
% \end{table}

\noindent \textbf{Non-Allocated vs. Allocated.}
We show that adding allocated slots significantly improves the performance of \textbf{Non-Allocated} by 31.9\% (ID 1 vs. 3). The results suggest the need to revisit the design of slot attention for our task.
% atomic activity recognition. 

\noindent \textbf{Recurrent vs. Parallel.} We demonstrate that updating action slots in a parallel fashion obtains a 17.9\% performance gain (ID 2 vs. 3), compared to conventional methods using a recurrent manner.

\noindent \textbf{Background slot and attention guidance $L_{\mathtt{bg}}$}.
We find that simply adding a non-allocated background slot without mask supervision slightly harms the performance (ID 3 vs. 4), which is different from the observation in previous slot attention literature~\cite{bao2022discovering,zhou2022slot}. 
Our insight is that the non-allocated slot may distract the action slots' attention.
With the background mask supervision $L_{\mathtt{bg}}$, the model shows additional performance gain compared to the one only using the allocated slot (ID 3 vs. 5).

\noindent \textbf{Action slots regularization $L_{\mathtt{neg}}$.}
We regularize action slots with the $L_{\mathtt{neg}}$ term to discourage action slots allocated to negative classes from attending any regions in spatial-temporal features.
%
% i.e., discourage slots with negative classes (i.e., not present in the video) from paying attention to any region of features. 
%In this way, other action slots associated with positive classes would have a better chance to identify the region of interest.
%
%We show that adding the regularization term $L_{\mathtt{neg}}$ can further improve the performance of the Allocated slot.
This design enhances the likelihood of other action slots allocated to positive classes identifying the region of interest more effectively. 
We demonstrate that the inclusion of the regularization term $L_{\mathtt{neg}}$ further improves the one only using the allocated slot (ID 3 vs. 6).

The final model (ID 7 in Table~\ref{tab:ablation}) that incorporates allocated slots, parallel updating, background slot with attention guidance, and action slot regularization, performs the best on OATS, which validates the efficacy of Action-slot.

%achieves competitive performance. 
%
% Combining with background mask supervision $L_{bg}$, Action-slot outperform the one without any guidance by 7.4\%.

\begin{table*}[!t]
% \small
\begin{center}
\begin{tabular}{cc@{\;}c@{\;}c@{\;}c@{\;}c@{\;}c}
\vspace{.2cm}
% % ------
% \hspace{-3mm} 
\scriptsize{Action-slot}
\hspace{-4mm}
& \adjustimage{height=1.9cm,valign=m}{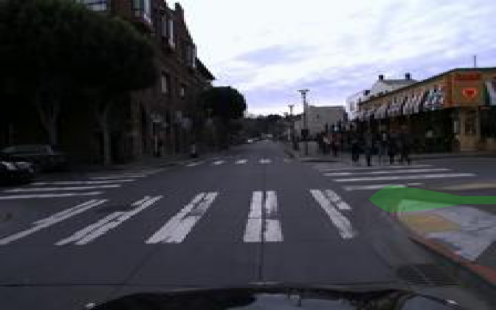}
& \adjustimage{height=1.9cm,valign=m}{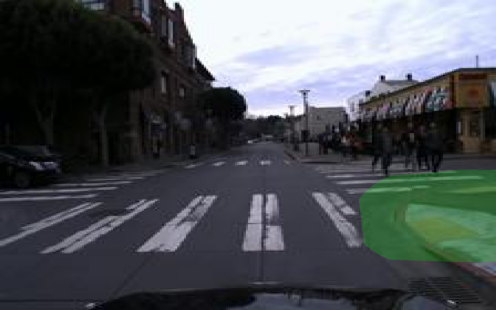}
& \adjustimage{height=1.9cm,valign=m}{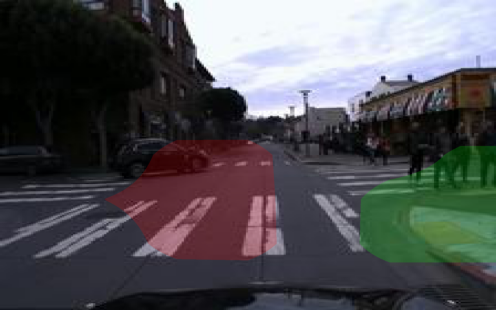}
& \adjustimage{height=1.9cm,valign=m}{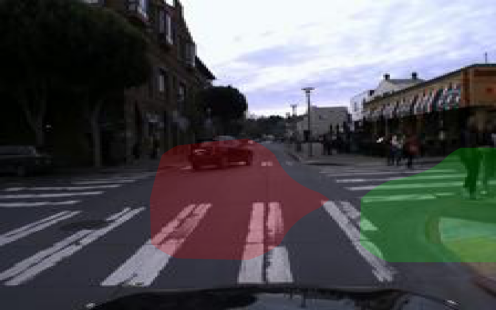}
& \adjustimage{height=1.9cm,valign=m}{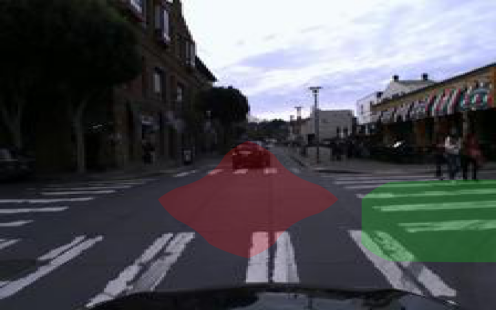}
\hspace{.04cm}
\\
\vspace{-.2cm}

% ------
\hspace{-3mm} 
\scriptsize{MO$\ddag$~\cite{bao2022discovering}}
\hspace{-8mm}
& \adjustimage{height=1.9cm,valign=m}{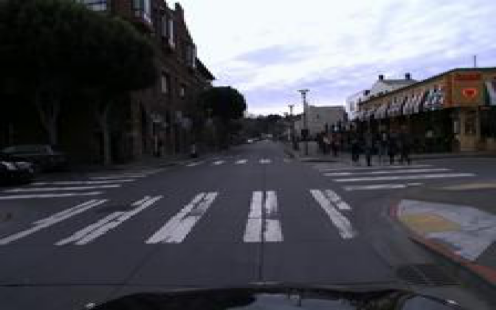}
& \adjustimage{height=1.9cm,valign=m}{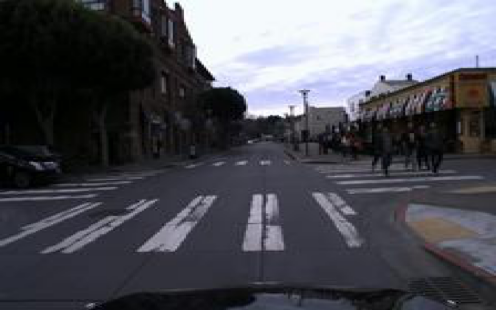}
& \adjustimage{height=1.9cm,valign=m}{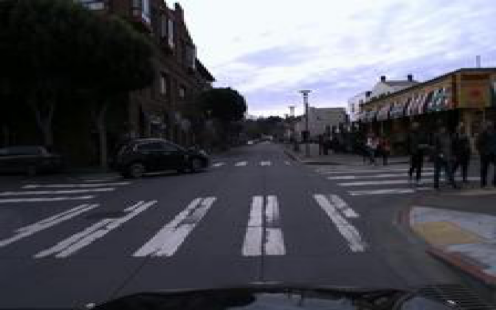}
& \adjustimage{height=1.9cm,valign=m}{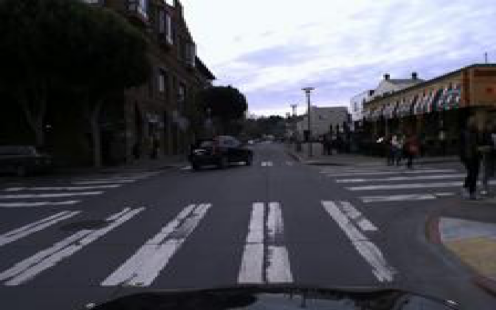}
& \adjustimage{height=1.9cm,valign=m}{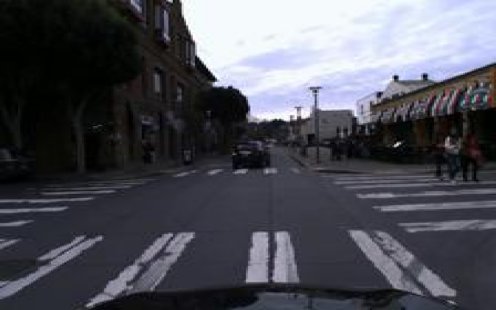}
\hspace{.04cm}

\end{tabular}
\end{center}
\vspace{-10pt}
\captionof{figure}{
Visualization of attention maps learned from OATS. Colored masks represent the attention of the activity slots \textcolor{red}{Z4-Z3:C} and \textcolor{green}{C2-C1:P+}. 
Note that, while MO successfully predicts the activity \textcolor{green}{C2-C1:P+}, the corresponding attention scores are very low.}
\label{fig:attention_oats}
\end{table*}

\begin{table*}[!t]
% \small
\begin{center}
\begin{tabular}{cc@{\;}c@{\;}c@{\;}c@{\;}c@{\;}c}

\vspace{.2cm}
% % ------
% \hspace{-3mm} 
\scriptsize{Action-slot}
\hspace{-4mm}
& \adjustimage{height=1.03cm,valign=m}{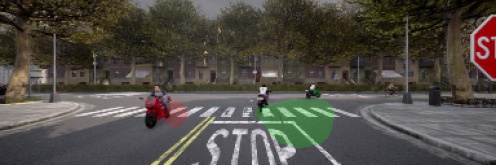}
& \adjustimage{height=1.03cm,valign=m}{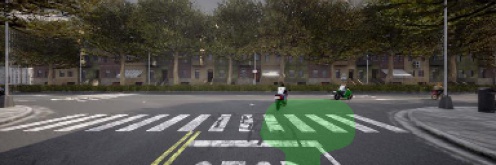}
& \adjustimage{height=1.03cm,valign=m}{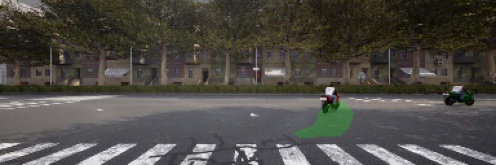}
& \adjustimage{height=1.03cm,valign=m}{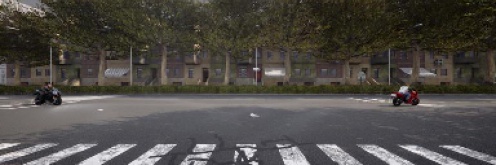}
& \adjustimage{height=1.03cm,valign=m}{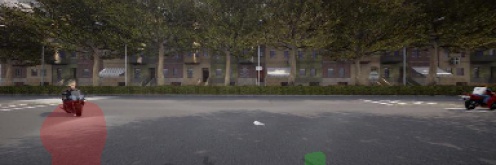}
\\
\hspace{.04cm}

\vspace{-.2cm}

% ------
\hspace{-3mm} 
\scriptsize{MO$\ddag$~\cite{bao2022discovering}}
\hspace{-8mm}
& \adjustimage{height=1.03cm,valign=m}{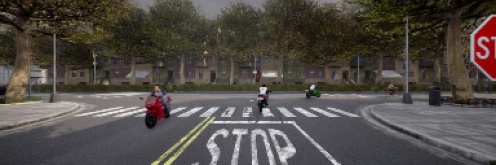}
& \adjustimage{height=1.03cm,valign=m}{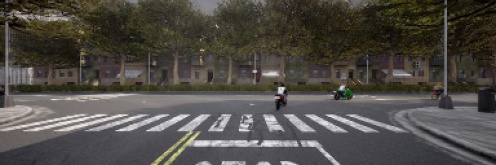}
& \adjustimage{height=1.03cm,valign=m}{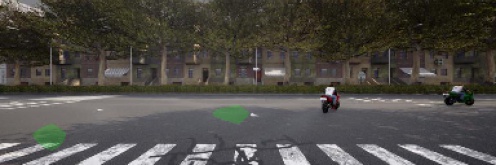}
& \adjustimage{height=1.03cm,valign=m}{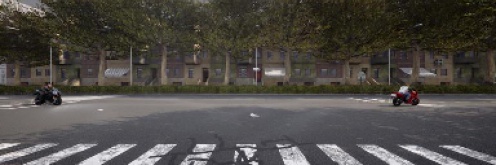}
& \adjustimage{height=1.03cm,valign=m}{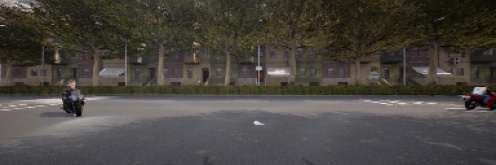}

\end{tabular}
\end{center}
\vspace{-10pt}
\captionof{figure}{
Visualization of attention maps learned from TACO. Colored masks are from the allocated slots of activity \textcolor{red}{Z4-Z1:K} and \textcolor{green}{Z1-Z2:K+}. Note that \textcolor{red}{Z4-Z1:K} appears twice in the 1st and 5th frames independently. 
% We set the visualized attention threshold to 0.2 due to a large amount of classes. 
% Note that MO predicts the activity \textcolor{green}{Z2-Z2:K+} correctly.
}
\label{fig:attention_taco}
\vspace{-2mm}
\end{table*}

\begin{table}[!t]

\centering
\scriptsize
\caption{Effectiveness of using our TACO dataset for representation pretraining and then finetuning on real-world OATS and nuScenes datasets. Kinetics denotes finetuning from the Kinetics-400 pretrained weights.}
\vspace{-2mm}
\resizebox{0.9\linewidth}{!}{
\begin{tabular}
            {@{}l  c c   c  c  }
            \toprule
            \multirow{2}{*}{ \begin{tabular}{@{\;}c@{\;}} \\\end{tabular}} & 

            \multicolumn{2}{c}{OATS}  & 
             \multicolumn{2}{c}{nuScenes}  

             \\
             \cmidrule(lr){2-3} \cmidrule(lr){4-5}
            &
             \begin{tabular}{c@{\;}}  Kinetics  \end{tabular} & 
             \begin{tabular}{c}  +TACO   \end{tabular} & 
            \begin{tabular}{c@{\;}} Kinetics   \end{tabular} & 
             \begin{tabular}{c@{\;}} +TACO  \end{tabular} 
            
              \\
            \midrule
            X3D~\cite{feichtenhofer2020x3d} & 31.4& 34.9 (+3.5) & 19.8 & 27.8 (+8.0)
            \\
            ARG~\cite{CVPR2019_ARG} & 26.7 & 31.3 (+4.6) & 12.2 & 17.0 (+4.8)
            \\
            % Slot-VPS & 25.5 &  & 20.7 &
            %  \\
             Action-slot & 48.2 & \textbf{59.5} (+11.3) & 23.6 & \textbf{32.3} (+8.7)
             \\
            \bottomrule
        \end{tabular}
        
        \label{table:transfer}
        }
        \vspace{-3mm}
\end{table}

% \begin{table}[!t]
% \centering
% \scriptsize
% \caption{Results of the different backbone of Action-slot on the TACO dataset.}
% \begin{tabular}
%             {@{}l@{\;} @{\;} c @{\;}@{\;}}
%             \toprule
%             \multirow{1}{*}{ \begin{tabular}{@{\;}c@{\;}} \end{tabular}} & 
%              \multicolumn{1}{c}{mAP}  

%              \\
%              \midrule
% \
%              I3D & 29.7
%              \\
%              Action-slot &37.6 (+7.9)
%              \\
%             \midrule
%              X3D~\cite{feichtenhofer2020x3d}&  37.8 \\
%              Action-slot & \textbf{54.4} (+16.6)
%               \\
%              \midrule
%              SlowFast~\cite{feichtenhofer2019slowfast}& 35.2
%              \\
%              Action-slot & 46.7 (+11.5)
%               \\
             
%             \bottomrule
%         \end{tabular}
%         \label{table:backbone}
% \end{table}

\label{sec:discussion}
\section{More Analysis and Discussions}

\noindent \textbf{Action-slot vs. Object-level guidance.}
We conduct experiments on the TACO dataset to evaluate the 
% robustness
impact of different numbers of road users in videos with respect to different attention guidance mechanisms in our Actio-slot framework.
%across different numbers of road users in videos.
%, using the TACO dataset.
%
Specifically, we compare the two attention guidance mechanisms discussed in Section~\ref{sec:action_slot} and
%background mask and negative class mask 
%with 
the object-level guidance.
% explored in~\cite{kipf2022conditional,elsayed2022savi++,bao2022discovering}.} 
%
For object-level guidance, we use instance segmentation collected in TACO to obtain masks for vehicles and pedestrians.
Hungarian matcher is used to assign object masks to slots, following MO~\cite{bao2022discovering}.
%We apply Hungarian matching for allocated slots and objects.
%
%Table~\ref{table:object_guidance} compares these two types of attention guidance under various traffic density conditions. 
%
% Specifically, we measure mAP in splits where different numbers of objects $N$ appear in the scenarios.
%
Table~\ref{table:object_guidance} shows that Action-slot with strong ground truth object-level supervisions performs worse, compared to the proposed attention guidance.
% , i.e., $L_{bg}$ and $L_{neg}$.
%
The observation indicates that only considering object cues may not effectively tackle our task because not every object is always involved in the same atomic activity.
% or 2) multiple objects can be involved in the same atomic activity.
%object mask and allocated slots, the performance is slightly worse than our approach
%
%We find the performance of the objects-guided Action-slot becomes worse when the number of objects in a scene increases. On the other hand, Action-slot with the proposed supervision, i.e., $L_{bg}$ and $L_{neg}$ shows superior robustness in scenarios with high traffic density because the action-centric representations focus on activities instead of objects.

\vspace{1mm}
\noindent \textbf{Action-slot with different backbones.}
%We evaluate Action-slot with various backbones on TACO Table~\ref{table:backbone}. 
Table~\ref{table:backbone} reports the performance of Action-slot using different backbone.
 % encoders.
%
We choose I3D~\cite{carreira2017quo}, X3D~\cite{feichtenhofer2020x3d}, and SlowFast~\cite{feichtenhofer2019slowfast}  
%, and \ychen{VideoMAE~\cite{tong2022videomae}} 
for experiments.
Action-slot significantly improves all 3D CNN methods, showing great generalization ability.
 % of the proposed method.
%
Among them, X3D achieves the most improvement.
% \textcolor{red}{In addition, we also trained Action-Slot with CSN as the backbone, but observe that the performance slightly degrades.}
We hypothesize that this is because X3D's features retain the original temporal dimension, affording Action-slot greater spatial-temporal information to analyze atomic activities.
% the features of X3D remain in the original temporal dimension, providing Action-slot more spatial-temporal information to explore atomic activities. 
%
In contrast, other backbone encoders downsample the temporal dimension; for instance, SlowFast takes a video sequence of length 16 as input and reduces the temporal dimension to 4.
%
%Note that we do not include all 3D CNN methods because the temporal dimension of feature maps is downsampled significantly by some of them, e.g., 1/8 in CSN. This can hinder the action slots in finding relevant regions and background mask guidance.

\vspace{1mm}
\noindent \textbf{Representation pretraining on TACO.}
We demonstrate that the proposed TACO dataset can enhance the efficacy of atomic activity recognition on real-world datasets through representation pretraining.
%
%To verify the real-world application value of the proposed TACO dataset, 
%
Specifically, we first pretrain the model on TACO and then fine-tune it on real-world datasets, including OATS~\cite{Agarwal_2023_ICCV} and a newly annotated nuScenes dataset~\cite{nuscenes}. 
We report the average mAP of the three splits in the OATS dataset in Table~\ref{table:transfer}. 
Due to the large-scale and balanced distribution of TACO, all models show a significant improvement on both real-world datasets when pretrained on TACO.
Notably, Action-slot outperforms all baselines, showcasing its robustness and generalizability.
% 
% It is worth noting that Action-slot outperforms all baselines, demonstrating its robustness and generalizability.
%, demonstrating the \textcolor{red}{xxx}. 
%
% Moreover, Action-slot demonstrates its strong generalization and achieves state-of-the-art results in the setting. 
%
% Please see the supplementary for more details.

\vspace{1mm}
\noindent \textbf{Qualitative Results.}
We visualize the attention maps learned using the re-implemented MO~\cite{bao2022discovering} and Action-slot on the OATS and TACO datasets in Figure~\ref{fig:attention_oats} and Figure~\ref{fig:attention_taco}, respectively.
%
% We visualize the learned attention maps from OATS and TACO datasets in Figure \ref{fig:attention_oats} and Figure \ref{fig:attention_taco}, respectively. 
%
% We compare the existing recurrent slot updating strategy~\cite {bao2022discovering} and Action-slot. 
%
Action-slot demonstrates the capability to 
% identify 
decompose multiple atomic activities from intricate scenes in both datasets. 
In Figure~\ref{fig:attention_oats}, Action-slot localizes the group activity \textbf{C2-C1:P+}, while MO~\cite{bao2022discovering} fails to attend any relevant regions. 
Moreover, Action-slot recognizes two independent \textbf{Z4-Z1:K} activities, instead of predicting them as a group activity, i.e., \textbf{Z4-Z1:K+} in Figure~\ref{fig:attention_taco}.
The evidence justifies the capability of Action-slot that learns action-centric representations. 
More interestingly, 
%we found Action-slot is able to localize when an activity happens. 
%
%Specifically, 
Action-slot
% begins attending 
attends to relevant regions when road users initiate an action and cease attention once the action is completed, as seen in the case of the bicyclists in the fourth frame of Figure~\ref{fig:attention_taco}.
% only starts to attend relevant regions when road users start performing an action and stops to attend when the action is accomplished. 
%
% For example, the bicyclists in the fourth frame.
\label{sec:conclusion}
\section{Conclusions}

We present Action-slot, a slot attention-based approach that learns visual action-centric representations, capturing both motion and contextual information.
% , without explicit perception guidance.
% we present Action-slot that learns action-centric representation to improve atomic activity recognition.
We introduce several crucial designs in slot attentions that enable decomposing multiple atomic activities in videos, without the need for explicit perception guidance. 
%
%and suggest modifications for multi-label atomic activity recognition. 
%
We conduct comprehensive experiments to validate the proposed method on both OATS and our TACO dataset, comparing it against various action recognition baseline methods.
% Extensive experiments are provided to validate the proposed method on OATS and our TACO dataset against several action recognition baseline methods.
%
In addition, we demonstrate that Action-slot learns meaningful attention maps, identifying objects in actions without explicit object-level guidance.
Lastly, we show that the TACO dataset enhances the efficacy of multi-label atomic activity recognition on real-world datasets through representation pretraining.
% Moreover, we showcase that Action-slot can learn reasonable attention maps, where objects perform actions in the traffic scene are identified, without explicit object-level guidance or supervision. 
% %
% Last but not least, we conduct transfer learning to the nuScenes dataset showing the real-world value of the proposed TACO dataset.

\noindent \textbf{Limitation.}
Although Action-slot demonstrates favorable quantitative and qualitative performance, we observe that it is still challenging to handle cases when two activities occlude with each other.
In these cases, the corresponding action slots may confuse where they should attend.
This observation is closely relevant to the tracking task under occlusion, where frequent ID switches are often observed.
We hope our findings can encourage the community to explore more advanced designs for action-centric representations.

\noindent \textbf{Acknowledgement.}
The work is sponsored in part by the Higher Education Sprout Project of the National Yang Ming Chiao Tung University and Ministry of Education (MOE), the Yushan Fellow Program Administrative Support Grant, and the National Science and Technology Council (NSTC) under grants 110-2222-E-A49-001-MY3, 111-2634-F-002-022-, 113-2923-E-A49 -003 -MY2, Mobile Drive Technology Co., Ltd (MobileDrive), and Industrial Technology Research Institute Mechanical and Mechatronics Systems Lab.
% \clearpage

{
    \small
    \bibliographystyle{ieeenat_fullname}
    \bibliography{main}

\begin{thebibliography}{82}
\providecommand{\natexlab}[1]{#1}
\providecommand{\url}[1]{\texttt{#1}}
\expandafter\ifx\csname urlstyle\endcsname\relax
  \providecommand{\doi}[1]{doi: #1}\else
  \providecommand{\doi}{doi: \begingroup \urlstyle{rm}\Url}\fi

\bibitem[car(2022)]{carlachallenge2022}
{CARLA Autonomous Driving Challenge}.
\newblock \url{https://carlachallenge.org/}, 2022.

\bibitem[sce(2023)]{scenario_runner}
{ScenarioRunner for CARLA}.
\newblock \url{https://github.com/carla-simulator/scenario_runner}, 2023.

\bibitem[Agarwal and Chen(2023)]{Agarwal_2023_ICCV}
Nakul Agarwal and Yi-Ting Chen.
\newblock Ordered atomic activity for fine-grained interactive traffic scenario understanding.
\newblock In \emph{Proceedings of the IEEE/CVF International Conference on Computer Vision (ICCV)}, pages 8624--8636, 2023.

\bibitem[Arnab et~al.(2021)Arnab, Dehghani, Heigold, Sun, Lu{\v{c}}i{\'c}, and Schmid]{arnab2021vivit}
Anurag Arnab, Mostafa Dehghani, Georg Heigold, Chen Sun, Mario Lu{\v{c}}i{\'c}, and Cordelia Schmid.
\newblock {ViViT: A Video Vision Transformer}.
\newblock In \emph{Proceedings of the IEEE/CVF International Conference on Computer Vision (ICCV)}, pages 6836--6846, 2021.

\bibitem[Bao et~al.(2022)Bao, Tokmakov, Jabri, Wang, Gaidon, and Hebert]{bao2022discovering}
Zhipeng Bao, Pavel Tokmakov, Allan Jabri, Yu-Xiong Wang, Adrien Gaidon, and Martial Hebert.
\newblock {Discovering Objects that Can Move}.
\newblock In \emph{Proceedings of the IEEE/CVF Conference on Computer Vision and Pattern Recognition}, pages 11789--11798, 2022.

\bibitem[Baradel et~al.(2018)Baradel, Neverova, Wolf, Mille, and Mori]{Baradel_2018_ECCV}
Fabien Baradel, Natalia Neverova, Christian Wolf, Julien Mille, and Greg Mori.
\newblock {Object Level Visual Reasoning in Videos}.
\newblock In \emph{Proceedings of the European Conference on Computer Vision (ECCV)}, 2018.

\bibitem[Bertasius et~al.(2021)Bertasius, Wang, and Torresani]{bertasius2021space}
Gedas Bertasius, Heng Wang, and Lorenzo Torresani.
\newblock {Is space-time attention all you need for video understanding?}
\newblock In \emph{International Conference on Machine Learning (ICML)}, 2021.

\bibitem[Biza et~al.(2023)Biza, van Steenkiste, Sajjadi, Elsayed, Mahendran, and Kipf]{biza2023invariant}
Ondrej Biza, Sjoerd van Steenkiste, Mehdi~SM Sajjadi, Gamaleldin~F Elsayed, Aravindh Mahendran, and Thomas Kipf.
\newblock {Invariant Slot Attention: Object Discovery with Slot-Centric Reference Frames}.
\newblock \emph{International Conference on Machine Learning (ICML)}, 2023.

\bibitem[Caesar et~al.(2020)Caesar, Bankiti, Lang, Vora, Liong, Xu, Krishnan, Pan, Baldan, and Beijbom]{nuscenes}
Holger Caesar, Varun Bankiti, Alex~H Lang, Sourabh Vora, Venice~Erin Liong, Qiang Xu, Anush Krishnan, Yu Pan, Giancarlo Baldan, and Oscar Beijbom.
\newblock {nuscenes: A Multimodal Dataset for Autonomous Driving}.
\newblock In \emph{Proceedings of the IEEE Conference on Computer Vision and Pattern Recognition (CVPR)}, 2020.

\bibitem[Cao et~al.(2022)Cao, Xiao, Anandkumar, Xu, and Pavone]{cao2022advdo}
Yulong Cao, Chaowei Xiao, Anima Anandkumar, Danfei Xu, and Marco Pavone.
\newblock {AdvDO: Realistic Adversarial Attacks for Trajectory Prediction}.
\newblock In \emph{European Conference on Computer Vision (ECCV)}. Springer, 2022.

\bibitem[Carion et~al.(2020)Carion, Massa, Synnaeve, Usunier, Kirillov, and Zagoruyko]{carion2020end}
Nicolas Carion, Francisco Massa, Gabriel Synnaeve, Nicolas Usunier, Alexander Kirillov, and Sergey Zagoruyko.
\newblock {End-to-end Object Detection with Transformers}.
\newblock In \emph{European Conference on Computer Vision (ECCV)}, pages 213--229. Springer, 2020.

\bibitem[Carreira and Zisserman(2017)]{carreira2017quo}
Joao Carreira and Andrew Zisserman.
\newblock {Quo vadis, Action Recognition? A New Model and the Kinetics Dataset}.
\newblock In \emph{Proceedings of the IEEE Conference on Computer Vision and Pattern Recognition (CVPR)}, pages 6299--6308, 2017.

\bibitem[Chen et~al.(2016)Chen, Choi, and Chandraker]{chen2016atomic}
Chao-Yeh Chen, Wongun Choi, and Manmohan Chandraker.
\newblock {Atomic Scenes for Scalable Traffic Scene Recognition in Monocular Videos}.
\newblock In \emph{IEEE Winter Conference on Applications of Computer Vision (WACV)}, pages 1--9, 2016.

\bibitem[Chen and Kr\"ahenb\"uhl(2022)]{Chen_2022_CVPR}
Dian Chen and Philipp Kr\"ahenb\"uhl.
\newblock Learning from all vehicles.
\newblock In \emph{Proceedings of the IEEE/CVF Conference on Computer Vision and Pattern Recognition (CVPR)}, pages 17222--17231, 2022.

\bibitem[Chen et~al.(2018)Chen, Zhu, Papandreou, Schroff, and Adam]{chen2018encoder}
Liang-Chieh Chen, Yukun Zhu, George Papandreou, Florian Schroff, and Hartwig Adam.
\newblock Encoder-decoder with atrous separable convolution for semantic image segmentation.
\newblock In \emph{Proceedings of the European conference on computer vision (ECCV)}, pages 801--818, 2018.

\bibitem[Chitta et~al.(2021)Chitta, Prakash, and Geiger]{chitta2021neat}
Kashyap Chitta, Aditya Prakash, and Andreas Geiger.
\newblock Neat: Neural attention fields for end-to-end autonomous driving.
\newblock In \emph{Proceedings of the IEEE/CVF International Conference on Computer Vision}, pages 15793--15803, 2021.

\bibitem[Cho et~al.(2014)Cho, van Merri{\"e}nboer, Gulcehre, Bahdanau, Bougares, Schwenk, and Bengio]{cho-etal-2014-learning}
Kyunghyun Cho, Bart van Merri{\"e}nboer, Caglar Gulcehre, Dzmitry Bahdanau, Fethi Bougares, Holger Schwenk, and Yoshua Bengio.
\newblock {Learning Phrase Representations using {RNN} Encoder{--}Decoder for Statistical Machine Translation}.
\newblock In \emph{Conference on Empirical Methods in Natural Language Processing ({EMNLP})}, 2014.

\bibitem[Choi et~al.(2009)Choi, Shahid, and Savarese]{choi2009they}
Wongun Choi, Khuram Shahid, and Silvio Savarese.
\newblock What are they doing?: Collective activity classification using spatio-temporal relationship among people.
\newblock In \emph{IEEE International Conference on Computer Vision Workshops (ICCV-W)}, pages 1282--1289, 2009.

\bibitem[Cui et~al.(2019)Cui, Jia, Lin, Song, and Belongie]{cui2019class}
Yin Cui, Menglin Jia, Tsung-Yi Lin, Yang Song, and Serge Belongie.
\newblock Class-balanced loss based on effective number of samples.
\newblock In \emph{Proceedings of the IEEE/CVF conference on computer vision and pattern recognition}, pages 9268--9277, 2019.

\bibitem[Deng et~al.(2009)Deng, Dong, Socher, Li, Li, and Fei-Fei]{deng2009imagenet}
Jia Deng, Wei Dong, Richard Socher, Li-Jia Li, Kai Li, and Li Fei-Fei.
\newblock {Imagenet: A Large-scale Hierarchical Image Database}.
\newblock In \emph{IEEE Conference on Computer Vision and Pattern Recognition (CVPR)}, pages 248--255, 2009.

\bibitem[Ding et~al.(2023)Ding, Lin, Li, and Zhao]{ding2023causalaf}
Wenhao Ding, Haohong Lin, Bo Li, and Ding Zhao.
\newblock Causalaf: causal autoregressive flow for safety-critical driving scenario generation.
\newblock In \emph{Conference on Robot Learning}, pages 812--823. PMLR, 2023.

\bibitem[Dosovitskiy et~al.(2017)Dosovitskiy, Ros, Codevilla, Lopez, and Koltun]{Dosovitskiy17}
Alexey Dosovitskiy, German Ros, Felipe Codevilla, Antonio Lopez, and Vladlen Koltun.
\newblock {CARLA: An Open Urban Driving Simulator}.
\newblock In \emph{Proceedings of the 1st Annual Conference on Robot Learning}, pages 1--16, 2017.

\bibitem[Dosovitskiy et~al.(2021)Dosovitskiy, Beyer, Kolesnikov, Weissenborn, Zhai, Unterthiner, Dehghani, Minderer, Heigold, Gelly, Uszkoreit, and Houlsby]{dosovitskiy2021an}
Alexey Dosovitskiy, Lucas Beyer, Alexander Kolesnikov, Dirk Weissenborn, Xiaohua Zhai, Thomas Unterthiner, Mostafa Dehghani, Matthias Minderer, Georg Heigold, Sylvain Gelly, Jakob Uszkoreit, and Neil Houlsby.
\newblock {An Image is Worth 16x16 Words: Transformers for Image Recognition at Scale}.
\newblock In \emph{International Conference on Learning Representations (ICLR)}, 2021.

\bibitem[Elsayed et~al.(2022)Elsayed, Mahendran, van Steenkiste, Greff, Mozer, and Kipf]{elsayed2022savi++}
Gamaleldin Elsayed, Aravindh Mahendran, Sjoerd van Steenkiste, Klaus Greff, Michael~C Mozer, and Thomas Kipf.
\newblock {SAVi++: Towards End-to-End Object-centric Learning from Real-world Videos}.
\newblock \emph{Conference on Neural Information Processing Systems (NeurIPS)}, 2022.

\bibitem[Fan et~al.(2021{\natexlab{a}})Fan, Murrell, Wang, Alwala, Li, Li, Xiong, Ravi, Li, Yang, Malik, Girshick, Feiszli, Adcock, Lo, and Feichtenhofer]{fan2021pytorchvideo}
Haoqi Fan, Tullie Murrell, Heng Wang, Kalyan~Vasudev Alwala, Yanghao Li, Yilei Li, Bo Xiong, Nikhila Ravi, Meng Li, Haichuan Yang, Jitendra Malik, Ross Girshick, Matt Feiszli, Aaron Adcock, Wan-Yen Lo, and Christoph Feichtenhofer.
\newblock {PyTorchVideo: A Deep Learning Library for Video Understanding}.
\newblock In \emph{Proceedings of the 29th ACM International Conference on Multimedia}, 2021{\natexlab{a}}.
\newblock \url{https://pytorchvideo.org/}.

\bibitem[Fan et~al.(2021{\natexlab{b}})Fan, Xiong, Mangalam, Li, Yan, Malik, and Feichtenhofer]{fan2021multiscale}
Haoqi Fan, Bo Xiong, Karttikeya Mangalam, Yanghao Li, Zhicheng Yan, Jitendra Malik, and Christoph Feichtenhofer.
\newblock Multiscale vision transformers.
\newblock In \emph{Proceedings of the IEEE/CVF International Conference on Computer Vision (ICCV)}, pages 6824--6835, 2021{\natexlab{b}}.

\bibitem[Feichtenhofer(2020)]{feichtenhofer2020x3d}
Christoph Feichtenhofer.
\newblock {X3D: Expanding Architectures for Efficient Video Recognition}.
\newblock In \emph{Proceedings of the IEEE Conference on Computer Vision and Pattern Recognition (CVPR)}, pages 203--213, 2020.

\bibitem[Feichtenhofer et~al.(2019)Feichtenhofer, Fan, Malik, and He]{feichtenhofer2019slowfast}
Christoph Feichtenhofer, Haoqi Fan, Jitendra Malik, and Kaiming He.
\newblock {SlowFast Networks for Video Recognition}.
\newblock In \emph{Proceedings of the IEEE/CVF International Conference on Computer Vision (ICCV)}, pages 6202--6211, 2019.

\bibitem[Geiger et~al.(2012)Geiger, Lenz, and Urtasun]{Geiger2012CVPR}
Andreas Geiger, Philip Lenz, and Raquel Urtasun.
\newblock {Are we ready for Autonomous Driving? The KITTI Vision Benchmark Suite}.
\newblock In \emph{The IEEE Conference on Computer Vision and Pattern Recognition (CVPR)}, 2012.

\bibitem[Gu et~al.(2018)Gu, Sun, Ross, Vondrick, Pantofaru, Li, Vijayanarasimhan, Toderici, Ricco, Sukthankar, et~al.]{gu2018ava}
Chunhui Gu, Chen Sun, David~A Ross, Carl Vondrick, Caroline Pantofaru, Yeqing Li, Sudheendra Vijayanarasimhan, George Toderici, Susanna Ricco, Rahul Sukthankar, et~al.
\newblock {AVA: A Video Dataset of Spatio-temporally Localized Atomic Visual Actions}.
\newblock In \emph{Proceedings of the IEEE Conference on Computer Vision and Pattern Recognition (CVPR)}, pages 6047--6056, 2018.

\bibitem[Gu et~al.(2023)Gu, Hu, Zhang, Chen, Wang, Wang, and Zhao]{vip3d}
Junru Gu, Chenxu Hu, Tianyuan Zhang, Xuanyao Chen, Yilun Wang, Yue Wang, and Hang Zhao.
\newblock Vip3d: End-to-end visual trajectory prediction via 3d agent queries.
\newblock In \emph{Proceedings of the IEEE/CVF Conference on Computer Vision and Pattern Recognition}, pages 5496--5506, 2023.

\bibitem[Hanselmann et~al.(2022)Hanselmann, Renz, Chitta, Bhattacharyya, and Geiger]{hanselmann2022king}
Niklas Hanselmann, Katrin Renz, Kashyap Chitta, Apratim Bhattacharyya, and Andreas Geiger.
\newblock King: Generating safety-critical driving scenarios for robust imitation via kinematics gradients.
\newblock In \emph{European Conference on Computer Vision}, pages 335--352. Springer, 2022.

\bibitem[He et~al.(2016)He, Zhang, Ren, and Sun]{he2016deep}
Kaiming He, Xiangyu Zhang, Shaoqing Ren, and Jian Sun.
\newblock Deep residual learning for image recognition.
\newblock In \emph{Proceedings of the IEEE Conference on Computer Vision and Pattern Recognition (CVPR)}, pages 770--778, 2016.

\bibitem[He et~al.(2017)He, Gkioxari, Doll{\'a}r, and Girshick]{he2017mask}
Kaiming He, Georgia Gkioxari, Piotr Doll{\'a}r, and Ross Girshick.
\newblock {Mask R-CNN}.
\newblock In \emph{Proceedings of the IEEE/CVF International Conference on Computer Vision (ICCV)}, pages 2961--2969, 2017.

\bibitem[Hu et~al.(2023)Hu, Yang, Chen, Li, Sima, Zhu, Chai, Du, Lin, Wang, Lu, Jia, Liu, Dai, Qiao, and Li]{hu2023_uniad}
Yihan Hu, Jiazhi Yang, Li Chen, Keyu Li, Chonghao Sima, Xizhou Zhu, Siqi Chai, Senyao Du, Tianwei Lin, Wenhai Wang, Lewei Lu, Xiaosong Jia, Qiang Liu, Jifeng Dai, Yu Qiao, and Hongyang Li.
\newblock Planning-oriented autonomous driving.
\newblock In \emph{Proceedings of the IEEE/CVF Conference on Computer Vision and Pattern Recognition}, 2023.

\bibitem[Ibrahim et~al.(2016)Ibrahim, Muralidharan, Deng, Vahdat, and Mori]{volleyball}
Mostafa~S Ibrahim, Srikanth Muralidharan, Zhiwei Deng, Arash Vahdat, and Greg Mori.
\newblock A hierarchical deep temporal model for group activity recognition.
\newblock In \emph{Proceedings of the IEEE Conference on Computer Vision and Pattern Recognition (CVPR)}, pages 1971--1980, 2016.

\bibitem[Jain et~al.(2015)Jain, Koppula, Raghavan, Soh, and Saxena]{brain4car2015}
Ashesh Jain, Hema~S. Koppula, Bharad Raghavan, Shane Soh, and Ashutosh Saxena.
\newblock {Car that Knows Before You Do: Anticipating Maneuvers via Learning Temporal Driving Models}.
\newblock In \emph{IEEE International Conference on Computer Vision (ICCV)}, 2015.

\bibitem[Jiang et~al.(2014)Jiang, Liu, Roshan~Zamir, Toderici, Laptev, Shah, and Sukthankar]{THUMOS14}
Y.-G. Jiang, J. Liu, A. Roshan~Zamir, G. Toderici, I. Laptev, M. Shah, and R. Sukthankar.
\newblock {THUMOS} challenge: Action recognition with a large number of classes.
\newblock \url{http://crcv.ucf.edu/THUMOS14/}, 2014.

\bibitem[Johnson et~al.(2017)Johnson, Hariharan, Van Der~Maaten, Fei-Fei, Lawrence~Zitnick, and Girshick]{johnson2017clevr}
Justin Johnson, Bharath Hariharan, Laurens Van Der~Maaten, Li Fei-Fei, C Lawrence~Zitnick, and Ross Girshick.
\newblock {CLEVR: A Diagnostic Dataset for Compositional Language and Elementary Visual Reasoning}.
\newblock In \emph{Proceedings of the IEEE Conference on Computer Vision and Pattern Recognition (CVPR)}, pages 2901--2910, 2017.

\bibitem[Kay et~al.(2017)Kay, Carreira, Simonyan, Zhang, Hillier, Vijayanarasimhan, Viola, Green, Back, Natsev, et~al.]{kay2017kinetics}
Will Kay, Joao Carreira, Karen Simonyan, Brian Zhang, Chloe Hillier, Sudheendra Vijayanarasimhan, Fabio Viola, Tim Green, Trevor Back, Paul Natsev, et~al.
\newblock The kinetics human action video dataset.
\newblock \emph{arXiv preprint arXiv:1705.06950}, 2017.

\bibitem[Kipf et~al.(2022)Kipf, Elsayed, Mahendran, Stone, Sabour, Heigold, Jonschkowski, Dosovitskiy, and Greff]{kipf2022conditional}
Thomas Kipf, Gamaleldin~F. Elsayed, Aravindh Mahendran, Austin Stone, Sara Sabour, Georg Heigold, Rico Jonschkowski, Alexey Dosovitskiy, and Klaus Greff.
\newblock {Conditional Object-Centric Learning from Video}.
\newblock In \emph{International Conference on Learning Representations (ICLR)}, 2022.

\bibitem[Kipf and Welling(2016)]{kipf2016semi}
Thomas~N Kipf and Max Welling.
\newblock Semi-supervised classification with graph convolutional networks.
\newblock \emph{arXiv preprint arXiv:1609.02907}, 2016.

\bibitem[Kuehne et~al.(2011)Kuehne, Jhuang, Garrote, Poggio, and Serre]{kuehne2011hmdb}
Hildegard Kuehne, Hueihan Jhuang, Est{\'\i}baliz Garrote, Tomaso Poggio, and Thomas Serre.
\newblock {HMDB: A Large Video Database for Human Motion Recognition}.
\newblock In \emph{2011 International Conference on Computer Vision (ICCV)}, pages 2556--2563, 2011.

\bibitem[Kung et~al.(2023)Kung, Yang, Pao, Lu, Chen, Lu, and Chen]{kung2023riskbench}
Chi-Hsi Kung, Chieh-Chi Yang, Pang-Yuan Pao, Shu-Wei Lu, Pin-Lun Chen, Hsin-Cheng Lu, and Yi-Ting Chen.
\newblock Riskbench: A scenario-based benchmark for risk identification.
\newblock \emph{arXiv preprint arXiv:2312.01659}, 2023.

\bibitem[Li et~al.(2019)Li, Jiang, Che, Shi, Liu, Meng, Ye, and Liu]{li2019dbus}
Max~Guangyu Li, Bo Jiang, Zhengping Che, Xuefeng Shi, Mengyao Liu, Yiping Meng, Jieping Ye, and Yan Liu.
\newblock {DBUS: Human Driving Behavior Understanding System}.
\newblock In \emph{ICCV Workshops}, pages 2436--2444, 2019.

\bibitem[Lin et~al.(2014)Lin, Fidler, Kong, and Urtasun]{lin2014visual}
Dahua Lin, Sanja Fidler, Chen Kong, and Raquel Urtasun.
\newblock {Visual Semantic Search: Retrieving Videos via Complex Textual Queries}.
\newblock In \emph{Proceedings of the IEEE/CVF Conference on Computer Vision and Pattern Recognition}, 2014.

\bibitem[Liu et~al.(2020)Liu, Adeli, Cao, Lee, Shenoi, Gaidon, and Niebles]{stipicra2020}
Bingbin Liu, Ehsan Adeli, Zhangjie Cao, Kuan-Hui Lee, Abhijeet Shenoi, Adrien Gaidon, and Juan~Carlos Niebles.
\newblock Spatiotemporal relationship reasoning for pedestrian intent prediction.
\newblock pages 3485--3492, 2020.

\bibitem[Locatello et~al.(2020)Locatello, Weissenborn, Unterthiner, Mahendran, Heigold, Uszkoreit, Dosovitskiy, and Kipf]{locatello2020object}
Francesco Locatello, Dirk Weissenborn, Thomas Unterthiner, Aravindh Mahendran, Georg Heigold, Jakob Uszkoreit, Alexey Dosovitskiy, and Thomas Kipf.
\newblock {Object-centric Learning with Slot Attention}.
\newblock \emph{Advances in Neural Information Processing Systems (NeurIPS)}, 33:\penalty0 11525--11538, 2020.

\bibitem[Loshchilov and Hutter(2019)]{loshchilov2018decoupled}
Ilya Loshchilov and Frank Hutter.
\newblock {Decoupled Weight Decay Regularization}.
\newblock In \emph{International Conference on Learning Representations}, 2019.

\bibitem[Maggiolino et~al.(2023)Maggiolino, Ahmad, Cao, and Kitani]{maggiolino2023deep}
Gerard Maggiolino, Adnan Ahmad, Jinkun Cao, and Kris Kitani.
\newblock {Deep OC-SORT: Multi-Pedestrian Tracking by Adaptive Re-Identification}.
\newblock \emph{arXiv preprint arXiv:2302.11813}, 2023.

\bibitem[Malla et~al.(2020)Malla, Dariush, and Choi]{malla2020titan}
Srikanth Malla, Behzad Dariush, and Chiho Choi.
\newblock {TITAN: Future Forecast using Action Priors}.
\newblock In \emph{Proceedings of the IEEE/CVF Conference on Computer Vision and Pattern Recognition}, pages 11186--11196, 2020.

\bibitem[Mohamed et~al.(2020)Mohamed, Qian, Elhoseiny, and Claudel]{mohamed2020social}
Abduallah Mohamed, Kun Qian, Mohamed Elhoseiny, and Christian Claudel.
\newblock Social-stgcnn: A social spatio-temporal graph convolutional neural network for human trajectory prediction.
\newblock In \emph{Proceedings of the IEEE/CVF conference on computer vision and pattern recognition}, pages 14424--14432, 2020.

\bibitem[Nagarajan et~al.(2022)Nagarajan, Ramakrishnan, Desai, Hillis, and Grauman]{ego-env}
Tushar Nagarajan, Santhosh~Kumar Ramakrishnan, Ruta Desai, James Hillis, and Kristen Grauman.
\newblock Egoenv: Human-centric environment representations from egocentric video.
\newblock \emph{arXiv preprint arXiv:2207.11365}, 2022.

\bibitem[Naphade et~al.(2023)Naphade, Wang, Anastasiu, Tang, Chang, Yao, Zheng, Rahman, Arya, Sharma, Feng, Ablavsky, Sclaroff, Chakraborty, Prajapati, Li, Li, Kunadharaju, Jiang, and Chellappa]{Naphade23AIC23}
Milind Naphade, Shuo Wang, David~C. Anastasiu, Zheng Tang, Ming-Ching Chang, Yue Yao, Liang Zheng, Mohammed~Shaiqur Rahman, Meenakshi~S. Arya, Anuj Sharma, Qi Feng, Vitaly Ablavsky, Stan Sclaroff, Pranamesh Chakraborty, Sanjita Prajapati, Alice Li, Shangru Li, Krishna Kunadharaju, Shenxin Jiang, and Rama Chellappa.
\newblock {The 7th AI City Challenge}.
\newblock In \emph{The IEEE Conference on Computer Vision and Pattern Recognition (CVPR) Workshops}, 2023.

\bibitem[Ohn-Bar et~al.(2020)Ohn-Bar, Prakash, Behl, Chitta, and Geiger]{Ohn-Bar_2020_CVPR}
Eshed Ohn-Bar, Aditya Prakash, Aseem Behl, Kashyap Chitta, and Andreas Geiger.
\newblock Learning situational driving.
\newblock In \emph{Proceedings of the IEEE/CVF Conference on Computer Vision and Pattern Recognition (CVPR)}, 2020.

\bibitem[Ramanishka et~al.(2018)Ramanishka, Chen, Misu, and Saenko]{ramanishka2018toward}
Vasili Ramanishka, Yi-Ting Chen, Teruhisa Misu, and Kate Saenko.
\newblock {Toward Driving Scene Understanding: A Dataset for Learning Driver Behavior and Causal Reasoning}.
\newblock In \emph{Proceedings of the IEEE Conference on Computer Vision and Pattern Recognition}, pages 7699--7707, 2018.

\bibitem[Rasouli et~al.(2019)Rasouli, Kotseruba, Kunic, and Tsotsos]{pieiccv2019}
Amir Rasouli, Iuliia Kotseruba, Toni Kunic, and John~K. Tsotsos.
\newblock {PIE: A Large-Scale Dataset and Models for Pedestrian Intention Estimation and Trajectory Prediction}.
\newblock In \emph{The IEEE International Conference on Computer Vision (ICCV)}, 2019.

\bibitem[Rempe et~al.(2022)Rempe, Philion, Guibas, Fidler, and Litany]{rempe2022strive}
Davis Rempe, Jonah Philion, Leonidas~J. Guibas, Sanja Fidler, and Or Litany.
\newblock Generating useful accident-prone driving scenarios via a learned traffic prior.
\newblock In \emph{Conference on Computer Vision and Pattern Recognition (CVPR)}, 2022.

\bibitem[Rempe et~al.(2023)Rempe, Luo, Peng, Yuan, Kitani, Kreis, Fidler, and Litany]{rempeluo2023tracepace}
Davis Rempe, Zhengyi Luo, Xue~Bin Peng, Ye Yuan, Kris Kitani, Karsten Kreis, Sanja Fidler, and Or Litany.
\newblock Trace and pace: Controllable pedestrian animation via guided trajectory diffusion.
\newblock In \emph{Conference on Computer Vision and Pattern Recognition (CVPR)}, 2023.

\bibitem[Renz et~al.(2022)Renz, Chitta, Mercea, Koepke, Akata, and Geiger]{renz2022plant}
Katrin Renz, Kashyap Chitta, Otniel-Bogdan Mercea, A Koepke, Zeynep Akata, and Andreas Geiger.
\newblock Plant: Explainable planning transformers via object-level representations.
\newblock \emph{arXiv preprint arXiv:2210.14222}, 2022.

\bibitem[Sajjadi et~al.(2022)Sajjadi, Duckworth, Mahendran, van Steenkiste, Pavetic, Lucic, Guibas, Greff, and Kipf]{NEURIPS2022_3dc83fcf}
Mehdi S.~M. Sajjadi, Daniel Duckworth, Aravindh Mahendran, Sjoerd van Steenkiste, Filip Pavetic, Mario Lucic, Leonidas~J Guibas, Klaus Greff, and Thomas Kipf.
\newblock Object scene representation transformer.
\newblock In \emph{Advances in Neural Information Processing Systems}, pages 9512--9524. Curran Associates, Inc., 2022.

\bibitem[Segal et~al.(2020)Segal, Kee, Luo, Sadat, Yumer, and Urtasun]{Segal_universal_corl2020}
Sean Segal, Eric Kee, Wenjie Luo, Abbas Sadat, Ersin Yumer, and Raquel Urtasun.
\newblock {Universal Embeddings for Spatio-Temporal Tagging of Self-Driving Logs}.
\newblock In \emph{Conference on Robot Learning (CoRL)}, 2020.

\bibitem[Shao et~al.(2023)Shao, Wang, Chen, Waslander, Li, and Liu]{shao2023reasonnet}
Hao Shao, Letian Wang, Ruobing Chen, Steven~L Waslander, Hongsheng Li, and Yu Liu.
\newblock Reasonnet: End-to-end driving with temporal and global reasoning.
\newblock In \emph{Proceedings of the IEEE/CVF Conference on Computer Vision and Pattern Recognition}, pages 13723--13733, 2023.

\bibitem[Sigurdsson et~al.(2016)Sigurdsson, Varol, Wang, Farhadi, Laptev, and Gupta]{sigurdsson2016hollywood}
Gunnar~A Sigurdsson, G{\"u}l Varol, Xiaolong Wang, Ali Farhadi, Ivan Laptev, and Abhinav Gupta.
\newblock {Hollywood in Homes: Crowdsourcing Data Collection for Activity Understanding}.
\newblock In \emph{Proceedings of the European Conference on Computer Vision (ECCV)}, pages 510--526. Springer, 2016.

\bibitem[Singh et~al.(2022)Singh, Akrigg, Di~Maio, Fontana, Alitappeh, Khan, Saha, Jeddisaravi, Yousefi, Culley, et~al.]{singh2022road}
Gurkirt Singh, Stephen Akrigg, Manuele Di~Maio, Valentina Fontana, Reza~Javanmard Alitappeh, Salman Khan, Suman Saha, Kossar Jeddisaravi, Farzad Yousefi, Jacob Culley, et~al.
\newblock {ROAD: The ROad event Awareness Dataset for Autonomous Driving}.
\newblock \emph{IEEE Transactions on Pattern Analysis and Machine Intelligence}, 45\penalty0 (1):\penalty0 1036--1054, 2022.

\bibitem[Sun et~al.(2020)Sun, Kretzschmar, Dotiwalla, Chouard, Patnaik, Tsui, Guo, Zhou, Chai, Caine, et~al.]{sun2020scalability}
Pei Sun, Henrik Kretzschmar, Xerxes Dotiwalla, Aurelien Chouard, Vijaysai Patnaik, Paul Tsui, James Guo, Yin Zhou, Yuning Chai, Benjamin Caine, et~al.
\newblock {Scalability in Perception for Autonomous Driving: Waymo Open Dataset}.
\newblock In \emph{Proceedings of the IEEE/CVF Conference on Computer Vision and Pattern Recognition}, 2020.

\bibitem[Sun et~al.(2022)Sun, Segu, Postels, Wang, Van~Gool, Schiele, Tombari, and Yu]{sun2022shift}
Tao Sun, Mattia Segu, Janis Postels, Yuxuan Wang, Luc Van~Gool, Bernt Schiele, Federico Tombari, and Fisher Yu.
\newblock Shift: a synthetic driving dataset for continuous multi-task domain adaptation.
\newblock In \emph{Proceedings of the IEEE/CVF Conference on Computer Vision and Pattern Recognition}, pages 21371--21382, 2022.

\bibitem[Taha et~al.(2020)Taha, Chen, Misu, Shrivastava, and Davis]{driverwacv2020}
Ahmed Taha, Yi-Ting Chen, Teruhisa Misu, Abhinav Shrivastava, and Larry Davis.
\newblock {Boosting Standard Classification Architectures Through a Ranking Regularizer}.
\newblock In \emph{IEEE/CVF Winter Conference on Applications of Computer Vision (WACV)}, 2020.

\bibitem[Tan et~al.(2023)Tan, Nagarajan, and Grauman]{tan2023egodistill}
Shuhan Tan, Tushar Nagarajan, and Kristen Grauman.
\newblock Egodistill: Egocentric head motion distillation for efficient video understanding.
\newblock \emph{arXiv preprint arXiv:2301.02217}, 2023.

\bibitem[Tong et~al.(2022)Tong, Song, Wang, and Wang]{tong2022videomae}
Zhan Tong, Yibing Song, Jue Wang, and Limin Wang.
\newblock Video{MAE}: Masked autoencoders are data-efficient learners for self-supervised video pre-training.
\newblock In \emph{Advances in Neural Information Processing Systems}, 2022.

\bibitem[Tran et~al.(2019)Tran, Wang, Torresani, and Feiszli]{tran2019video}
Du Tran, Heng Wang, Lorenzo Torresani, and Matt Feiszli.
\newblock {Video Classification with Channel-separated Convolutional Networks}.
\newblock In \emph{Proceedings of the IEEE/CVF International Conference on Computer Vision (ICCV)}, pages 5552--5561, 2019.

\bibitem[Vaswani et~al.(2017)Vaswani, Shazeer, Parmar, Uszkoreit, Jones, Gomez, Kaiser, and Polosukhin]{vaswani2017attention}
Ashish Vaswani, Noam Shazeer, Niki Parmar, Jakob Uszkoreit, Llion Jones, Aidan~N Gomez, {\L}ukasz Kaiser, and Illia Polosukhin.
\newblock Attention is all you need.
\newblock \emph{Conference on Neural Information Processing Systems}, 30, 2017.

\bibitem[Wang et~al.(2018)Wang, Girshick, Gupta, and He]{wang2018non}
Xiaolong Wang, Ross Girshick, Abhinav Gupta, and Kaiming He.
\newblock Non-local neural networks.
\newblock In \emph{Proceedings of the IEEE conference on computer vision and pattern recognition}, pages 7794--7803, 2018.

\bibitem[Wu et~al.(2019)Wu, Wang, Wang, Guo, and Wu]{CVPR2019_ARG}
Jianchao Wu, Limin Wang, Li Wang, Jie Guo, and Gangshan Wu.
\newblock {Learning Actor Relation Graphs for Group Activity Recognition}.
\newblock In \emph{Proceedings of the IEEE/CVF International Conference on Computer Vision (ICCV)}, 2019.

\bibitem[Wu et~al.(2020)Wu, Huang, Liu, Wang, and Lin]{DistributionBalancedLoss}
Tong Wu, Qingqiu Huang, Ziwei Liu, Yu Wang, and Dahua Lin.
\newblock Distribution-balanced loss for multi-label classification in long-tailed datasets.
\newblock In \emph{European Conference on Computer Vision (ECCV)}, 2020.

\bibitem[Xiao et~al.(2022)Xiao, Yuille, and Chen]{semanticregioncorl2022}
Zihao Xiao, Allan Yuille, and Yi-Ting Chen.
\newblock {Learning Road Scene-level Representations via Semantic Region Prediction}.
\newblock In \emph{Conference on Robot Learning (CoRL)}, 2022.

\bibitem[Xu et~al.(2022)Xu, Ding, Lyu, Liu, Wang, He, Hu, Zhao, and Li]{xu2022safebench}
Chejian Xu, Wenhao Ding, Weijie Lyu, Zuxin Liu, Shuai Wang, Yihan He, Hanjiang Hu, Ding Zhao, and Bo Li.
\newblock Safebench: A benchmarking platform for safety evaluation of autonomous vehicles.
\newblock In \emph{Thirty-sixth Conference on Neural Information Processing Systems Datasets and Benchmarks Track}, 2022.

\bibitem[Yang et~al.(2020)Yang, Xu, Shi, Dai, and Zhou]{yang2020temporal}
Ceyuan Yang, Yinghao Xu, Jianping Shi, Bo Dai, and Bolei Zhou.
\newblock Temporal pyramid network for action recognition.
\newblock In \emph{Proceedings of the IEEE/CVF conference on computer vision and pattern recognition}, pages 591--600, 2020.

\bibitem[Yeung et~al.(2017)Yeung, Russakovsky, Jin, Andriluka, Mori, and Fei-Fei]{yeung2015every}
Serena Yeung, Olga Russakovsky, Ning Jin, Mykhaylo Andriluka, Greg Mori, and Li Fei-Fei.
\newblock Every moment counts: Dense detailed labeling of actions in complex videos.
\newblock \emph{International Journal of Computer Vision}, 2017.

\bibitem[Zhang et~al.(2023)Zhang, Huang, and Ohn-Bar]{Zhang_2023_CVPR}
Jimuyang Zhang, Zanming Huang, and Eshed Ohn-Bar.
\newblock Coaching a teachable student.
\newblock In \emph{Proceedings of the IEEE/CVF Conference on Computer Vision and Pattern Recognition (CVPR)}, pages 7805--7815, 2023.

\bibitem[Zhou et~al.(2022)Zhou, Zhang, Lee, Sun, Li, Zhu, Yoo, Qi, and Han]{zhou2022slot}
Yi Zhou, Hui Zhang, Hana Lee, Shuyang Sun, Pingjun Li, Yangguang Zhu, ByungIn Yoo, Xiaojuan Qi, and Jae-Joon Han.
\newblock {Slot-VPS: Object-centric Representation Learning for Video Panoptic Segmentation}.
\newblock In \emph{Proceedings of the IEEE/CVF Conference on Computer Vision and Pattern Recognition (CVPR)}, pages 3093--3103, 2022.

\bibitem[Zhou et~al.(2023)Zhou, Wang, Li, and Huang]{zhou2023query}
Zikang Zhou, Jianping Wang, Yung-Hui Li, and Yu-Kai Huang.
\newblock Query-centric trajectory prediction.
\newblock In \emph{Proceedings of the IEEE/CVF Conference on Computer Vision and Pattern Recognition}, pages 17863--17873, 2023.

\end{thebibliography}
}
\clearpage
\appendix

\renewcommand\thesection{\Alph{section}}

\label{sec:sup_abstract}
 \section*{\LARGE Appendix}

In this appendix, we cover,
\begin{enumerate}[label=\Alph*)]

\item Construction of the TACO dataset

\item nuScenes annotation

\item More ablation study of Action-slot

\item Analysis in challenging scenarios

\item Analysis of the TACO Dataset

\item Limitations

\item Implementation details

\end{enumerate}    
\section{Construction of The TACO Dataset}
\label{sec:sup_taco}

We introduce the construction of the proposed Traffic Activity Recognition (TACO) dataset. We leverage the CARLA simulator~\cite{Dosovitskiy17} to collect arbitrary traffic scenarios for achieving a balanced activity class distribution in a large scale. 

\paragraph{Scenario Collection.}
Certain atomic activities are rare and difficult to collect in the real world, as shown in the class distribution comparison of TACO and OATS in Figure 2 of the main paper
%
% To collect arbitrary traffic scenarios, 
We propose to leverage the CARLA simulator~\cite{dosovitskiy2021an} to construct the synthetic dataset. 
% We want to highlight that certain atomic activities are rare and difficult to collect in the real world, as shown in the class distribution comparison of TACO and OATS in Figure 2.
%
We choose CARLA simulator because it is widely accepted and popular in the computer vision community, where it provides various sensor suites and high-fidelity simulations to facilitate autonomous driving development and test~\cite{Ohn-Bar_2020_CVPR,chitta2021neat,carlachallenge2022,xu2022safebench,Chen_2022_CVPR,renz2022plant,Zhang_2023_CVPR,shao2023reasonnet}, safety-critical scenario generation~\cite{rempe2022strive,hanselmann2022king,ding2023causalaf}, and domain adaption~\cite{sun2022shift}.

\begin{figure}[h]
\centering
    \includegraphics[width=8cm]{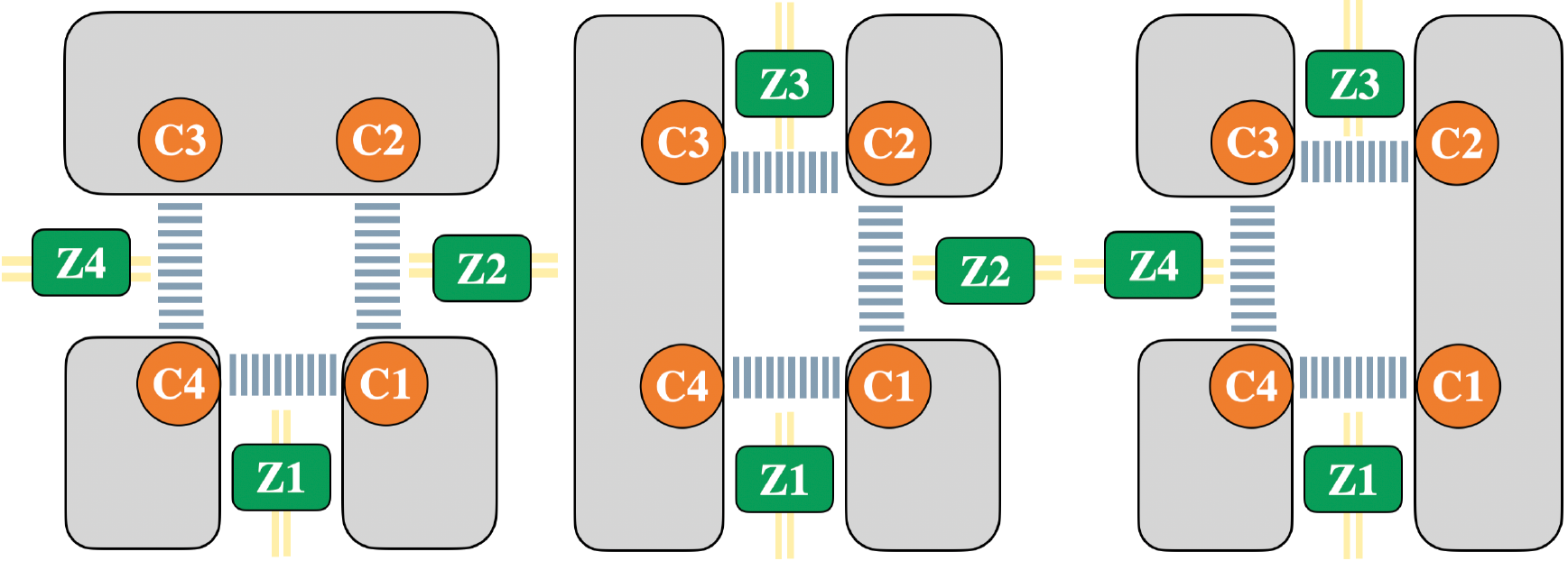}
        \caption{Illustration of the topology in atomic activity for three types of T-intersections.
        }
        \label{fig:t-intersection}
        \vspace{-4mm}
\end{figure}

\begin{figure}[h]
\centering
    \includegraphics[width=8cm]{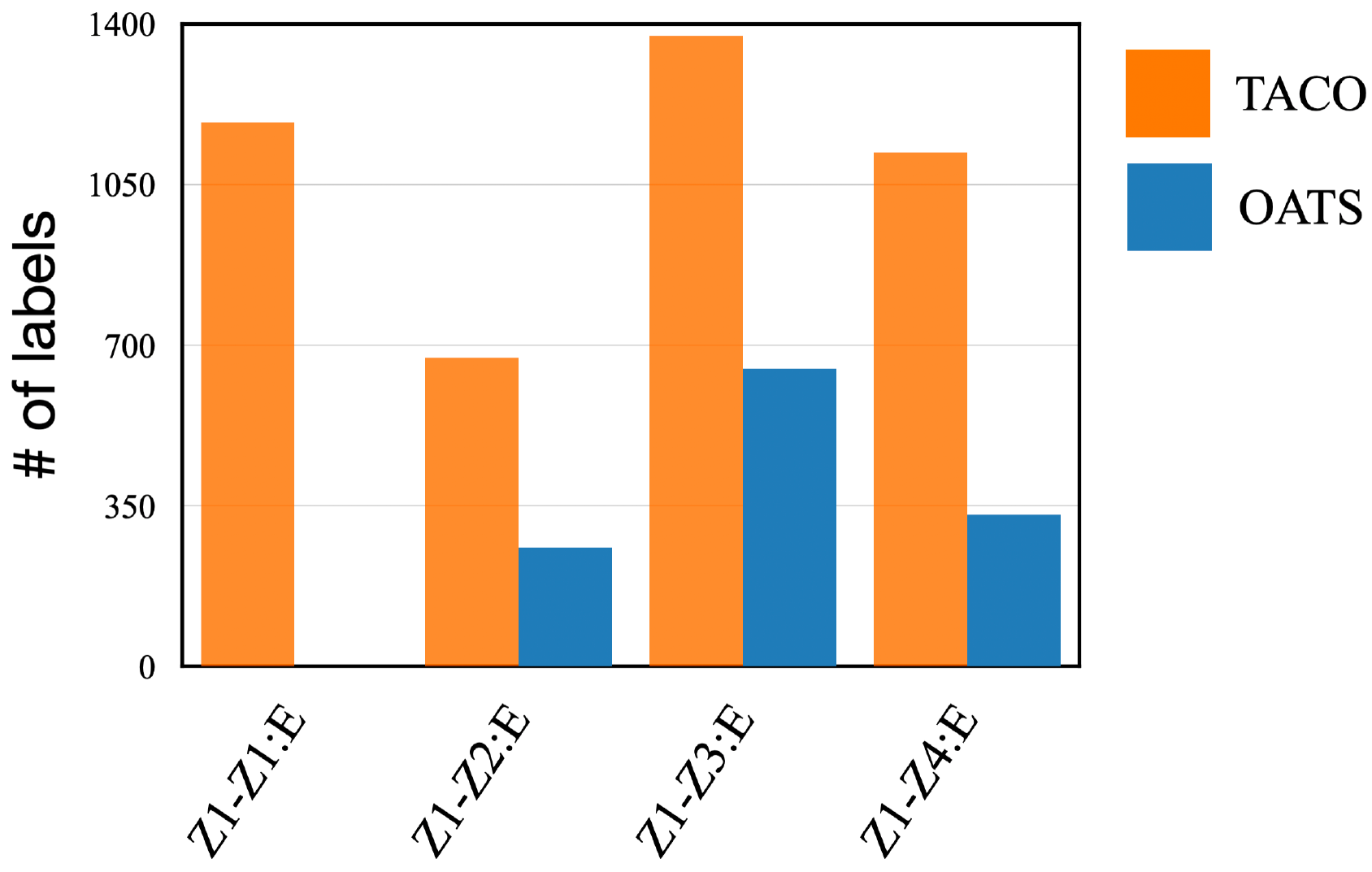}
        \caption{Ego-vehicle's action distribution in TACO and OATS.
        }
        \label{fig:ego_action}
        \vspace{-4mm}
\end{figure}

\begin{figure*}[h]
\centering
    \includegraphics[width=16cm]{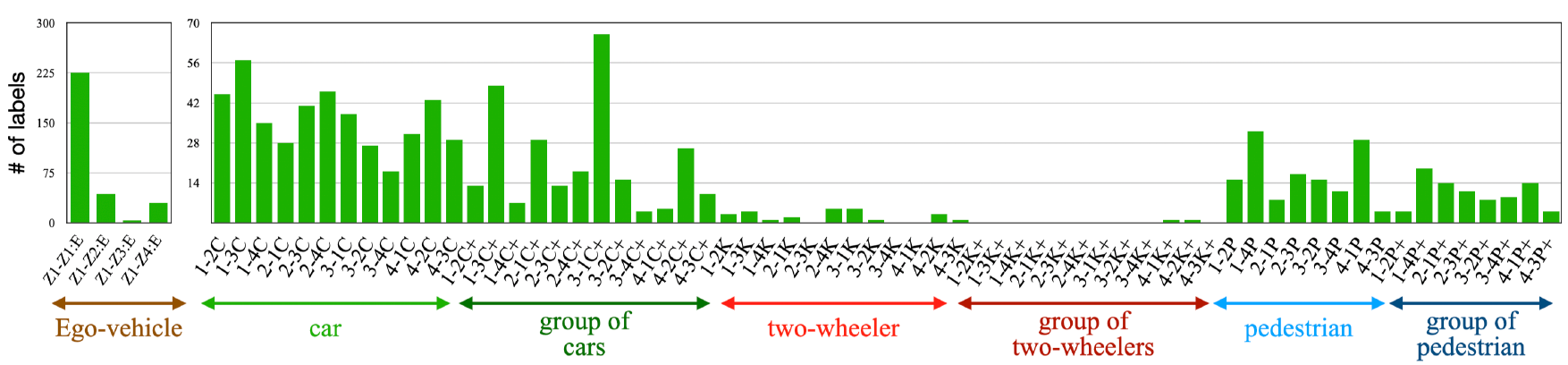}
        \caption{The distribution of atomic activities  
        % class and ego-vehicle's action distribution 
        in the nuScenes dataset~\cite{nuscenes}.
        Note that we neglect the notation of topology (i.e., roadway \textbf{Z} and corner \textbf{C}) in the x-axis due to limited space.
        }
        \label{fig:nuscenes_distribution}
        \vspace{-4mm}
\end{figure*}

Scenarios are collected in 
% We use 
the built-in maps (i.e., Town01, Town02, Town03, Town04, Town05, Town06, Town07, and Town10HD) defined in CARLA 0.9.14.
% , specifically, . 
We use Town10HD as the testing set.
The rest are for the training set. 
%
% We first identify all intersections in each map and then specifically collect scenarios on these intersections.
We pinpoint all intersections on each map and subsequently gather specific scenarios related to these intersections.
Note that, we also collect T-intersections, following 
% by extending 
the topology definition discussed in OATS~\cite{Agarwal_2023_ICCV}, as shown in Figure~\ref{fig:t-intersection}.

We collect scenarios with the following three approaches:

\begin{enumerate}[label=(\alph*)]
\item \textbf{Auto-pilot} 
We use the built-in auto-pilot to control ego-vehicle and road users, including vehicles and pedestrians. 
%
% and let them run without any destination constraints.
We do not set destinations for all road users.  
We program an automatic scenario collection process when %by asking the ego-vehicle to start recording a scenario when 
(1) ego-vehicle approaches an intersection and  
(2) ego-vehicle is surrounded by at least one road user. 
% surrounds.

The length of the recordings 
% time 
varies from 51 to 242 frames. 
We set the duration for capturing various ego motions.
%
% For the number of road users, 
We randomly set the numbers of road users and spawn them on a map. 
% of spawned vehicles and pedestrians empirically to 
% keep the scene crowded while avoiding traffic deadlock for each distinct map.
%
To collect diverse atomic activities, we set all road users (except for the ego-vehicle) to ignore any traffic rules, including traffic lights and stop signs.

\item \textbf{Automatic Scenario Generation~\cite{kung2023riskbench}}. We use existing pre-recorded \textit{basic scenarios} in RiskBench~\cite{kung2023riskbench} and automatically generate diverse scenarios from a \textit{basic scenario}. Specifically, we select the pre-recorded basic scenarios from \textit{interactive scenarios} where ego-vehicle interacts with other risky road users and \textit{non-interactive scenarios} where ego-vehicle does not interact with any road user. Then we follow the augmentation process proposed in RiskBench~\cite{kung2023riskbench} to automatically generate diverse scenarios via injecting random road users and changing the weather. The scenarios collected by RiskBench provide more human-like maneuvers and more risky interactions compared to the scenarios collected by the auto-pilot.

\item \textbf{Scenario Runner~\cite{scenario_runner}}. Scenario runner, a scenario collection tool developed by CARLA~\cite{Dosovitskiy17}, can explicitly generate a scripted scenario by defining a set of routes for road users. 
Although auto-pilot and automatic scenario generation can help collect diverse traffic activities, it is difficult to generate some specific classes of atomic activity frequently. 
For example, \textit{Z3-Z2:K+}, a group of bicyclists turns left from the opposite roadway of ego-vehicle. 
Therefore, we leverage scenario runner to explicitly collect the scenarios that are difficult for the auto-pilot method.
%
% Specifically, we first define the routes for each intersection. Then we generate road users to follow the route.
%
A scenario starts to collect when 
% the scenario runner start when
ego vehicle reaches a trigger point (i.e., location) predefined in the script. 
A scenario ends when all scripted actions are accomplished.
To further enhance the diversity of scenarios, 
% collected by the scenario runner, 
we randomly spawn road users surrounding the ego vehicle. 
Note that road users are not guaranteed to be involved in an atomic activity.
\end{enumerate}
For all collection methods, we randomly set the weather and light conditions. 
Note that we exclude night scenes because of the poor visibility.

\paragraph{Sensor Suites.}
We deploy a wide field-of-view camera (120 degrees)
% range of 120 degrees 
% of the camera 
% to better 
to record events taking place on the extreme left and right sides of the ego-vehicle.
For example, pedestrians crossing the street on the left (\textit{C3-C4:P} and \textit{C4-C3:P}), and a vehicle turns right from the left roadway (\textit{Z4-Z1:C}). 
We collect the corresponding images and instance segmentation.
% for the perspective view.
%
In addition, we collect instance segmentation from Bird's eye view.

% for Atomic Activities
\paragraph{Annotation Criterion.}
We follow the same annotation criterion defined in OATS~\cite{Agarwal_2023_ICCV}. 
%
% Since t
The annotation of atomic activities can be subjective, e.g., whether a turning right car that stops for a crossing pedestrian should be annotated. 
Thus, the whole annotation work is done by one person to ensure annotation consistency. 
In addition, we list a set of annotation criteria to enhance the quality of data. 
If any of the criteria is valid in a video, the annotator is asked to discard the whole scenario.
\begin{enumerate}
    \item Annotator cannot determine the starting roadway of a road user. 
    \item Annotator cannot determine a road user's destination.
    \item Annotator cannot determine any road user's action in a video.
    \item One of the road users has completed an atomic activity at the beginning of the video, e.g., driving away from \textbf{Z4} and almost arriving at \textbf{Z1}.
\end{enumerate}
% We further discard scenarios with zero atomic activity, i.e., there is no other road user performing any action, to alleviate the positive-negative imbalance issue in multi-label recognition
We additionally filter out scenarios with zero atomic activity, meaning there are no other road users engaged in any actions. This helps address the positive-negative imbalance issue in multi-label recognition~\cite{sigurdsson2016hollywood,cui2019class,DistributionBalancedLoss}.
%
% We also guarantee that each atomic activity should be able to be observed for at least a specific duration.
We also ensure that each atomic activity remains observable for a minimum specified duration.
%
% This is because the video will be subsampled to a fixed-length $K$ short clip before being fed to a model, where the length depends on the input sequence of a model.
This is due to the fact that the video will be subsampled into a fixed-length $K$ short clip before being inputted into a model.
%
% As most models we evaluated take 16-frame clips, we set $K$ as 16. Therefore, given an N frame of video, we guarantee that each atomic activity should appear for at least $N/K$ frames.
We set $K$ as 16 because the majority of models we benchmarked utilize 16-frame clips.
Consequently, for an N-frame video, we ensure that each atomic activity is present for a minimum of $N/K$ frames.
Otherwise, annotators will discard the entire scenario.

\begin{table*}[t!]
% \small
\begin{center}
\begin{tabular}{cc@{\;}c@{\;}c@{\;}c@{\;}c}
\vspace{.1cm}

\scriptsize{Object-guided}
\hspace{-.3cm}
& \adjustimage{trim=30mm 0mm 0mm 0mm, clip, height=1.65cm,valign=m}{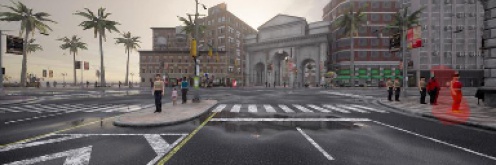}
& \adjustimage{trim=30mm 0mm 0mm 0mm, clip, height=1.65cm,valign=m}{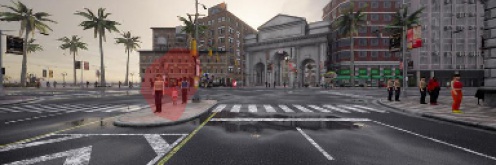}
& \adjustimage{trim=30mm 0mm 0mm 0mm, clip, height=1.65cm,valign=m}{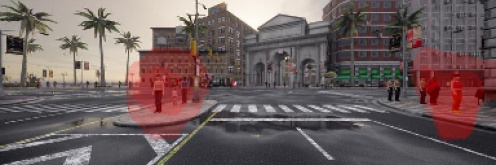}
& \adjustimage{trim=30mm 0mm 0mm 0mm, clip, height=1.65cm,valign=m}{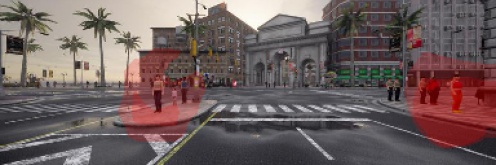}
\\

\vspace{.1cm}
\scriptsize{Action-slot (ours)}
\hspace{-.3cm}

& \adjustimage{trim=30mm 0mm 0mm 0mm, clip, height=1.65cm,valign=m}{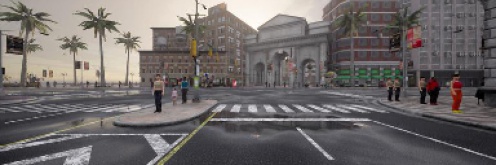}
& \adjustimage{trim=30mm 0mm 0mm 0mm, clip, height=1.65cm,valign=m}{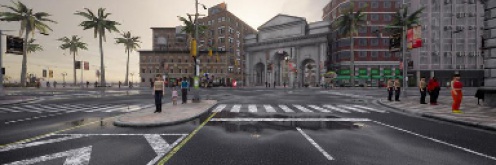}
& \adjustimage{trim=30mm 0mm 0mm 0mm, clip, height=1.65cm,valign=m}{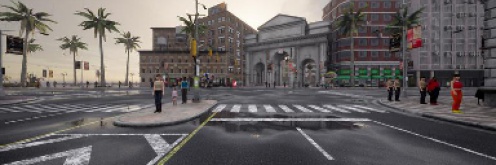}
& \adjustimage{trim=30mm 0mm 0mm 0mm, clip, height=1.65cm,valign=m}{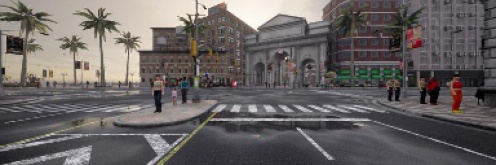}

\\
\\
\vspace{.1cm}

\scriptsize{Object-guided}
\hspace{-.3cm}
& \adjustimage{trim=30mm 0mm 0mm 0mm, clip, height=1.65cm,valign=m}{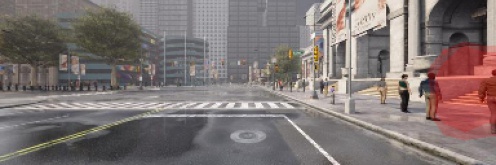}
& \adjustimage{trim=30mm 0mm 0mm 0mm, clip, height=1.65cm,valign=m}{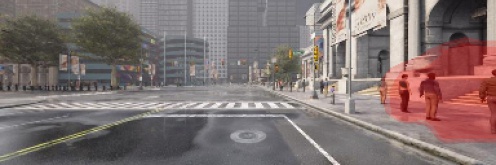}
& \adjustimage{trim=30mm 0mm 0mm 0mm, clip, height=1.65cm,valign=m}{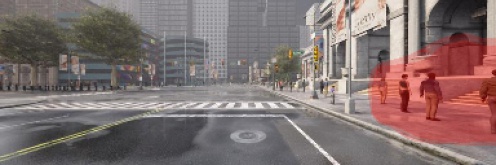}
& \adjustimage{trim=30mm 0mm 0mm 0mm, clip, height=1.65cm,valign=m}{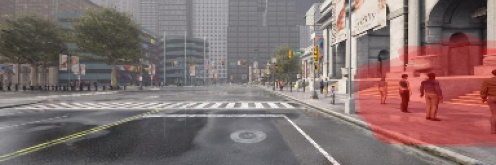}
\\

\vspace{.1cm}
\scriptsize{Action-slot (ours)}
\hspace{-.3cm}

& \adjustimage{trim=30mm 0mm 0mm 0mm, clip, height=1.65cm,valign=m}{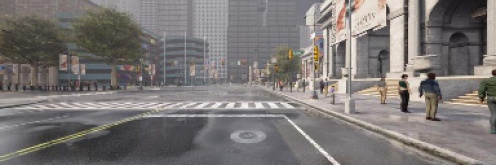}
& \adjustimage{trim=30mm 0mm 0mm 0mm, clip, height=1.65cm,valign=m}{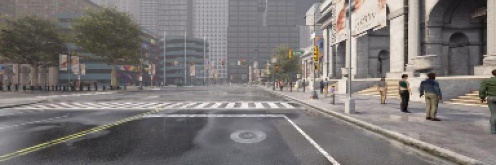}
& \adjustimage{trim=30mm 0mm 0mm 0mm, clip, height=1.65cm,valign=m}{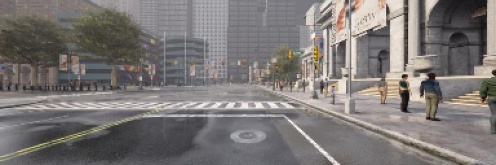}
& \adjustimage{trim=30mm 0mm 0mm 0mm, clip, height=1.65cm,valign=m}{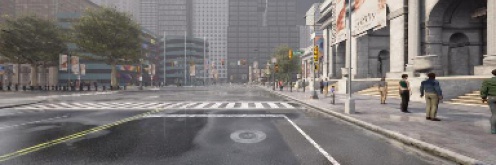}
\end{tabular}
\end{center}
\vspace{-10pt}
\vspace{-.3cm}

\captionof{figure}{
Visualization of attention maps of \textbf{Object-guided} and \textbf{Action-slot}
from slots that predict \textcolor{red}{false positives} on additional scenarios. The scenario presents zero atomic activity but many static road users. False positive predicted by the object-guided model: \textcolor{red}{C1-C4:P}, \textcolor{red}{C4-C1:P} (upper) and \textcolor{red}{C1-C2:P}, \textcolor{red}{C2-C1:P}, \textcolor{red}{C1-C2:P+} (bottom).
The attention scores within the Action-slot do not surpass the threshold in any region.
% The attention scores in the Action-slot are not greater than the threshold in any regions.
}
\label{fig:object_fp}
\end{table*}

\paragraph{Annotation for Ego-vehicle's Action.}
% We annotate the ego-vehicle's action for each scenario. For scenarios where the annotator can not determine the ego vehicle's action, the annotator will label it as \textit{Z1-Z1:E}. For example, the ego-vehicle stops at a traffic light during the entire scenario or the ego-vehicle slowly moves toward the intersection but has not shown a specific action. This is different from OATS where they discard any scenarios without a specific ego-vehicle action.
We provide annotations of the ego-vehicle's actions for every scenario. We annotate \textit{Z1-Z1:E} if the annotator is unable to determine the ego vehicle's action. For instance, when the ego-vehicle remains stationary at a traffic light throughout the entire scenario or moves slowly towards an intersection without exhibiting a specific action. This approach differs from OATS which excludes scenarios lacking a specific ego-vehicle action.

\paragraph{Detailed Dataset Statistics.}
% We have labeled a total of 16521 instances of atomic activity exhibited by various road users. The maximum number of road users in a frame is 37 and 63 in a video. On average, each video contains 2.43 labeled traffic pattern descriptions. The length of the collected videos varies, ranging from 51 frames to 242 frames. The average length of the videos is 109.341 frames with a frame rate of 20Hz.
We have annotated a total of 16,521 instances of atomic activity. 
The maximum number of road users observed in a single frame is 37.
In a video, it reaches up to 63.
On average, each video encompasses 2.43 labeled traffic pattern descriptions. 
The duration of the captured videos varies, ranging from 51 frames to 242 frames. 
The average length of the videos is 109.341 frames, with a frame rate of 20Hz.

\label{sec:sup_nuscenes}

\section{nuScenes Annotations}
We annotate nuScenes~\cite{nuscenes}, one of the most popular real-world traffic scene datasets, for additional experiments. The dataset was collected in Boston and Singapore. 
We follow the same annotation criterion in TACO for nuScenes. 
Specifically, we scan through the \textit{train\_val} set of nuScenes and annotate any scenarios in the 4-way and T-intersections. 
However, the data in nuScenes is collected with a very low frame per second (FPS), we thus select 16 consecutive frames as a short clip.
% We annotate the data collected in Singapore, where is left-hand-drive,  with the same annotation protocol.
Note that we discard the night scenes because of poor visibility.
To this end, we obtain 426 clips and 933 atomic activity labels for the new nuScenes dataset.
The atomic activity class distribution is presented in Figure~\ref{fig:nuscenes_distribution}.
We randomly divide the set with 340 clips for the training set and 86 clips for the testing set.
We downsample the image size to 256 $\times$ 768, which is the same as TACO. For transfer learning, we downsample the image size to the same size with pre-trained datasets, i.e., 224 $\times$ 224 for OATS pre-trained and 256 $\times$ 768 for TACO pre-trained. We neglect the atomic activities involved with grouped two-wheelers (\textit{K+}) when calculating mAP because there are only 2 labels in the whole dataset.
\label{sec:sup_ablation}
\section{More Ablation Study of Action-slot}

\begin{table}[h!]
\centering
\small
\caption{Comparison of using cross-attention and slot-attention in Action-slot on OATS. S1, S2, and S3 denote the three test splits in OATS.}
% \resizebox{0.65\columnwidth}{!}{
\begin{tabular}
            { lc c c |c }
            \toprule
            &
            \multicolumn{1}{c}{S1}  &
            \multicolumn{1}{c}{S2}  &
            \multicolumn{1}{c}{S3}  &
            \multicolumn{1}{c}{mAP}
            \\
            \midrule
            Cross attention & 29.0 & 32.8 & 35.4 & 32.4
            \\
            Slot attention & \textbf{48.1} & \textbf{47.7} & \textbf{48.8} & \textbf{48.2}
            \\
            \bottomrule
        \end{tabular}
\label{tab:query}
\end{table}

\begin{table*}[t!]
% \small
\begin{center}
\begin{tabular}{@{}c@{\hspace{.5em}}c@{\hspace{.5em}}c@{\hspace{.5em}}c@{\hspace{.5em}}c@{\hspace{.5em}}}

% \multicolumn{4}{c}{\includegraphics[width=4cm]{Supp_Figure/static_object/taco/static_object_1_topo.pdf}} \\
% \vspace{-.8cm}
% \\
\vspace{-.7cm}
% \adjustimage{height=1.4cm,valign=m}{Supp_Figure/static_object/taco/frame2.jpg} 
\adjustimage{trim=10mm 0mm 30mm 0mm, clip, height=2cm,valign=m}{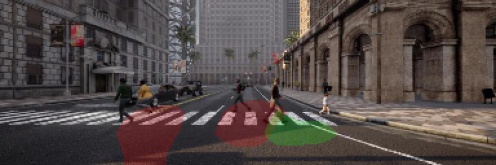}
& \adjustimage{trim=10mm 0mm 30mm 0mm, clip, height=2cm,valign=m}{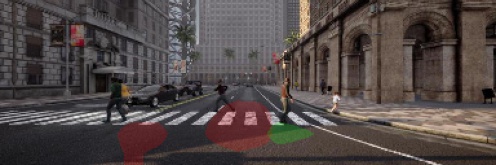}
& \adjustimage{trim=10mm 0mm 30mm 0mm, clip, height=2cm,valign=m}{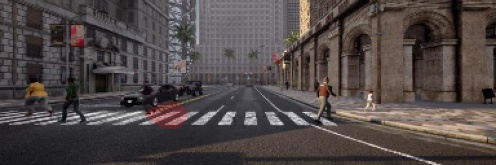} 
& \adjustimage{height=3.8cm,valign=m}{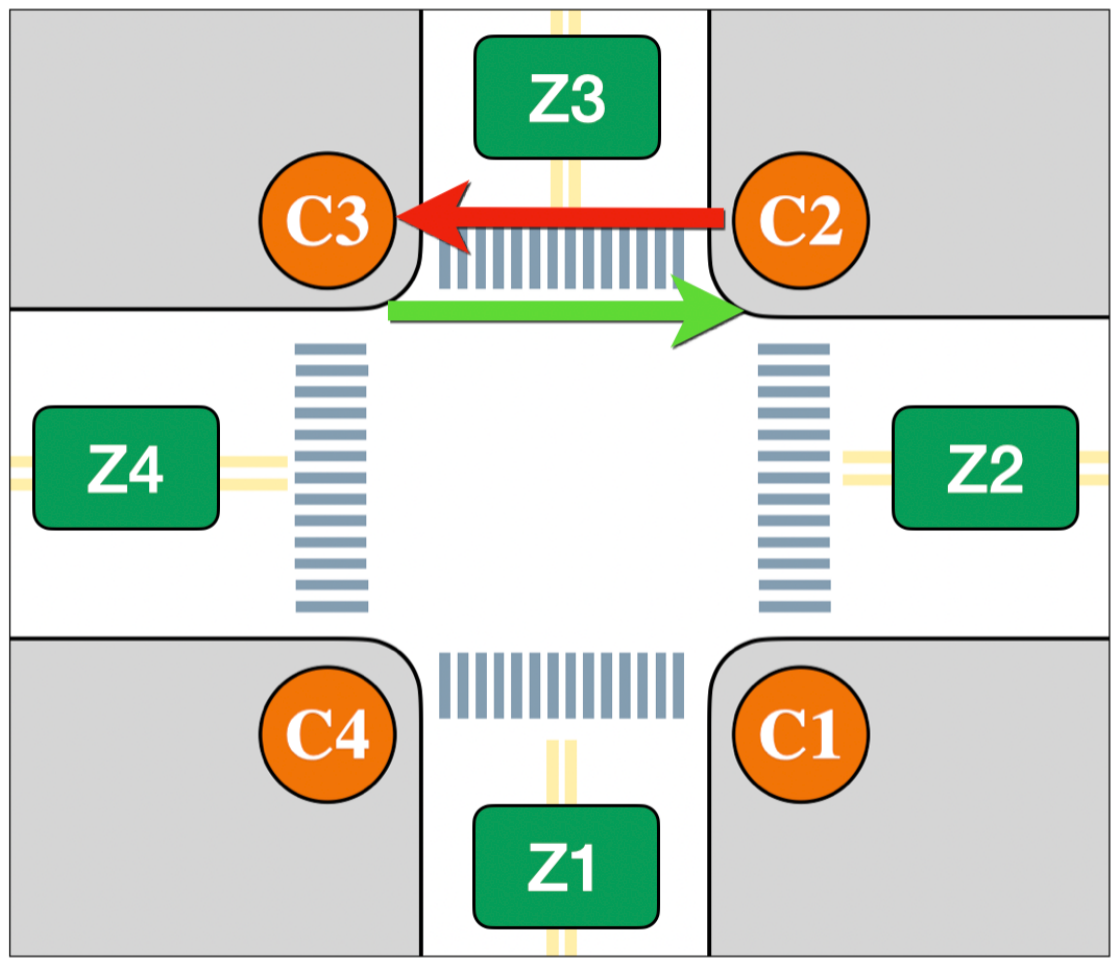} 
\\
\multicolumn{4}{l}{Black vehicles stop at \textit{Z3} in this clip. Atomic activities: \textcolor{red}{C2-C3:P+} and \textcolor{green}{C3-C2:P+}.} 
\vspace{-.8cm}
\\
% \vspace{-.6cm}
% \multicolumn{4}{c}{\includegraphics[width=4cm]{Supp_Figure/static_object/taco/static_object_1_topo.pdf}} \\
\vspace{-.8cm}
\adjustimage{height=2.5cm,valign=m}{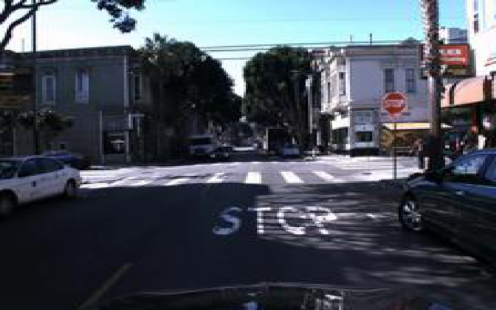}
& \adjustimage{height=2.5cm,valign=m}{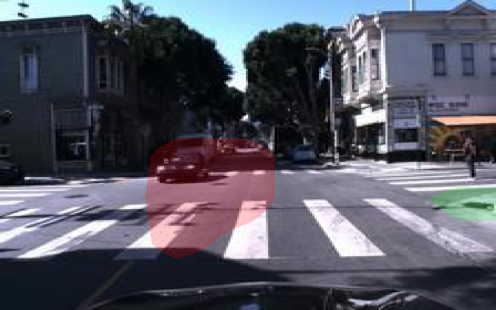}
& \adjustimage{height=2.5cm,valign=m}{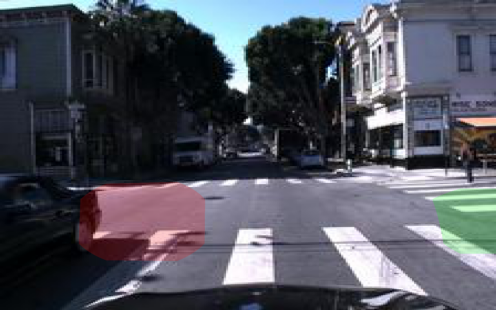}
% & \adjustimage{height=2.5cm,valign=m}{Supp_Figure/static_object/oats/frame13.png} 
& \adjustimage{height=4.5cm,valign=m}{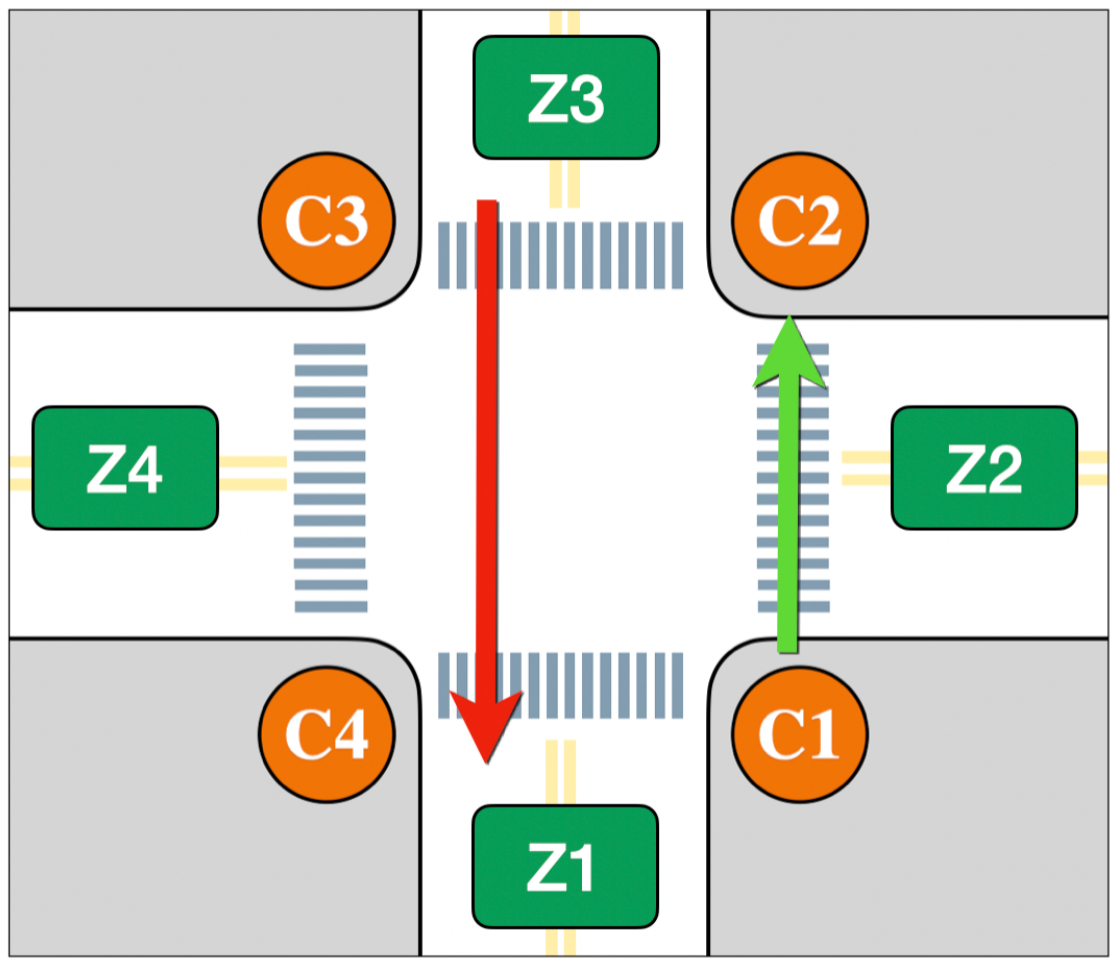}

\\
\multicolumn{4}{l}{Many parked cars on both sides of the street. 
% the ego-vehicle. 
Atomic activities: \textcolor{red}{Z3-Z1:C} and \textcolor{green}{C1-C2:P}.} 
\vspace{-.6cm}

\\
\vspace{-.6cm}

\adjustimage{height=2.3cm,valign=m}{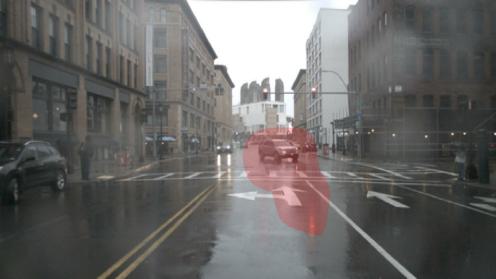}
& \adjustimage{height=2.3cm,valign=m}{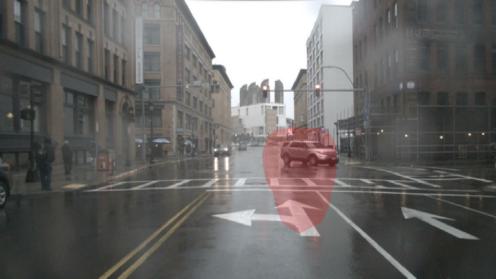}
& \adjustimage{height=2.3cm,valign=m}{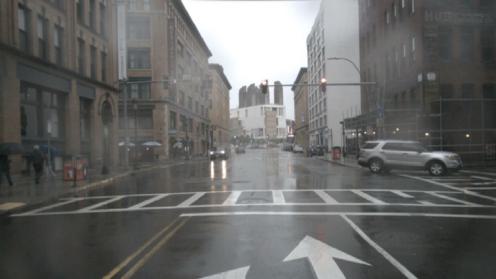}
& \adjustimage{height=4cm,valign=m}{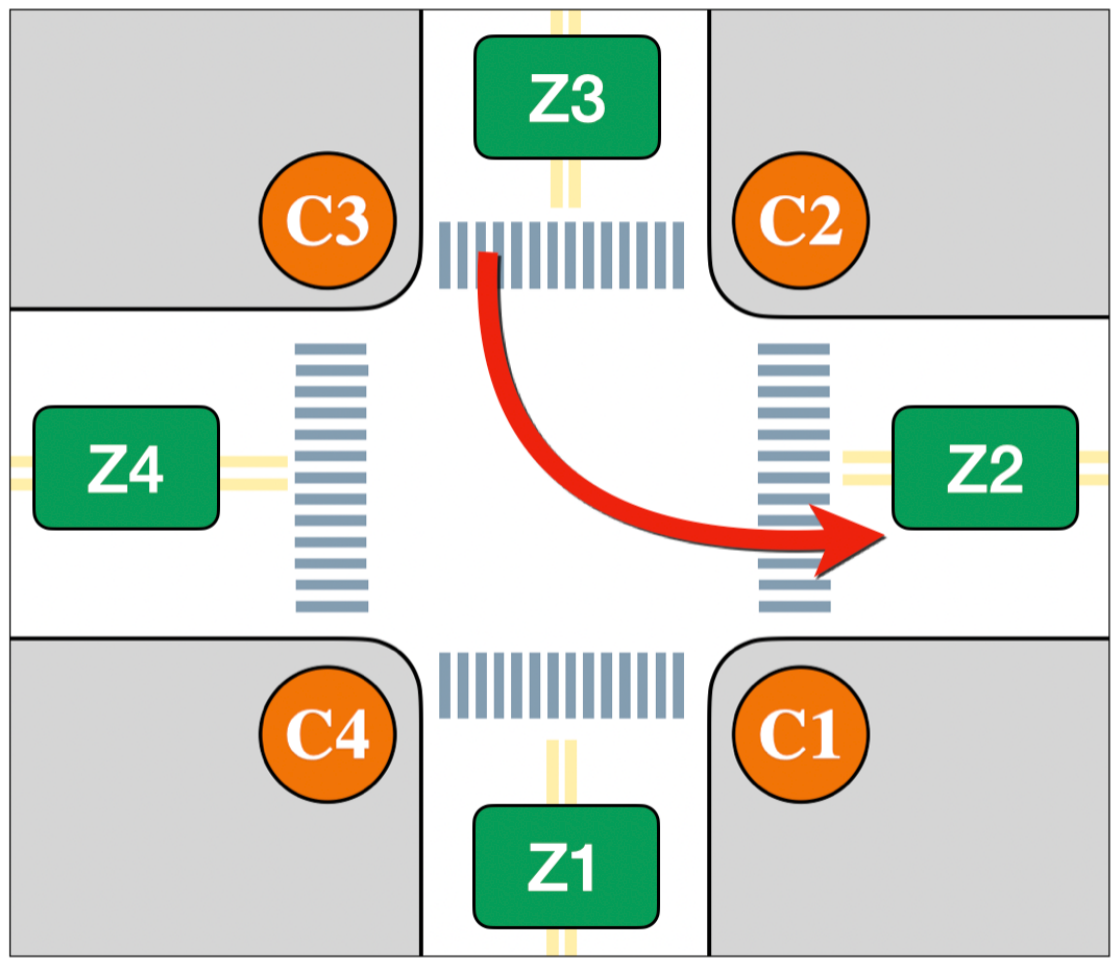}
% & \adjustimage{height=2.3cm,valign=m}{Supp_Figure/static_object/nuscenes/frame9.jpg} 
\\
\multicolumn{4}{l}{Parked car and pedestrians on the left and car stopping temperately at the \textit{Z3}. Atomic activity: \textcolor{red}{Z3-Z2:C}.}

\end{tabular}
\end{center}

\vspace{-.4cm}
\captionof{figure}{
Action-slot's attention visualization in scenarios where atomic activities and static road users are present. We show scenarios from the TACO, OATS~\cite{Agarwal_2023_ICCV}, and nuScenes~\cite{nuscenes} datasets in the first, second, and third row, respectively.}
\label{fig:static_objects}
\end{table*}

\begin{table}[t!]

\centering
% \small
\caption{Results of Boston and Singapore splits in nuScenes.}
\vspace{-1mm}
\begin{tabular}
            {@{}l    c  c  }
            \toprule
            \multirow{2}{*}{ \begin{tabular}{@{\;}c@{\;}} \\\end{tabular}} & 

             \multicolumn{2}{c}{nuScenes}  
             \\
            &

            \begin{tabular}{c@{\;}} Boston   \end{tabular} & 
             \begin{tabular}{c@{\;}} Singapore  \end{tabular} 
            
              \\
            \midrule
            X3D~\cite{feichtenhofer2020x3d} &33.7 &11.6
            \\
            ARG~\cite{CVPR2019_ARG} & 17.2 & 6.6
            \\
            % Slot-VPS & 25.5 &  & 20.7 &
            %  \\
             Action-slot & \textbf{34.7} & \textbf{18.3} 
             \\
            \bottomrule
        \end{tabular}
        
        \label{table:boston_vs_singapore}
        
        \vspace{-3mm}
\end{table}

\paragraph{Object-guided vs. Action-slot.}
% In Table 4 of the main paper, we compare the object guidance with Action-slot using background mask $L_{bg}$ and negative class mask $L_{neg}$, under various numbers of road users present in a video.

We further provide detailed insights into the failure cases of object guidance.
We hypothesize that the object guidance may mislead the model because not all objects are involved in an activity.
% and an activity may not be active all the time. 
% We thus propose to exclude the non-relevant regions and let slot attention to \textbf{dicover} the activities. 
%
% In Table~\ref{tab:object_mask_comparison}, the results justify our hypothesis. 
% %
% The precision score of object guidance is inferior to Action-slot by 5.2\%. 

We create scenarios where many pedestrians are static on sidewalks and not involved in any activities, as shown in Figure~\ref{fig:object_fp}, to better demonstrate the misleading signal caused by the object guidance. 
We visualize the attention from any slot that predicts \textbf{false positive} with red masks. The object-guided method pays attention to the static road users and produces false positive predictions. Moreover, the attention to the false positives is accumulated temporally. We hypothesize this is because the method is not robust to the spatial-temporal features extracted from the backbone~\cite{feichtenhofer2020x3d}. On the other hand, our Action-slot demonstrates the robustness in the scenarios with many static road users.
% by excluding non-relevant regions. Moreover, it verifies the effectiveness of using slot attention to \textbf{discover} activities.

\paragraph{Cross-Attention vs. Slot Attention.}
We study the difference between cross-attention and slot-attention for multi-label atomic activity recognition on OATS. 
% in Action-slot.
Cross-attention can be seen as a query-based method that recently has achieved remarkable success in many tasks~\cite{zhou2023query,hu2023_uniad,vip3d}.
The key difference between cross-attention and slot attention is the dimension to which the softmax operation is applied.
The classic cross-attention~\cite{vaswani2017attention} applies softmax on the tokens, i.e., tokens compete over queries. On the other hand, the softmax in slot attention is applied to the slot dimension, which makes slots compete with each other over the tokens. 
In Table~\ref{tab:query}, we conduct experiments by replacing the slot attention in Action-slot with the classic cross-attention. The results show inferior performance of cross-attention compared to Action-slot using slot attention.
Moreover, we observe that the background guidance loss $L_{bg}$ in the query-based method can not converge well. The results indicate that the cross-attention method may need a stronger supervision signal.
\label{sec:sup_analysis}

\section{Analysis in Challenging Scenarios}

\begin{table*}[t!]
% \small
\begin{center}
\begin{tabular}{c@{\;}c@{\;}c@{\;}c@{\;}c@{\;}c}
\vspace{.2cm}
% % ------
% \hspace{-3mm} 
% \hspace{-4mm}
% & \adjustimage{trim=10mm 0mm 20mm 0mm, clip, height=2cm,valign=m}{Supp_Figure/construction/1/frame6.jpg}
% & \adjustimage{trim=10mm 0mm 20mm 0mm, clip, height=2cm,valign=m}{Supp_Figure/construction/1/frame7.jpg}
% & \adjustimage{trim=10mm 0mm 20mm 0mm, clip, height=2cm,valign=m}{Supp_Figure/construction/1/frame10.jpg}
% & \adjustimage{trim=10mm 0mm 20mm 0mm, clip, height=2cm,valign=m}{Supp_Figure/construction/1/frame11.jpg}
% & \adjustimage{trim=10mm 0mm 20mm 0mm, clip, height=2cm,valign=m}{Supp_Figure/construction/1/frame14.jpg}
& \adjustimage{height=1.9cm,valign=m}{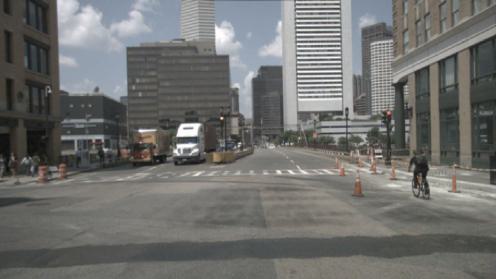}
& \adjustimage{height=1.9cm,valign=m}{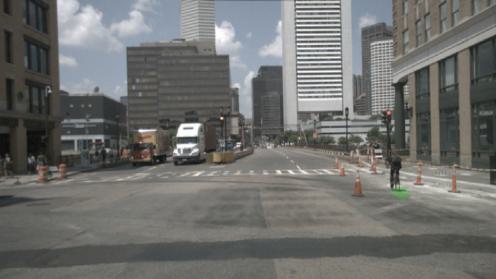}
& \adjustimage{height=1.9cm,valign=m}{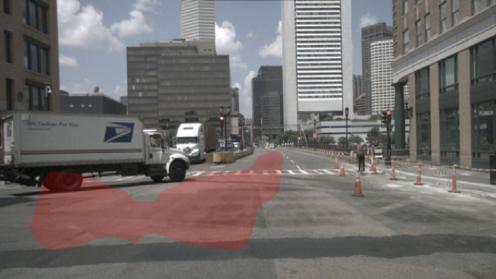}
% & \adjustimage{height=1.8cm,valign=m}{Supp_Figure/construction/1/frame11.jpg}
& \adjustimage{height=1.9cm,valign=m}{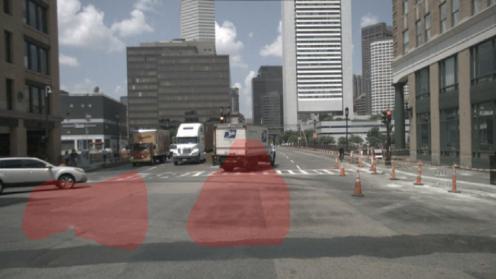}
& \adjustimage{height=1.9cm,valign=m}{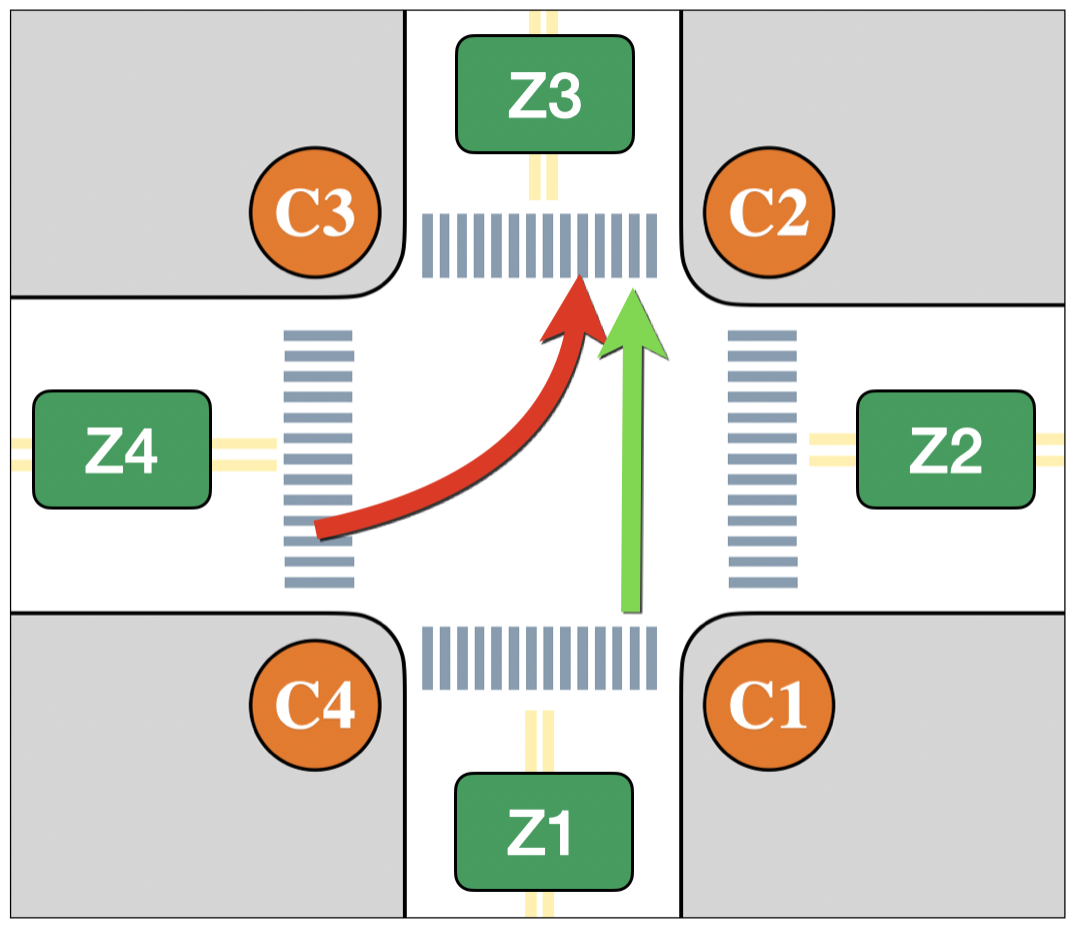}

% \hspace{.04cm}
\\
\vspace{.2cm}

& \multicolumn{5}{l}{Scenario presenting traffic cones in the intersection. Atomic activities: \textcolor{red}{Z4-Z3:C+} and \textcolor{green}{Z1-Z3:K}.} \\

\vspace{.2cm}

% ------
% \hspace{-3mm} 
% \hspace{-8mm}
% & \adjustimage{trim=10mm 0mm 20mm 0mm, clip, height=2cm,valign=m}{Supp_Figure/construction/2/frame0.jpg}
% & \adjustimage{trim=10mm 0mm 20mm 0mm, clip, height=2cm,valign=m}{Supp_Figure/construction/2/frame1.jpg}
% & \adjustimage{trim=10mm 0mm 20mm 0mm, clip, height=2cm,valign=m}{Supp_Figure/construction/2/frame5.jpg}
% & \adjustimage{trim=10mm 0mm 20mm 0mm, clip, height=2cm,valign=m}{Supp_Figure/construction/2/frame10.jpg}
% & \adjustimage{trim=10mm 0mm 20mm 0mm, clip, height=2cm,valign=m}{Supp_Figure/construction/2/frame12.jpg}
& \adjustimage{height=1.9cm,valign=m}{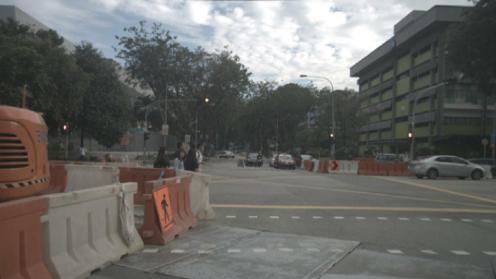}
& \adjustimage{height=1.9cm,valign=m}{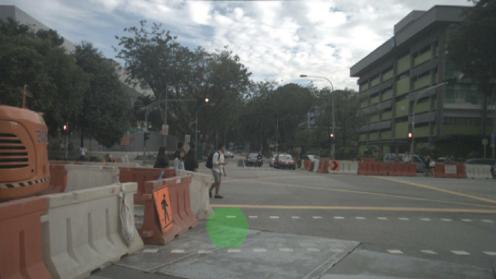}
& \adjustimage{height=1.9cm,valign=m}{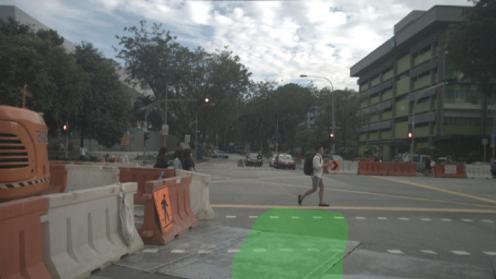}
& \adjustimage{height=1.9cm,valign=m}{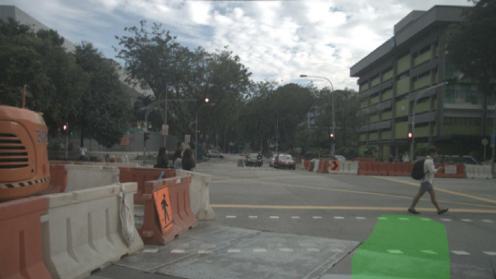}
% & \adjustimage{height=1.8cm,valign=m}{Supp_Figure/construction/2/frame12.jpg}
& \adjustimage{height=1.9cm,valign=m}{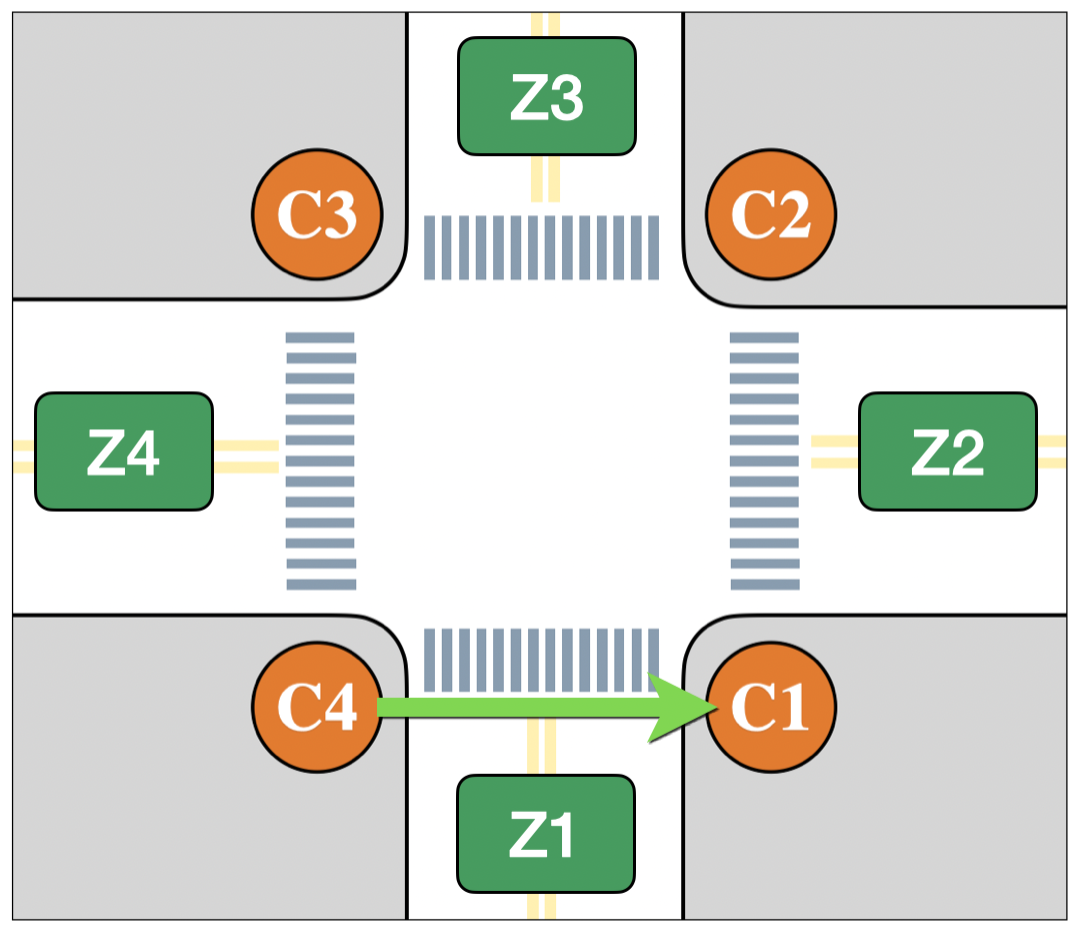}

\\
\vspace{.2cm}

& \multicolumn{5}{l}{Scenario presenting the construction cover the entire corner \textit{C4}. Atomic activities: \textcolor{green}{C4-C1:P}.} \\

\end{tabular}
\end{center}
\vspace{-15pt}
\captionof{figure}{
Action-slot's attention visualization in nuScenes~\cite{nuscenes}. In the two scenarios, road structures are partially occluded by the traffic cones and construction. Colored masks represented the action slots' attention on distinct activities.}
\label{fig:construction}
\end{table*}

% \begin{table}[h]
% \centering
% \scriptsize
% \caption{Comparisons of Action-slot and object-level guidance on TACO across different numbers of road users (denoted as $N$) present in a video. BG and Neg denote background slot and regularization in our method.}
% \begin{tabular}
%             {@{}l@{\;} @{\;} c @{\;}@{\;} @{\;}@{\;}c @{\;}@{\;} @{\;} c @{\;} @{\;} c @{\;}  }
%             \toprule
%             \multirow{1}{*}{ \begin{tabular}{@{\;}c@{\;}} \end{tabular}} & 
%              \multirow{1}*{ \begin{tabular}{@{}c@{}} $N$ $\leq$ 5 \end{tabular}} & 
%              \multirow{1}*{ \begin{tabular}{@{}c@{}}  5 $<$ $N$ $\leq$ 15 \end{tabular}} & 
%              \multirow{1}*{ \begin{tabular}{@{}c@{}} $N$ $>$ 15 \end{tabular}} 
%              &
%              \\
%              \midrule
%              ORN
%              \\
%              ARG~\cite{CVPR2019_ARG}
%              \\
%              \midrule
%              Action-slot (object) & 49.4 & 47.7 & 43.5
%              \\
%              Action-slot (BG+Neg) & \textbf{55.2} & \textbf{50.9} & \textbf{46.3}
%              \\
%             \bottomrule
%         \end{tabular}
        
%         \label{table:object_guidance}
% \end{table}

\paragraph{Atomic Activities and Static Road Users in Scenarios.}
We show qualitative results in scenarios where both atomic activities and multiple static road users are present. In Figure~\ref{fig:static_objects}, the attention learned by Action-slot focuses on the regions where activities occur instead of being distracted by static road users, e.g., vehicles waiting at traffic lights (first and third row), many parked cars (second row), and pedestrians walking on the sidewalk (third row). The results demonstrate the proposed action-centric representations are robust in crowded traffic scenes and can decompose the atomic activities and non-relevant regions from videos.

\paragraph{Boston v.s. Singapore.}
To verify the generalization of Action-slot, we evaluate models on the Boston split and Singapore split in nuScenes~\cite{nuscenes} in Table~\ref{table:boston_vs_singapore}. Note that the scenarios collected in TACO and OATS~\cite{Agarwal_2023_ICCV} are right-hand-traffic. The left-hand-traffic scenarios collected in Singapore thus pose a challenging domain discrepancy for atomic activities. We use the model pretrained on TACO for better performance. Experimental results show that the video-level and object-aware representations both perform inferior in the Singapore split. On the other hand, Action-slot shows the generalization in the scenes with significant domain discrepancy.

\begin{table}[h]
\centering
\small
\caption{Comparisons of pretrained representations from OATS and TACO. We perform transfer learning on the nuScenes dataset.
}
\vspace{-1mm}
\begin{tabular}
            {@{}l    c  c  c }
            \toprule
            \multirow{2}{*}{ \begin{tabular}{@{\;}c@{\;}} \\ \end{tabular}} & 

             \multicolumn{3}{c}{nuScenes}  
             \\
            &
            \begin{tabular}{c@{\;}} Kinetics   \end{tabular} & 
            \begin{tabular}{c@{\;}} + OATS   \end{tabular} & 
             \begin{tabular}{c@{\;}} +TACO  \end{tabular} 
            
              \\
            \midrule
            X3D~\cite{feichtenhofer2020x3d} & 19.8 & 18.9 & 27.8 
            \\
            ARG~\cite{CVPR2019_ARG} &12.2 & 12.7 & 17.0 
            \\
            % Slot-VPS & 25.5 &  & 20.7 &
            %  \\
             Action-slot & 23.6 & 23.6 & \textbf{32.3} 
             \\
            \bottomrule
        \end{tabular}
        \label{table:oats_vs_taco}  
        % \vspace{-3mm}
\end{table}

% \begin{table}[h]
% \centering
% \small
% \caption{Comparisons of direct test results on nuScenes for models pre-trained on TACO and OATS.
% }
% \vspace{-1mm}
% \begin{tabular}
%             {@{}l    c  c  }
%             \toprule
%             \multirow{2}{*}{ \begin{tabular}{@{\;}c@{\;}} \\ \end{tabular}} & 

%              \multicolumn{2}{c}{nuScenes}  
%              \\
%             &
%             \begin{tabular}{c@{\;}} + OATS   \end{tabular} & 
%              \begin{tabular}{c@{\;}} +TACO  \end{tabular} 
            
%               \\
%             \midrule
%             X3D~\cite{feichtenhofer2020x3d} & 14.0 & 19.1
%             \\
%              Action-slot & 17.7 & \textbf{20.7} 
%              \\
%             \bottomrule
%         \end{tabular}
%         \label{table:test}  
%         % \vspace{-3mm}
% \end{table}

\paragraph{Occluded Road Topology.}
We present the qualitative results to demonstrate that Action-slot can handle challenging scenarios where the road topology is significantly occluded in Figure.~\ref{fig:construction}. We observe Action-slot can accurately predict and localize the activities despite the intersection (upper) and corner (lower) occupied by the traffic cones and construction, respectively. This demonstrates the strong generalization of Action-slot on recognizing road topology and the reasoning ability of road topology.

\begin{table*}[t!]
% \small
\begin{center}
\begin{tabular}{c@{\;}c@{\;}c@{\;}c@{\;}c@{\;}c}
& \adjustimage{trim=10mm 0mm 40mm 0mm, clip, height=1.8cm,valign=m}{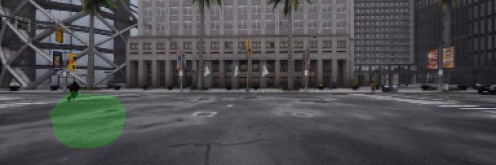}
& \adjustimage{trim=10mm 0mm 40mm 0mm, clip, height=1.8cm,valign=m}{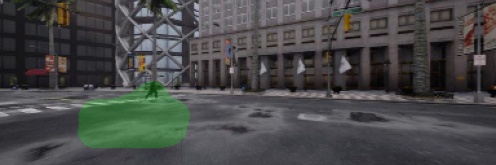}
& \adjustimage{trim=10mm 0mm 40mm 0mm, clip, height=1.8cm,valign=m}{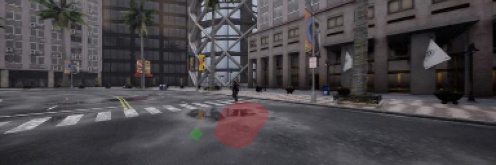}
& \adjustimage{trim=10mm 0mm 40mm 0mm, clip, height=1.8cm,valign=m}{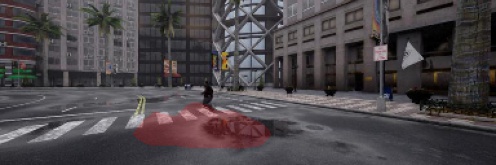}
& \adjustimage{trim=10mm 0mm 40mm 0mm, clip, height=1.8cm,valign=m}{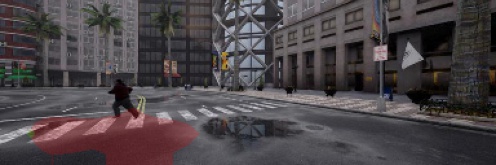}

\end{tabular}
\end{center}
\vspace{-10pt}
\captionof{figure}{
Action-slot's attention visualization in the TACO scenario where a crossing pedestrian first performs action \textcolor{red}{C4-C3:P} then returns with \textcolor{green}{C3-C4:P}.}
\label{fig:multiple_actions}
\end{table*}

\paragraph{Road User with Multiple Actions.}
We find that Action-slot can handle the scenarios where a road user performs multiple actions consecutively. In Figure~\ref{fig:multiple_actions}, Action-slot accurately predicts the two atomic activities involved with the crossing pedestrian and spatial-temporally localizes the transition of two actions. The result again demonstrates the effectiveness of the proposed action-centric representations for the action-aware task.
\label{sec:sup_analysis_taco}
\section{Analysis of The TACO Dataset}

\paragraph{OATS Pretrain v.s. TACO Pretrain.}
We compare the pertained representations learned from OATS and TACO by fine-tuning them on nuScenes in Table~\ref{table:oats_vs_taco}. The results show that models pretrained on TACO outperform the ones pretrained on OATS, verifying the real-world value and transferability of the proposed TACO dataset.

\begin{table*}[t!]
\vspace{1.cm}
\begin{center}

\begin{tabular}{c@{\;}c@{\;}c@{\;}c@{\;}c@{\;}c}
% \hspace{-3mm} 
\hspace{-4mm}
% & \adjustimage{trim=35mm 5mm 45mm 10mm, clip, height=1.9cm,valign=m}{Supp_Figure/occlusion/frame2.jpg}
& \adjustimage{trim=35mm 5mm 45mm 10mm, clip, height=2cm,valign=m}{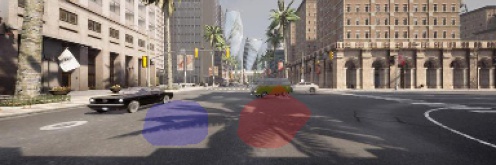}
& \adjustimage{trim=35mm 5mm 45mm 10mm, clip, height=2cm,valign=m}{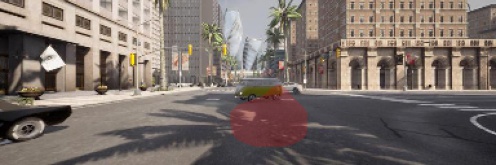}
& \adjustimage{trim=35mm 5mm 45mm 10mm, clip, height=2cm,valign=m}{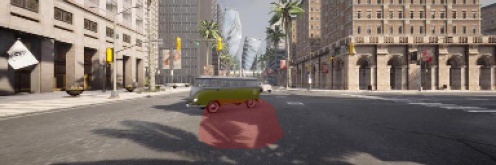}
& \adjustimage{trim=35mm 5mm 45mm 10mm, clip, height=2cm,valign=m}{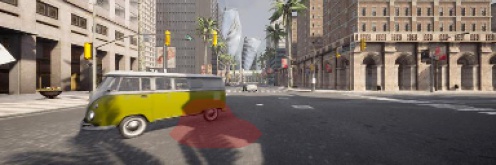}
& \adjustimage{height=1.9cm,valign=m}{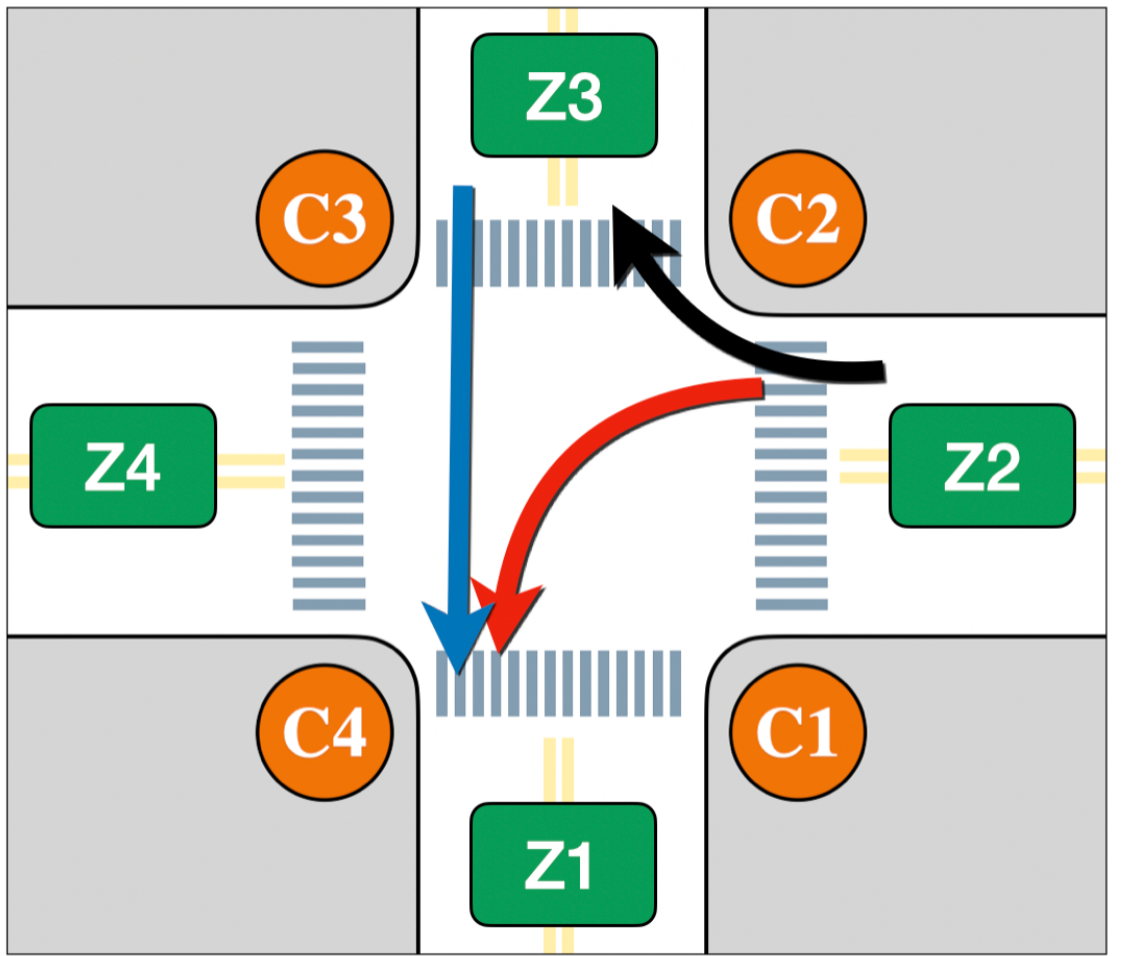}

\end{tabular}
\end{center}
\vspace{-.5cm}
\captionof{figure}{
Action-slot's attention visualization in the TACO scenario where an atomic activity is occluded. The yellow bus with red masks partially occludes the white car on the right side (Z2). Action-slot successfully predicts  \textcolor{blue}{Z3-Z1:C} and \textcolor{red}{Z2-Z1:C} but misses the occluded white car \textit{Z2-Z3:C} (black arrow in illustration).}
\label{fig:occlusion}
\end{table*}

\begin{table*}[t!]
\centering
\scriptsize
\caption{Results of all 64 classes of atomic activity. Each grouped row is divided by the type of road users involved.}
\vspace{-1mm}
\begin{tabular}
            {@{}l  @{\;}@{\;}c@{\;}@{\;}  @{\;}@{\;}c@{\;}@{\;}  @{\;}@{\;}c@{\;}@{\;} @{\;}@{\;}c@{\;}@{\;}  @{\;}@{\;}c@{\;}@{\;} @{\;}@{\;}c@{\;}@{\;}  @{\;}@{\;}c@{\;}@{\;}  @{\;}@{\;}c@{\;}@{\;}  @{\;}@{\;}c@{\;}@{\;}  @{\;}@{\;}c@{\;}@{\;} @{\;}@{\;}c@{\;}@{\;} c}
            \toprule
            \multirow{2}{*}{ \begin{tabular}{@{\;}c@{\;}} \\\end{tabular}} 

            & 
            Z1-Z2:C & Z1-Z3:C & Z1-Z4:C & 
            Z2-Z1:C & Z2-Z3:C & Z2-Z4:C &
            Z3-Z1:C & Z3-Z2:C & Z3-Z4:C &
            Z4-Z1:C & Z4-Z2:C & Z4-Z3:C 

              \\
            \midrule
            X3D~\cite{feichtenhofer2020x3d} 
            &
            21.1 & 27.1 & 31.5 & 28.2 & 33.4 & 21.8 &
            28.4 &  34.5 & 18.9 & 22.1 &  28.8 & 32.1
            \\
            ARG~\cite{CVPR2019_ARG} & 36.2 & 17.9 & 23.6 & 25.9 & 26.8 & 14.5 & 15.3 & 15.7 & 16.2 & 41.4 & 30.0 & 19.1
            \\
             Action-slot 
            &
            \textbf{48.5} & \textbf{47.9} & \textbf{53.1} & \textbf{57.7} & \textbf{54.1} & \textbf{45.9} & 
            \textbf{41.5} & \textbf{43.5} & \textbf{47.0} & \textbf{48.1} & \textbf{44.8} & \textbf{44.7}
             \\
            \bottomrule
            \\
            \\
            \\
            \toprule
            \multirow{2}{*}{ \begin{tabular}{@{\;}c@{\;}} \\\end{tabular}} 
            &
            Z1-Z2:C+ & Z1-Z3:C+ & Z1-Z4:C+ & 
            Z2-Z1:C+ & Z2-Z3:C+ & Z2-Z4:C+ &
            Z3-Z1:C+ & Z3-Z2:C+ & Z3-Z4:C+ &
            Z4-Z1:C+ & Z4-Z2:C+ & Z4-Z3:C+
            \\
            \midrule
            X3D~\cite{feichtenhofer2020x3d} 
            & 
            76.6 & 42.7 & 3.1 & 61.7 & 58.7 & 48.6 & 
            41.1 & 60.9 & 64.5 & 76.4 & 73.4& \textbf{70.8}
            \\
            ARG~\cite{CVPR2019_ARG} & 32.8 & 11.0 & 0.1 & 38.3 & 3.8 & 17.8 & 9.8 & 13.2 & 9.4 & 4.1 & 15.6 &  1.2
            \\
             Action-slot
             & 
             \textbf{86.9} & \textbf{63.7} & \textbf{28.7} & \textbf{74.9} & \textbf{59.5} & \textbf{75.2} & 
             \textbf{64.6} & \textbf{79.0} & \textbf{79.1} & \textbf{82.1} & \textbf{78.0}  &69.7
             \\
            \bottomrule

            \\
            \\
            \\
            \toprule
            \multirow{2}{*}{ \begin{tabular}{@{\;}c@{\;}} \\\end{tabular}} 
            &
            Z1-Z2:K & Z1-Z3:K & Z1-Z4:K & 
            Z2-Z1:K & Z2-Z3:K & Z2-Z4:K &
            Z3-Z1:K & Z3-Z2:K & Z3-Z4:K &
            Z4-Z1:K & Z4-Z2:K & Z4-Z3:K
            \\
            \midrule
            X3D~\cite{feichtenhofer2020x3d} 
            & 
            17.3 & 27.9 & 24.9 & 36.0 & 10.6 & 14.2 &
            30.5 & 21.5 & 9.1 & 13.7 & 21.7 & 13.0

            \\
            ARG~\cite{CVPR2019_ARG} & 
            \textbf{55.0} & 17.3 & 0.7 & 50.3 & 19.2 & 20.7 & 
            22.7 & \textbf{57.9} & 10.3 & \textbf{46.9} & 25.6 &  23.2
            \\
             Action-slot
             & 
             39.9 & \textbf{54.2} & \textbf{45.3} & \textbf{51.1} & \textbf{30.3} & \textbf{41.9} &
             \textbf{46.1} & 36.9 & \textbf{36.9} & 35.5 & \textbf{38.8} & \textbf{35.4}
             \\
            \bottomrule

            \\
            \\
            \\
            \toprule
            \multirow{2}{*}{ \begin{tabular}{@{\;}c@{\;}} \\\end{tabular}} 
            &
            Z1-Z2:K+ & Z1-Z3:K+ & Z1-Z4:K+ & 
            Z2-Z1:K+ & Z2-Z3:K+ & Z2-Z4:K+ &
            Z3-Z1:K+ & Z3-Z2:K+ & Z3-Z4:K+ &
            Z4-Z1:K+ & Z4-Z2:K+ & Z4-Z3:K+
            \\
            \midrule
            X3D~\cite{feichtenhofer2020x3d} 
            & 
            76.7 & 31.0 & 1.0 & \textbf{83.5} & 26.3 &  53.2 &
            54.1 & \textbf{75.0} & 36.6 & 40.4 & \textbf{79.8} & \textbf{55.2}
            \\
            ARG~\cite{CVPR2019_ARG} & 22.1 & \textbf{22.1} & 20.6 & 5.0 & 11.3 & 3.9 & 24.2 & 7.0 & 2.5 & 5.3 & 20.4 & 9.1
            
            \\
             Action-slot
             & 
             \textbf{93.7} & \textbf{56.7} & 7.7 & 67.3 & \textbf{51.9} & \textbf{65.7} &
             \textbf{74.3} & 74.0 & \textbf{68.3} & \textbf{60.6} & 76.8 & 48.4
             \\
            \bottomrule
            \\
            \\
            \\
            \toprule
            \multirow{2}{*}{ \begin{tabular}{@{\;}c@{\;}} \\\end{tabular}} 
            &
            C1-C2:P & C1-C4:P &
            C2-C1:P & C2-C3:P &
            C3-C2:P & C3-C4:P &
            C4-C1:P & C4-C3:P

            \\
            \midrule
            X3D~\cite{feichtenhofer2020x3d} 
            & 
            34.4 & 43.5 &
            38.9 & 35.0 &
            25.1 & 29.6 &
            45.5 & 27.6
            \\
            ARG~\cite{CVPR2019_ARG} & 31.1 & 38.7 & 23.7 & 12.8 & 10.6 & 22.5 & 39.3 & 15.3
            \\
             Action-slot
             & 
             \textbf{52.4} & \textbf{60.6} &
             \textbf{55.9} & \textbf{42.5} & 
             \textbf{44.3} & \textbf{38.2} &
             \textbf{65.2} & \textbf{34.2}
             \\
            \bottomrule
            \\
            \\
            \\
            \toprule
            \multirow{2}{*}{ \begin{tabular}{@{\;}c@{\;}} \\\end{tabular}} 
            &
            C1-C2:P+ & C1-C4:P+ &
            C2-C1:P+ & C2-C3:P+ &
            C3-C2:P+ & C3-C4:P+ &
            C4-C1:P+ & C4-C3:P+

            \\
            \midrule
            X3D~\cite{feichtenhofer2020x3d} 
            & 
            39.2 & 53.3 & 
            47.3 & 24.2 &
            23.2 & 32.4 &
            62.6 & 29.1
            \\
            ARG~\cite{CVPR2019_ARG} & 19.6 & 24.8 & 18.0 & 6.1 & 5.7 & 9.4 & 24.6 & 8.4
            \\
             Action-slot
             & 
             \textbf{61.0} & \textbf{74.5} & 
             \textbf{65.0} & \textbf{34.8} & 
             \textbf{33.0} & \textbf{37.4} &
             \textbf{80.8} & \textbf{35.3}
             \\
            \bottomrule
            
        \end{tabular}
        
        \label{table:class}
        
        \vspace{-3mm}
\end{table*}

\paragraph{Activity Classes Analysis.}

We report the performance of models for all 64 classes of atomic activities, which can not be achieved in OATS~\cite{Agarwal_2023_ICCV} and nuScenes~\cite{nuscenes}. We find two interesting observations. First, most results of activities involved with grouped road users (e.g., \textit{C1-C2:C+}) are better than the ones involved with a single road user (e.g., \textit{C1-C2:C}), in which one possible reason is that the larger regions of interest are easier to predict. Second, the smaller or more distant activities are more challenging, which can be attributed to the insufficient representations learned from the backbone.

\label{sec:limitations}

\section{Limitation}

\paragraph{Action-slot.}
We find Action-slot performs less effectively in the occluded scenarios where the activities are visually overlapped. In Figure~\ref{fig:occlusion}, a bus with \textit{Z2-Z1} action occludes a white car on the right side with action \textit{Z2-Z3}. Action-slot successfully predicts the bus's action \textit{Z2-Z1} but fails to predict \textit{Z2-Z3} and can not localize it via attention. 
We hypothesize that this is because the occlusion may confuse the competition mechanism in slot attention, i.e., two slots compete over the overlapped regions.
We hope our findings can inspire the community to discover more advanced action-centric representations that can handle occlusion issues.

\paragraph{The TACO Dataset.}
We observe the 64 classes of atomic activities in TACO can not fully cover the diverse events in traffic scenes. For example, vehicles can only move between roadways and pedestrians can only move between two near corners. However, two-wheelers can move between corners, e.g., \textit{C1-C2:K+}, and pedestrians can also move diagonally, e.g., \textit{C1-C3:P}. These atomic activities are important to many applications, such as safety-critical scenario generation~\cite{xu2022safebench,hanselmann2022king,rempe2022strive}. However, the existing auto-pilot mechanism in the CARLA simulator does not support the collection of such atomic activities. We aspire for our research to inspire collaborative efforts within the community to improve existing atomic activity datasets. This involves gathering larger, more diverse datasets from real-world scenarios and advancing the sophistication of simulator-based auto-pilots.

\label{sec:sup_implementation}

\section{Implementation Details}

\paragraph{Data Preprocessing.}
We downsample a video into a fixed-length short clip as models' input, which is common practice in video recognition~\cite{sigurdsson2016hollywood,Agarwal_2023_ICCV,tan2023egodistill,ego-env}. Specifically, we randomly sample subsequent with uniform intervals between frames for the training set and fix the subsequent for the testing set. Note that we neglect this process for nuScenes because we annotate each sample as a fixed 16-frame clip.

\paragraph{Ego-vehicle's action.}
To enhance the awareness of ego motion, we include a module for all models to predict ego-vehicle's action via global features. The global features $F_{ego} \in \mathbb{R}^{256}$ are generated by applying a Conv3D with kernel size 1. 
Since each video must have a label for the corresponding ego-vehicle's action, the models output a multi-class prediction with a fully connected layer and softmax operation. Note that we neglect the ego vehicle's action prediction in OATS~\cite{Agarwal_2023_ICCV} since the absence of the annotation in the released dataset.
In this paper, we do not report the accuracy of predicting ego-vehicle's pattern because the performance of all models saturates with nearly 100\%.

\paragraph{Architecture of Action-slot.}
In this work, we eliminate the GRU from the slot updating process~\cite{locatello2020object,kipf2022conditional,elsayed2022savi++,bao2022discovering,zhou2022slot} for Action-Slot. This is because of its negligible impact on enhancing performance in our experiments.

\paragraph{Training for transformer-based methods.}
We freeze the pretrained transformer blocks except for the first three and last three blocks during training. 
We downsample the input image to $224 \times 224$ for both MViT~\cite{fan2021multiscale} and VideoMAE~\cite{tong2022videomae} on all datasets. 
We attempt to adapt the pre-trained positional embedding of MViT as suggested in ViT~\cite{dosovitskiy2021an} by interpolating spatial positional embedding. For example, we adapt token numbers from $7 \times 7$ to $8 \times 24$ in our TACO dataset. However, we find the performance degraded with interpolation. We thus simply use image size $224 \times 224$ to match the original positional embedding size.

\paragraph{Training for object-aware methods.}
We follow OATS~\cite{Agarwal_2023_ICCV} to set the number of object proposals to 20. Specifically, we select the 20 largest bounding boxes in each frame.
To match the ground-truth atomic activity labels with the length of proposals, we pad the ground truth with the negative class and use a Hungarian matcher to associate them during training. It is worth noting that because object-aware models output multi-class results for each proposal, we rearrange the outputs to a set for calculating the metrics.

\paragraph{Training for slot-based methods.}
In order to adapt slot-based baselines to atomic activity recognition, we use the last state of slots as the input to the classifier for the recurrent fashion, i.e., SAVi~\cite{kipf2022conditional,elsayed2022savi++} and MO~\cite{bao2022discovering}. As for Slot-VPS~\cite{zhou2022slot}, we sum up the slots across temporal dimensions.
For all slot-based baselines, including our Action-Slot, we set both the dimensions of slots and image features to 256.

\paragraph{Backbone modification for Action-Slot.}
We use the features of the last convolution block as the input to action slots for all backbone encoders except for SlowFast. SlowFast processes two paths: path \emph{Fast} takes an original input sequence (16 frames) as input and the path \emph{Slow} takes the sub-sequence with $1/4$ length (4 frames). We apply a pooling operation to the output of path \emph{Fast} for aligning the length of path \emph{Slow} and combine them with channel-wise concatenation. We freeze the entire backbone except for the last ConvBlock.

\paragraph{Hyperparameters.}
All the models including our Action-Slot are trained for 50, 100, and 100 epochs on OATS~\cite{Agarwal_2023_ICCV}, TACO, and nuscenes~\cite{nuscenes}, respectively. We use AdamW optimizer~\cite{loshchilov2018decoupled} with a batch size of 8. The learning rate varies from 1e-4 to 5e-5 and weight decay varies from 0.1 to 0.0001. We apply a 50\% dropout to all models' last features layer, e.g., X3D's last ConvBlock.
We conduct all experiments with a single NVIDIA 3090 GPU with 24GB.


\end{document}